\documentclass[fleqn,10pt]{wlscirep}
\usepackage[utf8]{inputenc}
\usepackage[T1]{fontenc}
\usepackage{lineno}
\linenumbers

\usepackage{import}
\usepackage{amsmath,graphicx}
\usepackage{times}
\usepackage[dvipsnames]{xcolor}

\usepackage{tabularx,booktabs}
\usepackage{arydshln}
\usepackage{tabu}
\usepackage{adjustbox}
\usepackage{enumerate,multicol}
\usepackage{multirow}
\usepackage{multicol}
\usepackage{caption}
\usepackage{subcaption}
\usepackage{gensymb}
\usepackage{float}
\usepackage{balance}
\usepackage{xspace}
\usepackage{amssymb}
\usepackage{bbding,enumitem}
\usepackage{booktabs}
\usepackage{hyperref}
\usepackage{color, colortbl}
\hypersetup{
    colorlinks,
    linkcolor={red!50!black},
    citecolor={blue!50!black},
    urlcolor={blue}
}
\usepackage{overpic}
\usepackage{times}

\usepackage{soul}
\usepackage{url}
\usepackage[section]{placeins}

\usepackage{amsmath}
\usepackage{nicefrac, xfrac}
\usepackage{booktabs}

\definecolor{gray1}{RGB}{210,210,210}
\definecolor{gray2}{RGB}{50,50,50}
\definecolor{gray3}{RGB}{150,150,150}
\definecolor{RYB1}{RGB}{218,232,252}
\definecolor{RYB4}{RGB}{108,142,191}
\definecolor{shadecolor1}{rgb}{0.95,0.95,0.95}
\definecolor{blue2}{RGB}{230,240,250}

\makeatletter
\newcommand\avsuminner[2]{%
  {\sbox0{$\m@th#1\sum$}%
   \vphantom{\usebox0}%
   \ooalign{%
     \hidewidth
     \smash{\vrule height\dimexpr\ht0+1pt\relax depth\dimexpr\dp0+1pt\relax}%
     \hidewidth\cr
     $\m@th#1\sum$\cr
   }%
  }%
}

\usepackage{verbatim}

\usepackage{tikz}

\newcommand{\DrawPercentageBar}[1]{%
  \begin{tikzpicture}
    \fill[color=gray2]   (0.0 , 0.0) rectangle (#1*0.04ex , 1.5ex );
    \fill[color=gray1] (#1*0.04ex  , 0.0) rectangle (4.0ex, 1.5ex);
  \end{tikzpicture}%
}

\newcommand{\DrawCellBar}[1]{%
  \begin{tikzpicture}
    \fill[color=blue2]   (0 , -1ex) rectangle (#1*0.15ex , 2ex );
  \end{tikzpicture}%
}

\newcommand{\DP}[2]{%
  \begin{tikzpicture}
    \fill[color=#2]   (0.0 , 0.0) rectangle (#1*6.2ex , 2ex );
  \end{tikzpicture}%
}
\newcommand{\NP}[1]{
  #1 \DrawPercentageBar{#1}
}
\usepackage{calc}
\usepackage{stackengine}
\newcommand\clapp[3][0pt]{\stackengine{0pt}{#3}{\kern#1#2}{O}{c}{F}{F}{L}}

\newcommand{\NSP}[2]{
  #1 \textcolor{gray3}{\scriptsize{$\pm$ #2}} \DrawPercentageBar{#1}
}

\newcommand{\NPB}[2]{
  #1 \textcolor{gray2}{\scriptsize{(#2\%)}} 
}

\newcommand{\DCB}[2]{
\makebox[0.5ex][l]{\DrawCellBar{#2}}
  \NPB{#1}{#2}
}
\usepackage{pgfplots}
\usepackage{pgfplotstable}
\usetikzlibrary{pgfplots.groupplots}
\usepgfplotslibrary{colorbrewer}
\pgfplotsset{/pgfplots/error bars/error bar style={black,thick}}

\pgfplotsset{compat=1.11,
        /pgfplots/ybar legend/.style={
        /pgfplots/legend image code/.code={%
        \draw[##1,/tikz/.cd,bar width=3pt,yshift=-0.2em,bar shift=0pt]
                plot coordinates {(0cm,0.8em)};},
},}

\title{Cataract-1K: Cataract Surgery Dataset for Scene Segmentation, Phase Recognition, and Irregularity Detection}

\author[1]{Negin Ghamsarian}
\author[3]{Yosuf El-Shabrawi}
\author[2]{Sahar Nasirihaghighi}
\author[3]{Doris Putzgruber-Adamitsch}
\author[4]{Martin Zinkernagel}
\author[4]{Sebastian Wolf}
\author[2*]{Klaus Schoeffmann}
\author[1]{Raphael Sznitman}

\affil[1]{Center for Artificial Intelligence in Medicine (CAIM), Department of Medicine, University of Bern, Switzerland}
\affil[2]{Department of Information Technology, University of Klagenfurt, Austria}
\affil[3]{Department of Ophthalmology, Klinikum Klagenfurt, Austria}
\affil[4]{Department of Ophthalmology, Inselspital, Bern, Switzerland}

\affil[*]{corresponding author: (ks@itec.aau.at)}

\begin{abstract}

In recent years, the landscape of computer-assisted interventions and post-operative surgical video analysis has been dramatically reshaped by deep-learning techniques, resulting in significant advancements in surgeons' skills, operation room management, and overall surgical outcomes. 
However, the progression of deep-learning-powered surgical technologies is profoundly reliant on large-scale datasets and annotations. Particularly, surgical scene understanding and phase recognition stand as pivotal pillars within the realm of computer-assisted surgery and post-operative assessment of cataract surgery videos. In this context,  we present the largest cataract surgery video dataset that addresses diverse requisites for constructing computerized surgical workflow analysis and detecting post-operative irregularities in cataract surgery.
We validate the quality of annotations by benchmarking the performance of several state-of-the-art neural network architectures for phase recognition and surgical scene segmentation. Besides, we initiate the research on domain adaptation for instrument segmentation in cataract surgery by evaluating cross-domain instrument segmentation performance in cataract surgery videos. The dataset and annotations will be publicly available upon acceptance of the paper.

\end{abstract}
\begin{document}
\nolinenumbers
\flushbottom
\maketitle

\section*{Background \& Summary}

Following the technological advancements in surgery, operation rooms are evolving into intelligent environments. Context-aware systems (CAS) are emerging as pivotal components of this evolution, empowered to advance pre-operative surgical planning \cite{9787557, quon2020deep, 9336225}, automate skill assessment \cite{yanik2022deep, lam2022machine, wang2018deep, wang2018satr, soleymani2021surgical}, support operation room planning \cite{aksamentov2017deep, twinanda2018rsdnet, marafioti2021catanet}, and interpret the surgical context comprehensively. By enabling real-time alerts and offering decision-making support, these systems prove invaluable, especially but not only for less-experienced surgeons. Their capabilities extend to the automatic analysis of surgical videos, encompassing functions like indexing, documentation, and generating post-operative reports \cite{RBE}. The ever-increasing demand for such automatic systems has sparked machine-learning-based approaches to surgical video analysis. 

Cataract Surgery, renowned as the most commonly conducted ophthalmic surgical procedure and one of the most demanding surgeries worldwide, is a major operation where deep learning can be of great benefit. 
Cataract, characterized by the opacification of the eye's natural lens, is often attributed to aging and leads to impaired visual acuity, reduced brightness, visual distortion, double vision, and color perception degradation. Globally, cataracts stand as the primary cause of blindness \cite{VB2020}. Owing to the aging demographic and increased lifespans, the World Health Organization forecasts a surge in cataract-related blindness cases, estimating the number to reach 40 million by the year 2025 \cite{VB2020}.
This prevalent disease can be remedied through cataract surgery involving the substitution of the eye's natural lens with a synthetic counterpart known as an intraocular lens (IOL). Advancements in technology have driven the evolution of cataract surgery techniques. This evolution spans from intracapsular cataract extraction (ICCE) in the 1960s and 1970s to extracapsular cataract extraction (ECCE) in the 1980s and 1990s. Today, the primary method involves sutureless small-incision phacoemulsification surgery with an injectable intraocular lens (IOL) implantation\footnote{Throughout this paper, the term "Cataract Surgery" is synonymous with "Phacoemulsification Cataract Surgery."}. 
Due to the widespread occurrence of cataract surgery and its substantial influence on patients' quality of life, a significant focus has been directed towards the analysis of cataract surgery content using deep learning methodologies over the past decade. In particular, Surgical phase recognition and scene segmentation are joint building blocks in various applications related to cataract surgery video analysis \cite{RBE}. These applications include but are not limited to relevance detection \cite{LocalPhase}, relevance-based compression \cite{ghamsarian2020relevance}, irregularity detection \cite{LensID,sokolova2021automatic}, and surgical outcome prediction \cite{ghamsarian2023predicting}. The current public datasets for cataract surgery either provide annotations for a particular sub-task such as instrument recognition \cite{al2019cataracts}, scene and relevant anatomical structure segmentation \cite{CaDIS, ghamsarian2022deeppyramid, ReCal-Net, LocalPhase}, or offer small multi-task datasets targeting specific problems such as intraocular lens (IOL) irregularity detection \cite{LensID}. As a result of the lack of a comprehensive dataset, there exists a considerable gap in exploring deep-learning-based approaches and frameworks to enhance cataract surgery outcomes.
To facilitate the development of such systems and models, there is a compelling need for large-scale datasets that encompass multi-task annotations.

This paper introduces the largest cataract surgery video dataset, including 1000 videos of cataract surgery recorded in Klinikum Klagenfurt, Austria, between 2021 and 2023. We provide large-scale ground-truth annotations for the semantic segmentation of different instruments and relevant anatomical structures, as well as surgical phases. Besides, the dataset features two subsets for major irregularities in cataract surgery, which affect surgical workflow, including intraocular lens (IOL) rotation, and pupil contraction in cataract surgery. Together, these 1000 videos, annotated datasets, and irregularity subsets form a complete dataset to empower computer-assisted interventions (CAI) in cataract surgery.

\section*{Methods}

\subsection*{Cataract-1K Dataset Description}
The Cataract-1K dataset consists of 1000 videos of cataract surgeries performed in the eye clinic of Klinikum Klagenfurt from 2021 to 2023\footnote{The dataset will be publicly released in Synapse upon paper acceptance. For anonymizing purposes, however, our dataset is temporarily accessible via Figshare.}. From these videos, we provide surgical phase annotations for 56 regular videos and relevant anatomical plus instrument pixel-level annotations for 2256 frames out of 30 cataract surgery videos. Furthermore, we provide a small subset of surgeries with two major irregularities, including "pupil reaction" and "IOL rotation," to support further research on irregularity detection in cataract surgery. Except for the annotated videos and images, the remaining videos in the Cataract-1K dataset are encoded with a temporal resolution of 25 fps and a spatial resolution of $512 \times 324$. Besides, we assess the surgeons' skills by considering the cumulative count of their completed surgeries, which spans from 1,000 to over 40000 procedures in the Cataract-1K dataset. We delineate the challenges and annotation procedures for each subset in the following paragraphs.

\begin{figure*}[t!]
    \centering
    \begin{subfigure}[t]{0.19\textwidth}
    \centering
        \includegraphics[width=0.78\textwidth]{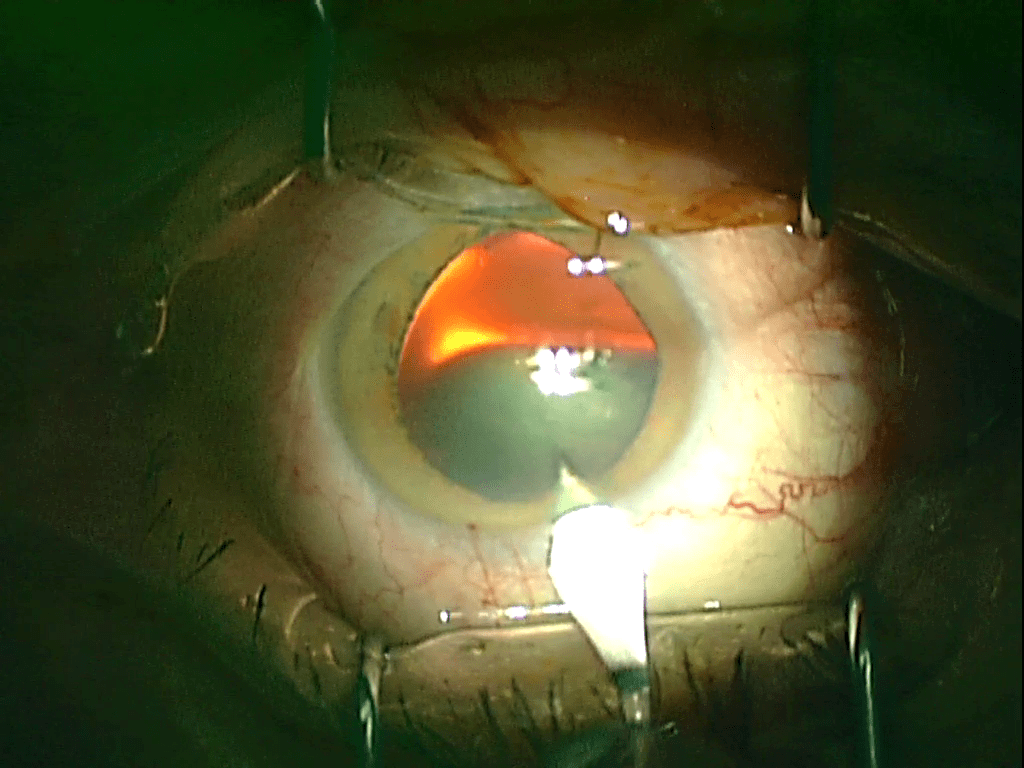}
        \caption*{Incision}    
    \end{subfigure}
    \begin{subfigure}[t]{0.19\textwidth}
    \centering
        \includegraphics[width=0.78\textwidth]{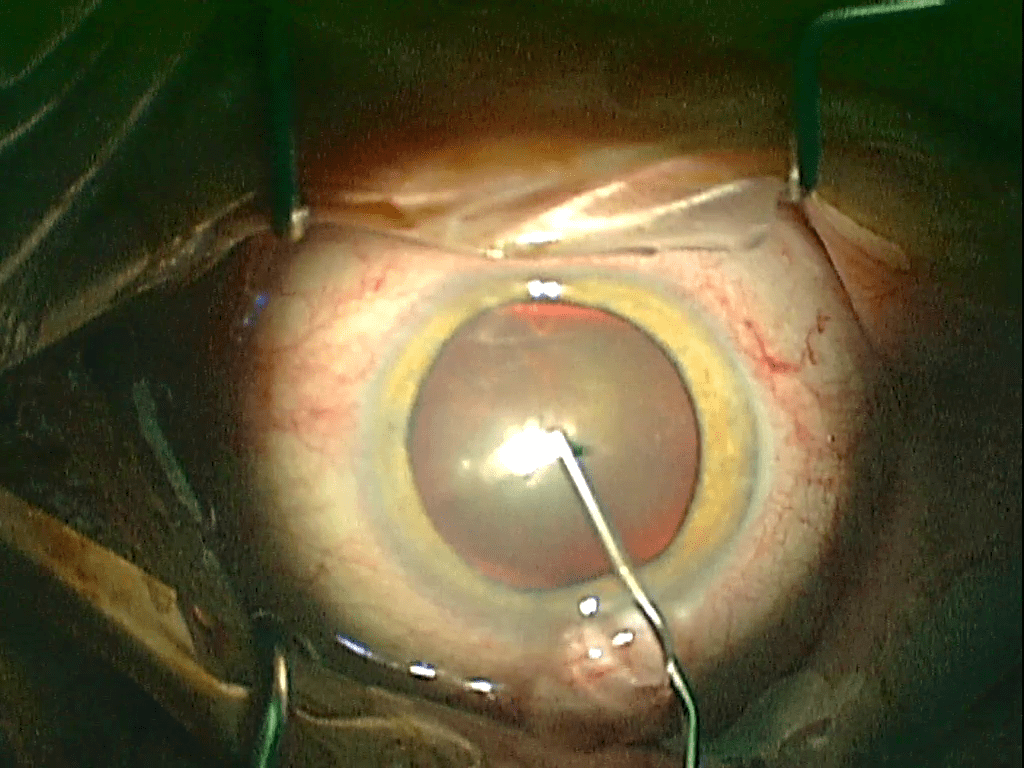}
        \caption*{Viscoelastic}    
    \end{subfigure}
    \begin{subfigure}[t]{0.19\textwidth}
    \centering
        \includegraphics[width=0.78\textwidth]{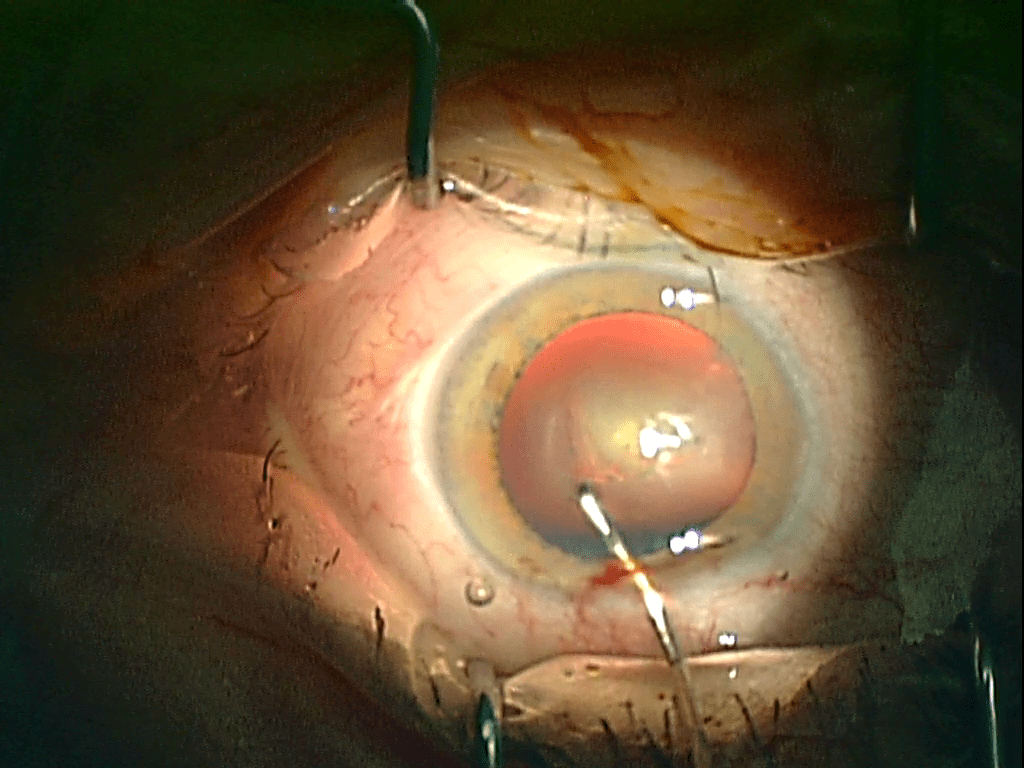}
        \caption*{Capsulorhexis}    
    \end{subfigure}
    \begin{subfigure}[t]{0.19\textwidth}
    \centering
        \includegraphics[width=0.78\textwidth]{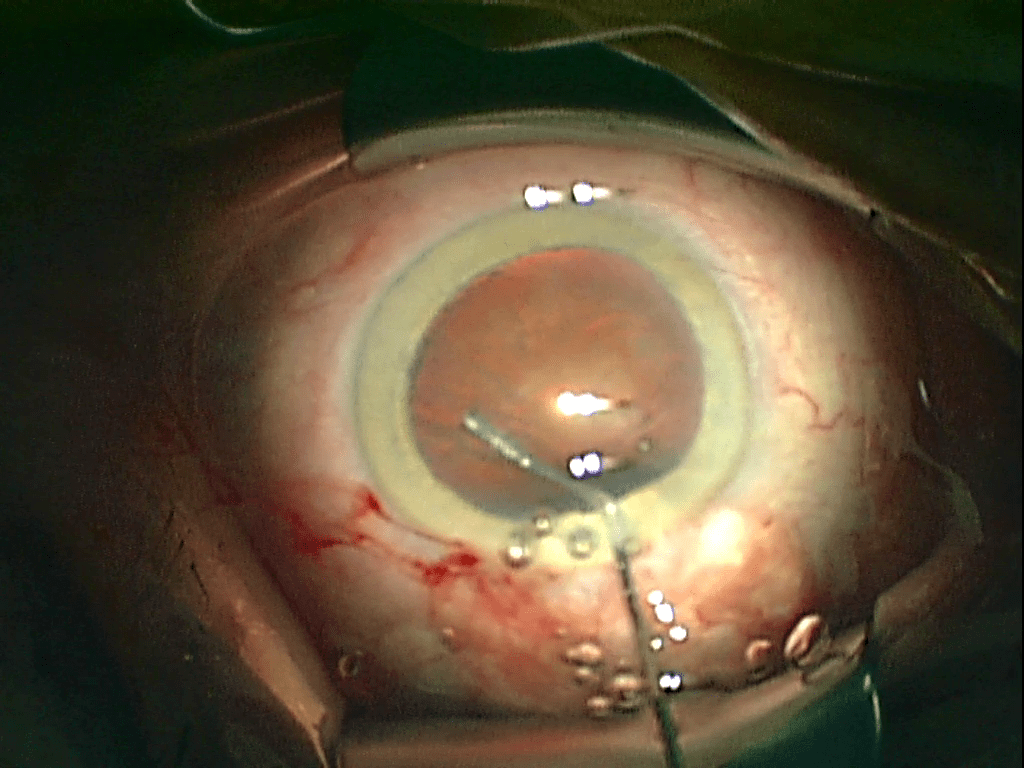}
        \caption*{Hydrodissection}    
    \end{subfigure}
    \begin{subfigure}[t]{0.19\textwidth}
    \centering
        \includegraphics[width=0.78\textwidth]{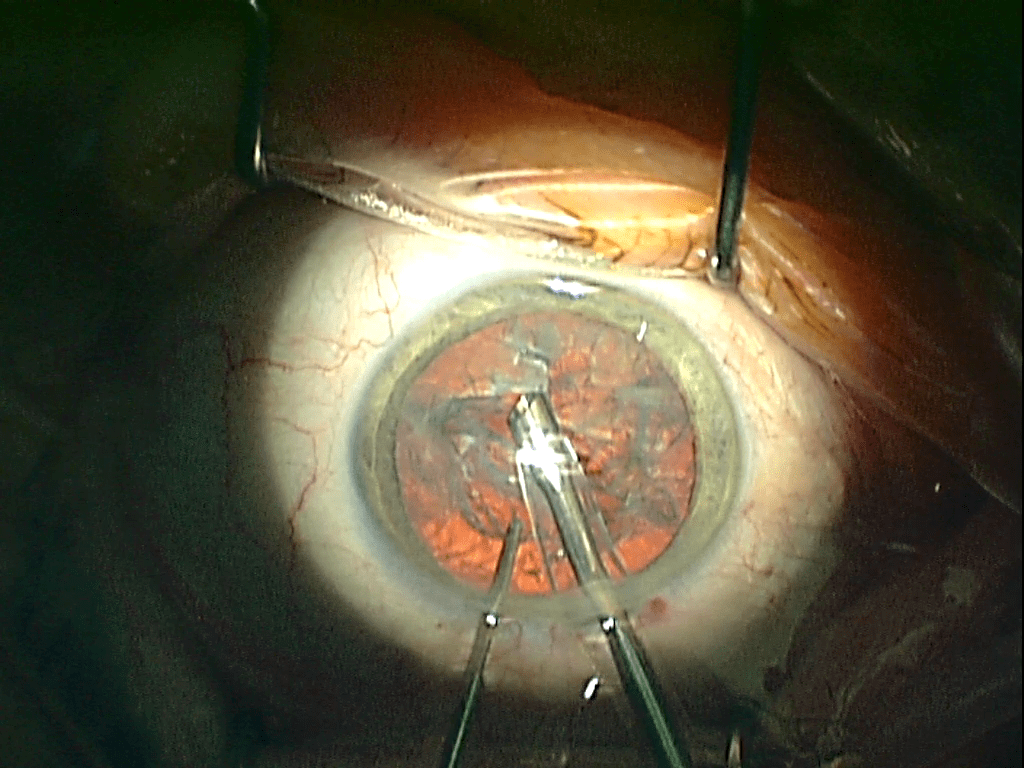}
        \caption*{Phacoemulsification}    
    \end{subfigure}\\
    \begin{subfigure}[t]{0.19\textwidth}
    \centering
        \includegraphics[width=0.78\textwidth]{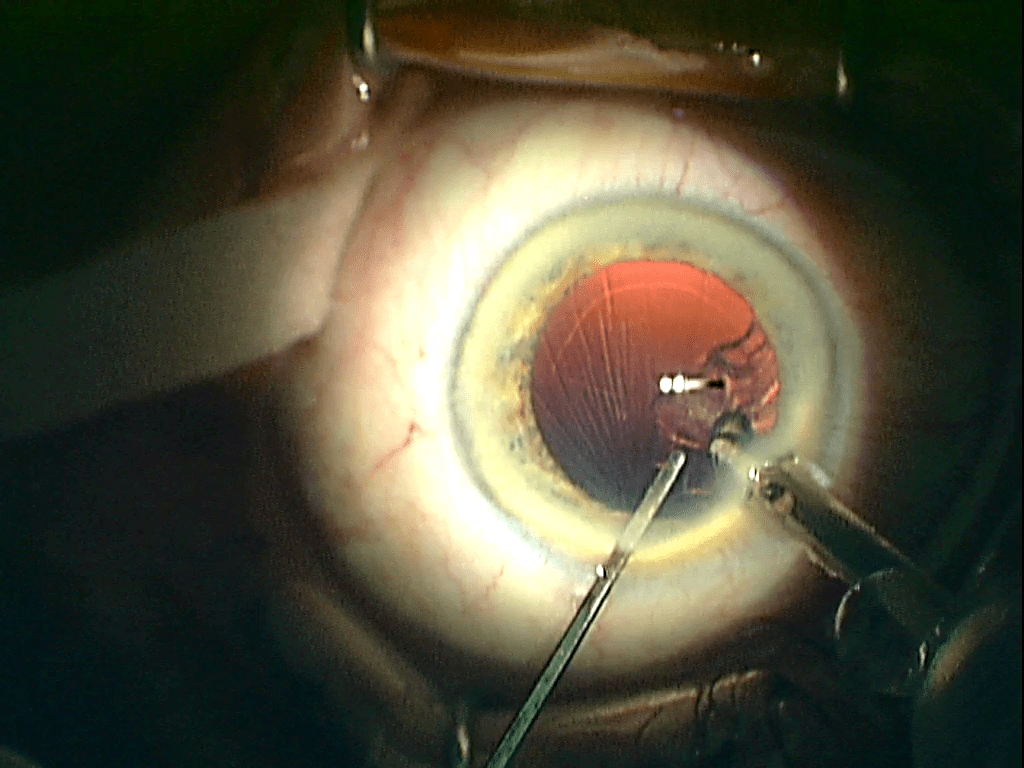}
        \caption*{Irrigation-Aspiration}    
    \end{subfigure}
    \begin{subfigure}[t]{0.19\textwidth}
    \centering
        \includegraphics[width=0.78\textwidth]{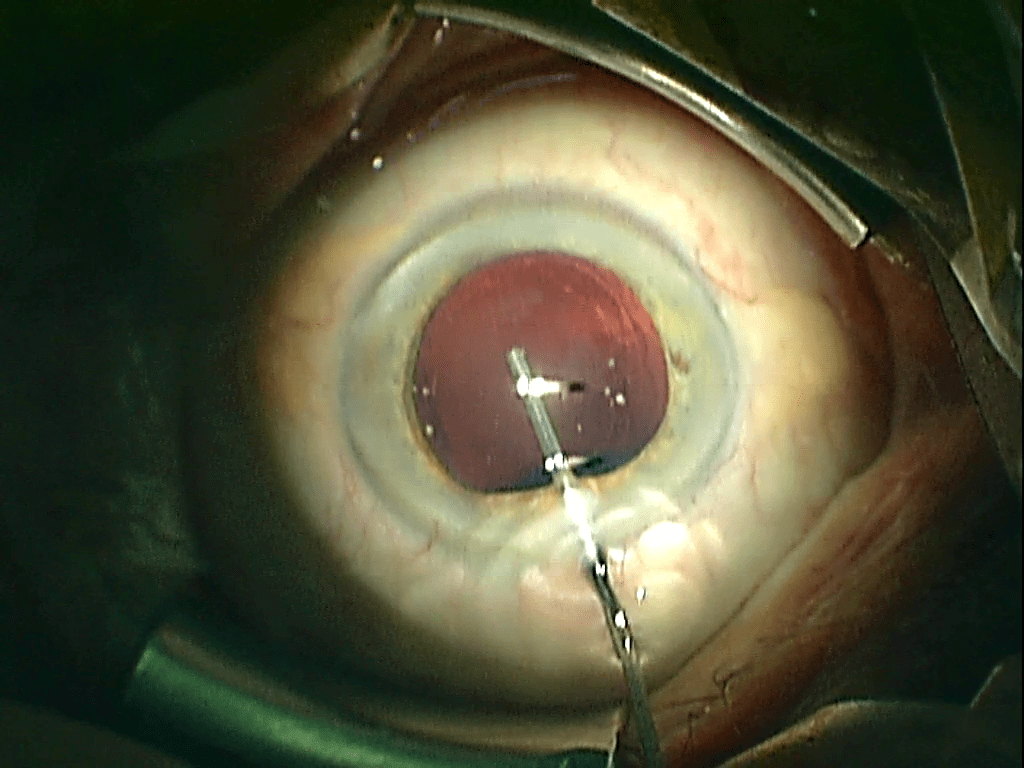}
        \caption*{Capsule Polishing}    
    \end{subfigure}
    \begin{subfigure}[t]{0.19\textwidth}
    \centering
        \includegraphics[width=0.78\textwidth]{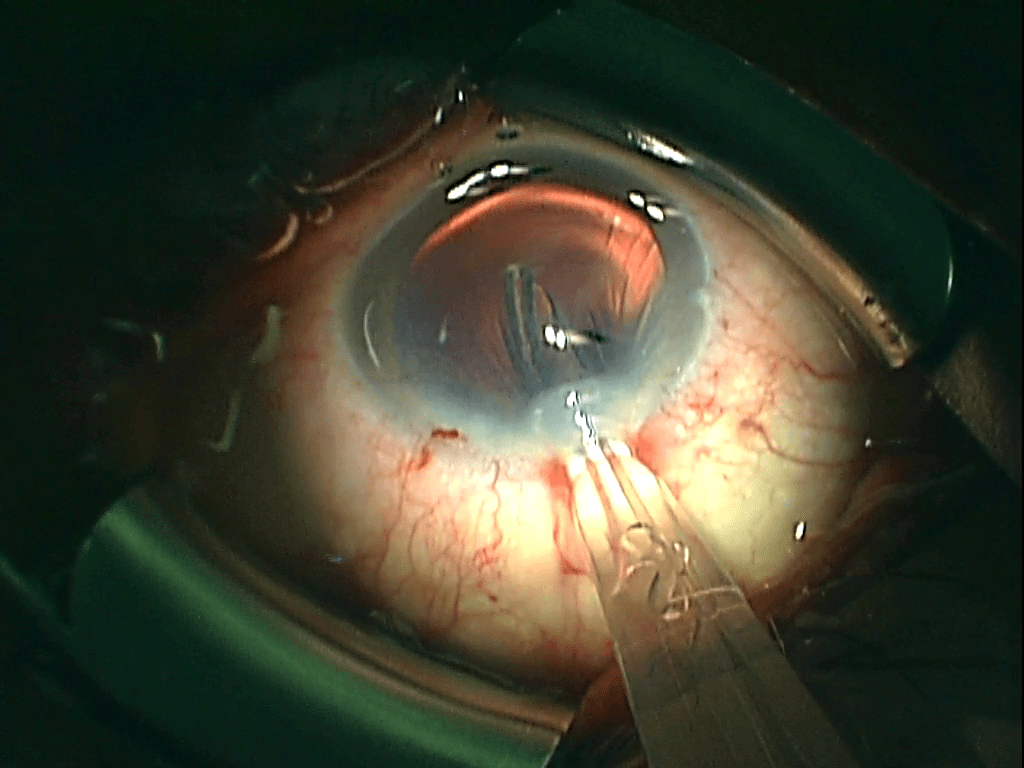}
        \caption*{Lens Implantation}    
    \end{subfigure}
    \begin{subfigure}[t]{0.19\textwidth}
    \centering
        \includegraphics[width=0.78\textwidth]{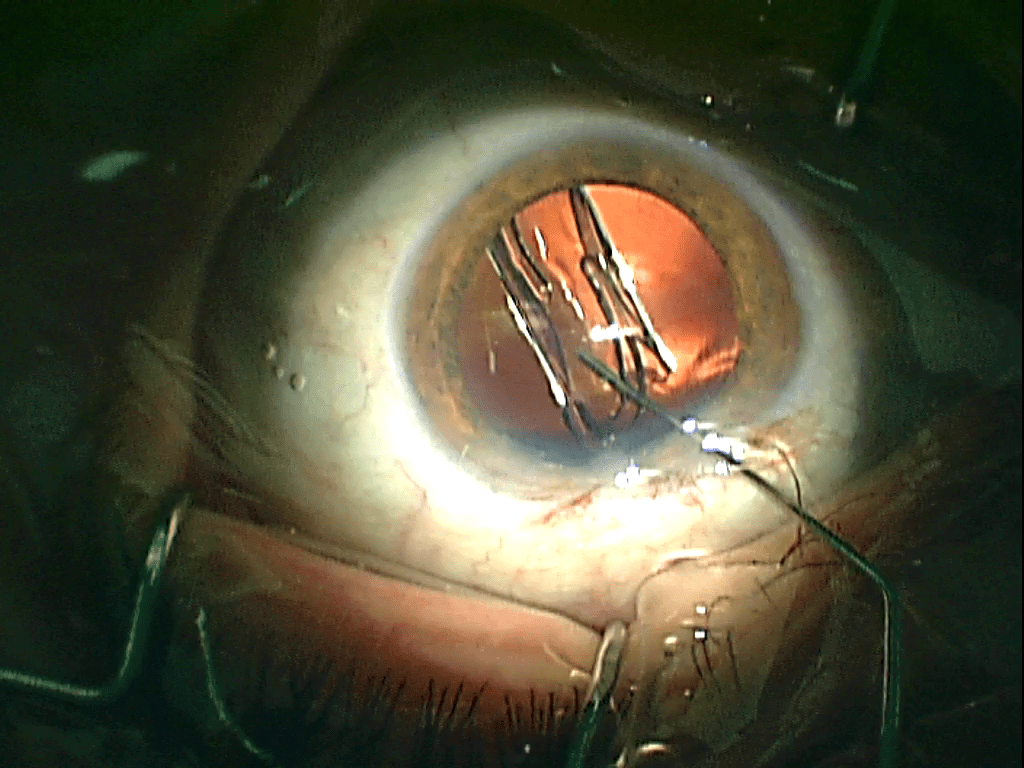}
        \caption*{Lens Positioning}    
    \end{subfigure}
    \centering
    \begin{subfigure}[t]{0.19\textwidth}
    \centering
        \includegraphics[width=0.78\textwidth]{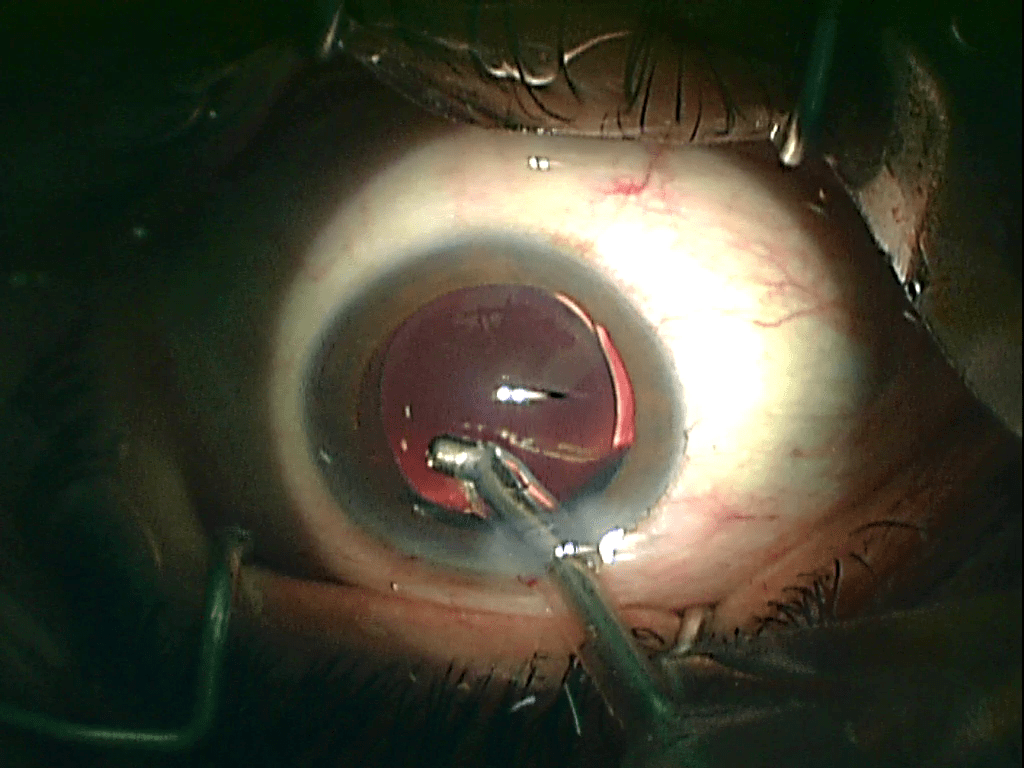}
        \caption*{Viscoelastic-Suction}    
    \end{subfigure}\\
    \begin{subfigure}[t]{0.25\textwidth}
    \centering
        \includegraphics[width=0.59\textwidth]{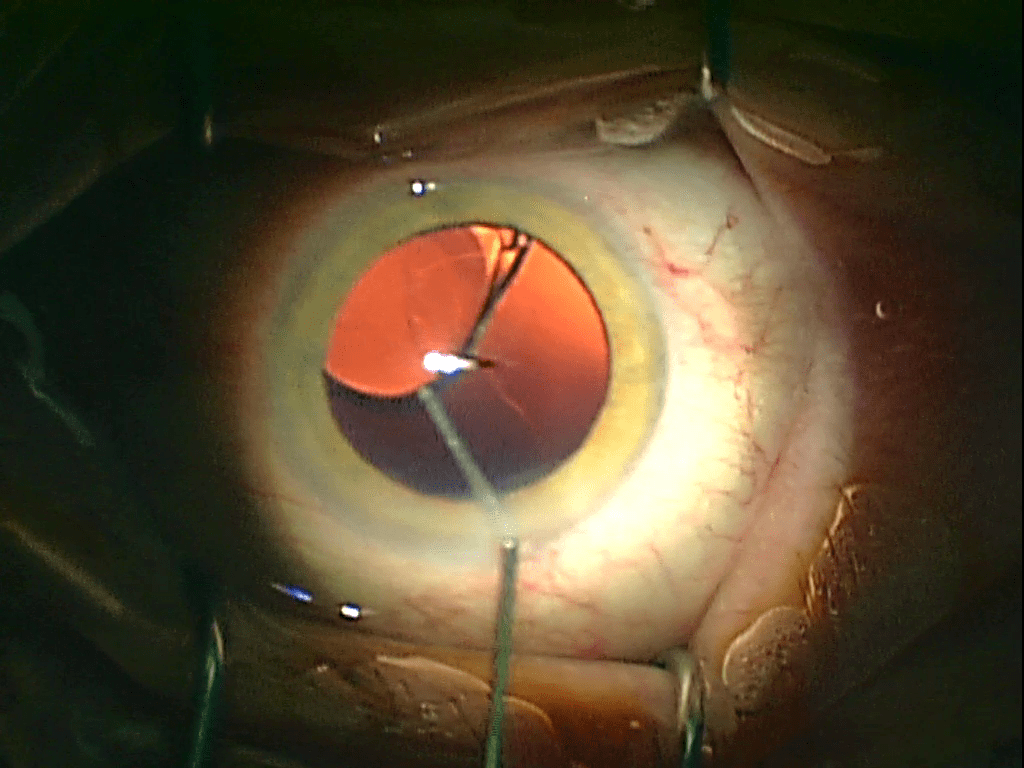}
        \caption*{Anterior-Chamber Flushing}    
    \end{subfigure}
    \begin{subfigure}[t]{0.25\textwidth}
    \centering
        \includegraphics[width=0.59\textwidth]{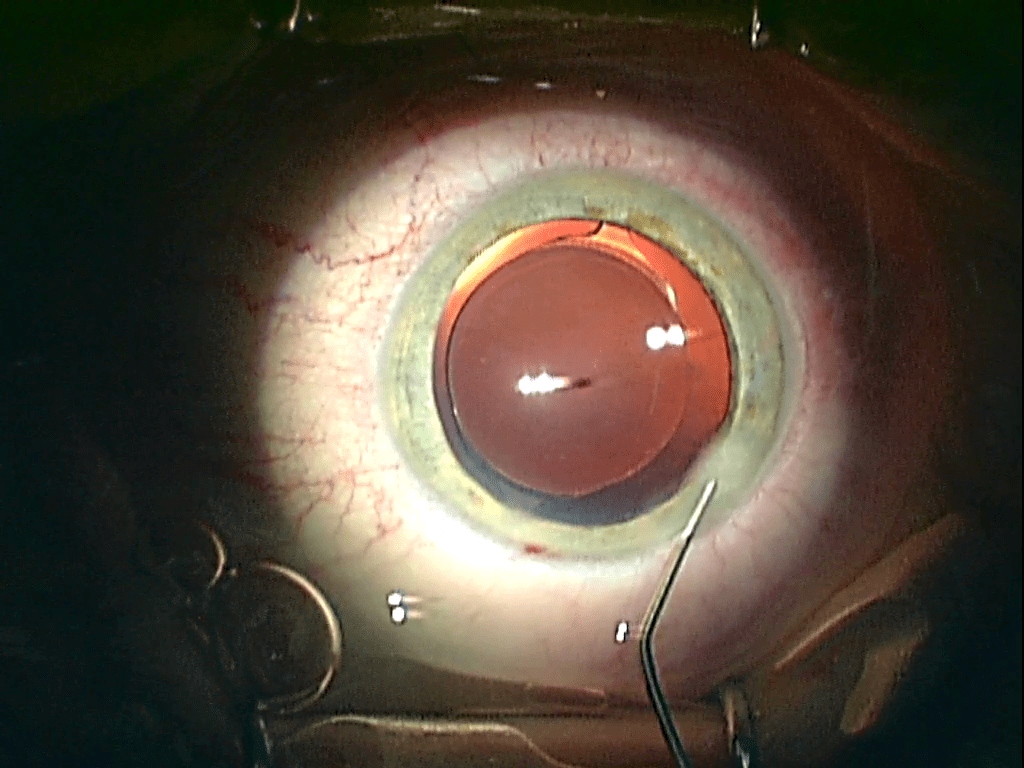}
        \caption*{Tonifying/Antibiotics}    
    \end{subfigure}
    \begin{subfigure}[t]{0.25\textwidth}
    \centering
        \includegraphics[width=0.59\textwidth]{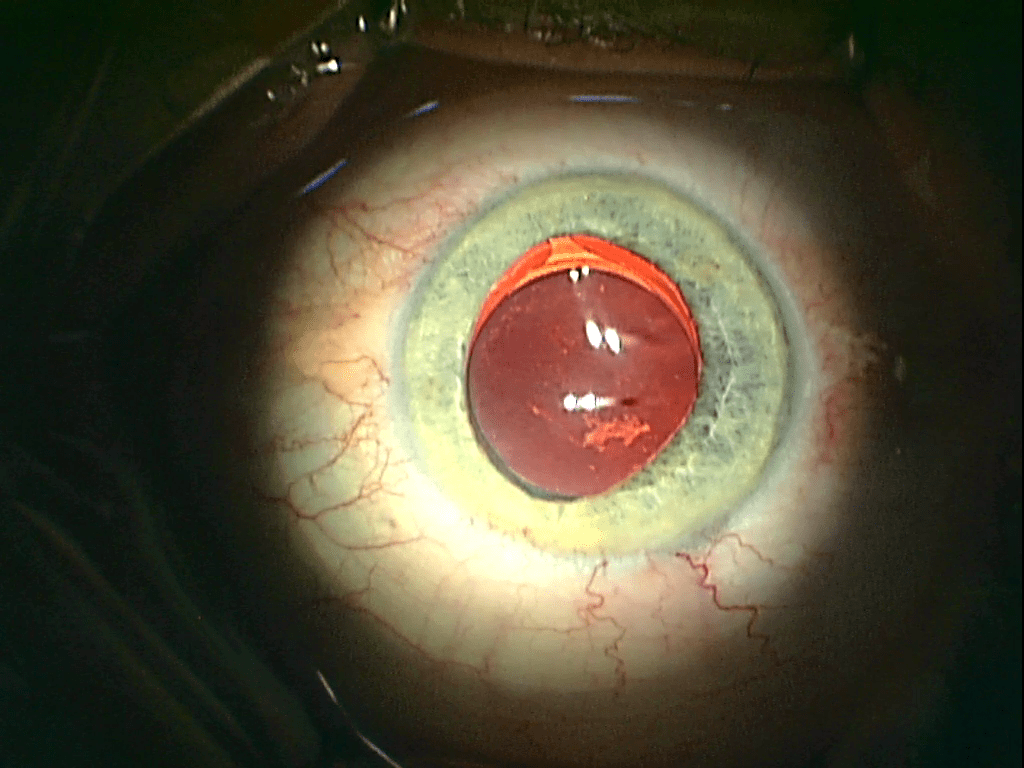}
        \caption*{Idle}    
    \end{subfigure}
    \\
    \caption{Sample frames from different phases in a regular cataract surgery.}
    \label{fig:phases-visualization}
\end{figure*}

\paragraph{Phase recognition dataset. }Crafting an approach to detect and classify significant phases within these videos, considering frame-by-frame temporal details, presents considerable challenges due to several factors:

\begin{table}[t!]
\centering
\caption{Visualizations of phase annotations for 56 normal cataract surgeries. The durations of the videos are different and normalized for better visualization.}
\label{tab:phase-annotations}
\resizebox{1\textwidth}{!}{%

\begin{tabular}{m{0.5cm}m{10cm}m{0.5cm}m{10cm}}
\specialrule{.12em}{.05em}{.05em}
Case & Phases & Case & Phases\\\midrule
1&\DP{0.11}{SkyBlue}\DP{0.10}{Maroon}\DP{0.20}{SkyBlue}\DP{0.13}{Maroon}\DP{0.20}{CarnationPink}\DP{0.07}{Maroon}\DP{1.16}{YellowGreen}\DP{0.07}{Maroon}\DP{0.89}{WildStrawberry}\DP{0.05}{Maroon}\DP{3.18}{ForestGreen}\DP{0.34}{Maroon}\DP{0.71}{Dandelion}\DP{0.12}{Maroon}\DP{0.19}{CarnationPink}\DP{0.10}{Maroon}\DP{0.23}{CarnationPink}\DP{0.10}{Maroon}\DP{0.43}{Purple}\DP{0.07}{Maroon}\DP{0.26}{Orchid}\DP{0.07}{Maroon}\DP{1.03}{RedOrange}\DP{0.09}{Maroon}\DP{0.10}{RoyalBlue}&2&\DP{0.13}{SkyBlue}\DP{0.10}{Maroon}\DP{0.27}{SkyBlue}\DP{0.11}{Maroon}\DP{0.15}{CarnationPink}\DP{0.05}{Maroon}\DP{0.99}{YellowGreen}\DP{0.08}{Maroon}\DP{0.40}{WildStrawberry}\DP{0.10}{Maroon}\DP{3.36}{ForestGreen}\DP{0.32}{Maroon}\DP{1.31}{Dandelion}\DP{0.07}{Maroon}\DP{0.19}{Tan}\DP{0.05}{Maroon}\DP{0.16}{CarnationPink}\DP{0.07}{Maroon}\DP{0.21}{Purple}\DP{0.05}{Maroon}\DP{0.22}{Orchid}\DP{0.05}{Maroon}\DP{0.83}{RedOrange}\DP{0.07}{Maroon}\DP{0.52}{RoyalBlue}\DP{0.06}{Maroon}\DP{0.08}{RoyalBlue}\\3&\DP{0.07}{SkyBlue}\DP{0.07}{Maroon}\DP{0.07}{SkyBlue}\DP{0.07}{Maroon}\DP{0.13}{CarnationPink}\DP{0.06}{Maroon}\DP{0.48}{YellowGreen}\DP{0.06}{Maroon}\DP{0.87}{WildStrawberry}\DP{0.08}{Maroon}\DP{4.54}{ForestGreen}\DP{0.15}{Maroon}\DP{0.82}{Dandelion}\DP{0.06}{Maroon}\DP{0.19}{Tan}\DP{0.01}{Maroon}\DP{0.09}{Tan}\DP{0.03}{Maroon}\DP{0.09}{CarnationPink}\DP{0.03}{Maroon}\DP{0.34}{Purple}\DP{0.04}{Maroon}\DP{0.16}{Orchid}\DP{0.04}{Maroon}\DP{0.67}{RedOrange}\DP{0.07}{Maroon}\DP{0.19}{Gray}\DP{0.05}{Maroon}\DP{0.27}{RedOrange}\DP{0.04}{Maroon}\DP{0.18}{RoyalBlue}&4&\DP{0.07}{SkyBlue}\DP{0.09}{Maroon}\DP{0.12}{SkyBlue}\DP{0.08}{Maroon}\DP{0.16}{CarnationPink}\DP{0.06}{Maroon}\DP{0.53}{YellowGreen}\DP{0.05}{Maroon}\DP{0.60}{WildStrawberry}\DP{0.06}{Maroon}\DP{3.95}{ForestGreen}\DP{0.19}{Maroon}\DP{0.95}{Dandelion}\DP{0.07}{Maroon}\DP{0.25}{Tan}\DP{0.07}{Maroon}\DP{0.12}{CarnationPink}\DP{0.13}{Maroon}\DP{0.32}{Purple}\DP{0.06}{Maroon}\DP{0.22}{Orchid}\DP{0.08}{Maroon}\DP{0.55}{RedOrange}\DP{0.06}{Maroon}\DP{0.42}{Gray}\DP{0.11}{Maroon}\DP{0.40}{RedOrange}\DP{0.08}{Maroon}\DP{0.17}{RoyalBlue}\\5&\DP{0.07}{SkyBlue}\DP{0.08}{Maroon}\DP{0.10}{SkyBlue}\DP{0.08}{Maroon}\DP{0.15}{CarnationPink}\DP{0.05}{Maroon}\DP{0.65}{YellowGreen}\DP{0.32}{Maroon}\DP{0.76}{WildStrawberry}\DP{0.05}{Maroon}\DP{3.12}{ForestGreen}\DP{0.17}{Maroon}\DP{0.93}{Dandelion}\DP{0.10}{Maroon}\DP{0.33}{Tan}\DP{0.07}{Maroon}\DP{0.13}{CarnationPink}\DP{0.12}{Maroon}\DP{0.49}{Purple}\DP{0.10}{Maroon}\DP{0.72}{Orchid}\DP{0.06}{Maroon}\DP{0.38}{RedOrange}\DP{0.06}{Maroon}\DP{0.14}{Gray}\DP{0.05}{Maroon}\DP{0.15}{RedOrange}\DP{0.03}{Maroon}\DP{0.28}{RoyalBlue}\DP{0.04}{Maroon}\DP{0.22}{RoyalBlue}&6&\DP{0.09}{SkyBlue}\DP{0.08}{Maroon}\DP{0.19}{SkyBlue}\DP{0.10}{Maroon}\DP{0.18}{CarnationPink}\DP{0.10}{Maroon}\DP{0.84}{YellowGreen}\DP{0.13}{Maroon}\DP{0.27}{WildStrawberry}\DP{0.02}{Maroon}\DP{0.66}{WildStrawberry}\DP{0.13}{Maroon}\DP{2.54}{ForestGreen}\DP{0.29}{Maroon}\DP{1.02}{Dandelion}\DP{0.11}{Maroon}\DP{0.28}{Tan}\DP{0.08}{Maroon}\DP{0.19}{CarnationPink}\DP{0.06}{Maroon}\DP{0.32}{Purple}\DP{0.06}{Maroon}\DP{0.36}{Orchid}\DP{0.08}{Maroon}\DP{0.73}{RedOrange}\DP{0.09}{Maroon}\DP{0.41}{Gray}\DP{0.09}{Maroon}\DP{0.25}{RedOrange}\DP{0.15}{Maroon}\DP{0.10}{RoyalBlue}\\7&\DP{0.10}{SkyBlue}\DP{0.08}{Maroon}\DP{0.13}{SkyBlue}\DP{0.06}{Maroon}\DP{0.12}{CarnationPink}\DP{0.07}{Maroon}\DP{0.65}{YellowGreen}\DP{0.08}{Maroon}\DP{0.65}{WildStrawberry}\DP{0.23}{Maroon}\DP{2.49}{ForestGreen}\DP{0.21}{Maroon}\DP{1.84}{Dandelion}\DP{0.06}{Maroon}\DP{0.10}{Tan}\DP{0.09}{Maroon}\DP{0.12}{CarnationPink}\DP{0.09}{Maroon}\DP{0.32}{Purple}\DP{0.07}{Maroon}\DP{0.12}{Orchid}\DP{0.05}{Maroon}\DP{0.37}{RedOrange}\DP{0.06}{Maroon}\DP{0.19}{Gray}\DP{0.06}{Maroon}\DP{0.27}{RedOrange}\DP{0.08}{Maroon}\DP{0.55}{RoyalBlue}\DP{0.07}{Maroon}\DP{0.15}{RedOrange}\DP{0.07}{Maroon}\DP{0.21}{RoyalBlue}\DP{0.06}{Maroon}\DP{0.15}{RoyalBlue}&8&\DP{0.12}{SkyBlue}\DP{0.07}{Maroon}\DP{0.22}{SkyBlue}\DP{0.32}{Maroon}\DP{0.21}{CarnationPink}\DP{0.06}{Maroon}\DP{0.82}{YellowGreen}\DP{0.06}{Maroon}\DP{0.40}{WildStrawberry}\DP{0.08}{Maroon}\DP{2.68}{ForestGreen}\DP{0.27}{Maroon}\DP{1.32}{Dandelion}\DP{0.06}{Maroon}\DP{0.14}{Tan}\DP{0.07}{Maroon}\DP{0.19}{CarnationPink}\DP{0.05}{Maroon}\DP{0.41}{Purple}\DP{0.07}{Maroon}\DP{0.13}{Orchid}\DP{0.05}{Maroon}\DP{1.43}{RedOrange}\DP{0.08}{Maroon}\DP{0.28}{RoyalBlue}\DP{0.07}{Maroon}\DP{0.22}{RoyalBlue}\DP{0.08}{Maroon}\DP{0.06}{RoyalBlue}\\9&\DP{0.06}{SkyBlue}\DP{0.06}{Maroon}\DP{0.13}{SkyBlue}\DP{0.06}{Maroon}\DP{0.08}{CarnationPink}\DP{0.05}{Maroon}\DP{0.45}{YellowGreen}\DP{0.07}{Maroon}\DP{0.51}{WildStrawberry}\DP{0.04}{Maroon}\DP{0.43}{WildStrawberry}\DP{0.08}{Maroon}\DP{3.07}{ForestGreen}\DP{0.19}{Maroon}\DP{0.68}{Dandelion}\DP{0.09}{Maroon}\DP{0.24}{Tan}\DP{0.05}{Maroon}\DP{0.37}{Tan}\DP{0.01}{Maroon}\DP{0.16}{Tan}\DP{0.04}{Maroon}\DP{0.10}{CarnationPink}\DP{0.07}{Maroon}\DP{0.60}{Purple}\DP{0.05}{Maroon}\DP{0.26}{Orchid}\DP{0.08}{Maroon}\DP{0.51}{RedOrange}\DP{0.05}{Maroon}\DP{0.27}{Gray}\DP{0.05}{Maroon}\DP{0.42}{RedOrange}\DP{0.07}{Maroon}\DP{0.28}{RoyalBlue}\DP{0.05}{Maroon}\DP{0.22}{RoyalBlue}&10&\DP{0.11}{SkyBlue}\DP{0.07}{Maroon}\DP{0.09}{SkyBlue}\DP{0.21}{Maroon}\DP{0.16}{CarnationPink}\DP{0.04}{Maroon}\DP{0.39}{YellowGreen}\DP{0.05}{Maroon}\DP{0.61}{WildStrawberry}\DP{0.04}{Maroon}\DP{2.57}{ForestGreen}\DP{0.12}{Maroon}\DP{1.25}{Dandelion}\DP{0.07}{Maroon}\DP{0.55}{Tan}\DP{0.14}{Maroon}\DP{0.12}{CarnationPink}\DP{0.04}{Maroon}\DP{0.61}{Purple}\DP{0.05}{Maroon}\DP{0.47}{Orchid}\DP{0.05}{Maroon}\DP{0.64}{RedOrange}\DP{0.03}{Maroon}\DP{0.22}{Gray}\DP{0.03}{Maroon}\DP{0.17}{Gray}\DP{0.05}{Maroon}\DP{0.40}{RedOrange}\DP{0.04}{Maroon}\DP{0.27}{RoyalBlue}\DP{0.03}{Maroon}\DP{0.29}{RoyalBlue}\\11&\DP{0.10}{SkyBlue}\DP{0.06}{Maroon}\DP{0.09}{SkyBlue}\DP{0.06}{Maroon}\DP{0.11}{CarnationPink}\DP{0.04}{Maroon}\DP{0.80}{YellowGreen}\DP{0.04}{Maroon}\DP{0.35}{WildStrawberry}\DP{0.10}{Maroon}\DP{0.50}{WildStrawberry}\DP{0.05}{Maroon}\DP{3.34}{ForestGreen}\DP{0.20}{Maroon}\DP{0.96}{Dandelion}\DP{0.06}{Maroon}\DP{0.12}{Tan}\DP{0.02}{Maroon}\DP{0.16}{Tan}\DP{0.04}{Maroon}\DP{0.12}{CarnationPink}\DP{0.04}{Maroon}\DP{0.43}{Purple}\DP{0.04}{Maroon}\DP{0.45}{Orchid}\DP{0.05}{Maroon}\DP{0.48}{RedOrange}\DP{0.09}{Maroon}\DP{0.27}{Gray}\DP{0.03}{Maroon}\DP{0.28}{RedOrange}\DP{0.03}{Maroon}\DP{0.27}{RoyalBlue}\DP{0.03}{Maroon}\DP{0.09}{RoyalBlue}\DP{0.00}{Maroon}\DP{0.11}{RoyalBlue}&12&\DP{0.10}{SkyBlue}\DP{0.13}{Maroon}\DP{0.13}{SkyBlue}\DP{0.13}{Maroon}\DP{0.20}{CarnationPink}\DP{0.40}{Maroon}\DP{0.93}{YellowGreen}\DP{0.07}{Maroon}\DP{0.71}{WildStrawberry}\DP{0.11}{Maroon}\DP{2.19}{ForestGreen}\DP{0.28}{Maroon}\DP{1.74}{Dandelion}\DP{0.10}{Maroon}\DP{0.20}{Tan}\DP{0.10}{Maroon}\DP{0.20}{CarnationPink}\DP{0.06}{Maroon}\DP{0.16}{Purple}\DP{0.04}{Maroon}\DP{0.32}{Orchid}\DP{0.05}{Maroon}\DP{0.94}{RedOrange}\DP{0.09}{Maroon}\DP{0.62}{RoyalBlue}\\13&\DP{0.10}{SkyBlue}\DP{0.10}{Maroon}\DP{0.11}{SkyBlue}\DP{0.09}{Maroon}\DP{0.16}{CarnationPink}\DP{0.09}{Maroon}\DP{0.94}{YellowGreen}\DP{0.06}{Maroon}\DP{0.38}{WildStrawberry}\DP{0.11}{Maroon}\DP{3.96}{ForestGreen}\DP{0.20}{Maroon}\DP{0.96}{Dandelion}\DP{0.16}{Maroon}\DP{0.08}{Tan}\DP{0.05}{Maroon}\DP{0.15}{CarnationPink}\DP{0.06}{Maroon}\DP{0.28}{Purple}\DP{0.04}{Maroon}\DP{0.12}{Orchid}\DP{0.05}{Maroon}\DP{0.27}{RedOrange}\DP{0.07}{Maroon}\DP{0.45}{Gray}\DP{0.04}{Maroon}\DP{0.22}{RedOrange}\DP{0.05}{Maroon}\DP{0.21}{Gray}\DP{0.02}{Maroon}\DP{0.23}{RoyalBlue}\DP{0.08}{Maroon}\DP{0.10}{RoyalBlue}&14&\DP{0.12}{SkyBlue}\DP{0.20}{Maroon}\DP{0.13}{SkyBlue}\DP{0.25}{Maroon}\DP{0.19}{CarnationPink}\DP{0.05}{Maroon}\DP{1.01}{YellowGreen}\DP{0.06}{Maroon}\DP{0.89}{WildStrawberry}\DP{0.09}{Maroon}\DP{2.19}{ForestGreen}\DP{0.18}{Maroon}\DP{1.57}{Dandelion}\DP{0.09}{Maroon}\DP{0.14}{Tan}\DP{0.05}{Maroon}\DP{0.26}{CarnationPink}\DP{0.05}{Maroon}\DP{0.34}{Purple}\DP{0.04}{Maroon}\DP{0.15}{Orchid}\DP{0.04}{Maroon}\DP{0.53}{RedOrange}\DP{0.07}{Maroon}\DP{0.30}{Gray}\DP{0.04}{Maroon}\DP{0.31}{RedOrange}\DP{0.07}{Maroon}\DP{0.34}{RoyalBlue}\DP{0.09}{Maroon}\DP{0.14}{RoyalBlue}\\15&\DP{0.08}{SkyBlue}\DP{0.09}{Maroon}\DP{0.12}{SkyBlue}\DP{0.08}{Maroon}\DP{0.13}{CarnationPink}\DP{0.06}{Maroon}\DP{0.72}{YellowGreen}\DP{0.57}{Maroon}\DP{0.53}{WildStrawberry}\DP{0.09}{Maroon}\DP{2.17}{ForestGreen}\DP{0.21}{Maroon}\DP{1.93}{Dandelion}\DP{0.07}{Maroon}\DP{0.30}{Tan}\DP{0.07}{Maroon}\DP{0.18}{CarnationPink}\DP{0.06}{Maroon}\DP{0.23}{Purple}\DP{0.03}{Maroon}\DP{0.38}{Orchid}\DP{0.04}{Maroon}\DP{0.57}{RedOrange}\DP{0.06}{Maroon}\DP{0.20}{Gray}\DP{0.04}{Maroon}\DP{0.36}{RedOrange}\DP{0.05}{Maroon}\DP{0.34}{RoyalBlue}\DP{0.09}{Maroon}\DP{0.14}{RoyalBlue}&16&\DP{0.04}{SkyBlue}\DP{0.10}{Maroon}\DP{0.10}{SkyBlue}\DP{0.31}{Maroon}\DP{0.18}{CarnationPink}\DP{0.18}{Maroon}\DP{0.91}{YellowGreen}\DP{0.06}{Maroon}\DP{0.55}{WildStrawberry}\DP{0.12}{Maroon}\DP{2.96}{ForestGreen}\DP{0.32}{Maroon}\DP{1.34}{Dandelion}\DP{0.10}{Maroon}\DP{0.12}{Tan}\DP{0.08}{Maroon}\DP{0.13}{CarnationPink}\DP{0.07}{Maroon}\DP{0.24}{Purple}\DP{0.05}{Maroon}\DP{0.13}{Orchid}\DP{0.06}{Maroon}\DP{0.51}{RedOrange}\DP{0.08}{Maroon}\DP{0.24}{Gray}\DP{0.12}{Maroon}\DP{0.26}{RedOrange}\DP{0.07}{Maroon}\DP{0.35}{RoyalBlue}\DP{0.06}{Maroon}\DP{0.16}{RoyalBlue}\\17&\DP{0.08}{SkyBlue}\DP{0.10}{Maroon}\DP{0.12}{SkyBlue}\DP{0.09}{Maroon}\DP{0.24}{CarnationPink}\DP{0.06}{Maroon}\DP{1.36}{YellowGreen}\DP{0.07}{Maroon}\DP{0.37}{WildStrawberry}\DP{0.14}{Maroon}\DP{2.69}{ForestGreen}\DP{0.37}{Maroon}\DP{1.67}{Dandelion}\DP{0.12}{Maroon}\DP{0.14}{Tan}\DP{0.09}{Maroon}\DP{0.14}{CarnationPink}\DP{0.11}{Maroon}\DP{0.23}{Purple}\DP{0.06}{Maroon}\DP{0.12}{Orchid}\DP{0.07}{Maroon}\DP{0.36}{RedOrange}\DP{0.08}{Maroon}\DP{0.28}{Gray}\DP{0.04}{Maroon}\DP{0.24}{RedOrange}\DP{0.06}{Maroon}\DP{0.31}{RoyalBlue}\DP{0.07}{Maroon}\DP{0.11}{RoyalBlue}&18&\DP{0.10}{SkyBlue}\DP{0.09}{Maroon}\DP{0.11}{SkyBlue}\DP{0.07}{Maroon}\DP{0.13}{CarnationPink}\DP{0.08}{Maroon}\DP{0.68}{YellowGreen}\DP{0.08}{Maroon}\DP{0.28}{WildStrawberry}\DP{0.17}{Maroon}\DP{3.37}{ForestGreen}\DP{0.25}{Maroon}\DP{2.36}{Dandelion}\DP{0.10}{Maroon}\DP{0.09}{Tan}\DP{0.07}{Maroon}\DP{0.14}{CarnationPink}\DP{0.05}{Maroon}\DP{0.21}{Purple}\DP{0.06}{Maroon}\DP{0.09}{Orchid}\DP{0.05}{Maroon}\DP{0.29}{RedOrange}\DP{0.08}{Maroon}\DP{0.24}{Gray}\DP{0.04}{Maroon}\DP{0.18}{RedOrange}\DP{0.06}{Maroon}\DP{0.12}{RoyalBlue}\DP{0.01}{Maroon}\DP{0.21}{RoyalBlue}\DP{0.07}{Maroon}\DP{0.09}{RoyalBlue}\\19&\DP{0.12}{SkyBlue}\DP{0.10}{Maroon}\DP{0.10}{SkyBlue}\DP{0.09}{Maroon}\DP{0.19}{CarnationPink}\DP{0.07}{Maroon}\DP{0.86}{YellowGreen}\DP{0.06}{Maroon}\DP{0.38}{WildStrawberry}\DP{0.14}{Maroon}\DP{2.82}{ForestGreen}\DP{0.24}{Maroon}\DP{1.36}{Dandelion}\DP{0.11}{Maroon}\DP{0.25}{Tan}\DP{0.05}{Maroon}\DP{0.16}{CarnationPink}\DP{0.05}{Maroon}\DP{0.44}{Purple}\DP{0.04}{Maroon}\DP{0.19}{Orchid}\DP{0.05}{Maroon}\DP{0.54}{RedOrange}\DP{0.09}{Maroon}\DP{0.38}{Gray}\DP{0.04}{Maroon}\DP{0.40}{RedOrange}\DP{0.06}{Maroon}\DP{0.42}{RoyalBlue}\DP{0.07}{Maroon}\DP{0.11}{RoyalBlue}&20&\DP{0.09}{SkyBlue}\DP{0.18}{Maroon}\DP{0.10}{SkyBlue}\DP{0.10}{Maroon}\DP{0.24}{CarnationPink}\DP{0.05}{Maroon}\DP{0.70}{YellowGreen}\DP{0.07}{Maroon}\DP{0.71}{WildStrawberry}\DP{0.16}{Maroon}\DP{2.49}{ForestGreen}\DP{0.23}{Maroon}\DP{1.09}{Dandelion}\DP{0.17}{Maroon}\DP{0.11}{Tan}\DP{0.12}{Maroon}\DP{0.15}{CarnationPink}\DP{0.22}{Maroon}\DP{0.38}{Purple}\DP{0.07}{Maroon}\DP{0.18}{Orchid}\DP{0.05}{Maroon}\DP{0.36}{RedOrange}\DP{0.15}{Maroon}\DP{0.36}{Gray}\DP{0.03}{Maroon}\DP{0.32}{RedOrange}\DP{0.11}{Maroon}\DP{0.32}{RoyalBlue}\DP{0.04}{Maroon}\DP{0.22}{RoyalBlue}\DP{0.05}{Maroon}\DP{0.11}{RedOrange}\DP{0.09}{Maroon}\DP{0.16}{RoyalBlue}\\21&\DP{0.14}{SkyBlue}\DP{0.11}{Maroon}\DP{0.08}{SkyBlue}\DP{0.11}{Maroon}\DP{0.14}{CarnationPink}\DP{0.08}{Maroon}\DP{0.54}{YellowGreen}\DP{0.08}{Maroon}\DP{0.37}{WildStrawberry}\DP{0.12}{Maroon}\DP{2.03}{ForestGreen}\DP{0.31}{Maroon}\DP{2.31}{Dandelion}\DP{0.09}{Maroon}\DP{0.19}{Tan}\DP{0.10}{Maroon}\DP{0.12}{CarnationPink}\DP{0.11}{Maroon}\DP{0.63}{Purple}\DP{0.07}{Maroon}\DP{0.11}{Orchid}\DP{0.12}{Maroon}\DP{1.17}{RedOrange}\DP{0.07}{Maroon}\DP{0.61}{RoyalBlue}\DP{0.07}{Maroon}\DP{0.13}{RoyalBlue}&22&\DP{0.11}{SkyBlue}\DP{0.14}{Maroon}\DP{0.16}{SkyBlue}\DP{0.10}{Maroon}\DP{0.13}{CarnationPink}\DP{0.09}{Maroon}\DP{1.10}{YellowGreen}\DP{0.10}{Maroon}\DP{0.73}{WildStrawberry}\DP{0.11}{Maroon}\DP{2.47}{ForestGreen}\DP{0.28}{Maroon}\DP{0.91}{Dandelion}\DP{0.16}{Maroon}\DP{0.17}{Tan}\DP{0.13}{Maroon}\DP{0.18}{CarnationPink}\DP{0.16}{Maroon}\DP{0.44}{Purple}\DP{0.20}{Maroon}\DP{0.48}{Orchid}\DP{0.07}{Maroon}\DP{0.21}{RedOrange}\DP{0.11}{Maroon}\DP{0.38}{Gray}\DP{0.07}{Maroon}\DP{0.22}{RedOrange}\DP{0.09}{Maroon}\DP{0.51}{RoyalBlue}\\23&\DP{0.11}{SkyBlue}\DP{0.08}{Maroon}\DP{0.12}{SkyBlue}\DP{0.09}{Maroon}\DP{0.19}{CarnationPink}\DP{0.05}{Maroon}\DP{1.16}{YellowGreen}\DP{0.10}{Maroon}\DP{0.58}{WildStrawberry}\DP{0.24}{Maroon}\DP{2.83}{ForestGreen}\DP{0.23}{Maroon}\DP{0.96}{Dandelion}\DP{0.08}{Maroon}\DP{0.44}{Tan}\DP{0.05}{Maroon}\DP{0.19}{CarnationPink}\DP{0.06}{Maroon}\DP{0.49}{Purple}\DP{0.05}{Maroon}\DP{0.30}{Orchid}\DP{0.06}{Maroon}\DP{0.25}{RedOrange}\DP{0.08}{Maroon}\DP{0.43}{Gray}\DP{0.06}{Maroon}\DP{0.32}{RedOrange}\DP{0.08}{Maroon}\DP{0.33}{RoyalBlue}&24&\DP{0.09}{SkyBlue}\DP{0.06}{Maroon}\DP{0.14}{SkyBlue}\DP{0.08}{Maroon}\DP{0.10}{CarnationPink}\DP{0.05}{Maroon}\DP{0.89}{YellowGreen}\DP{0.06}{Maroon}\DP{0.43}{WildStrawberry}\DP{0.12}{Maroon}\DP{2.21}{ForestGreen}\DP{0.29}{Maroon}\DP{1.20}{Dandelion}\DP{0.09}{Maroon}\DP{0.25}{Tan}\DP{0.12}{Maroon}\DP{0.17}{Tan}\DP{0.05}{Maroon}\DP{0.14}{CarnationPink}\DP{0.06}{Maroon}\DP{0.42}{Purple}\DP{0.03}{Maroon}\DP{0.25}{Orchid}\DP{0.05}{Maroon}\DP{0.96}{RedOrange}\DP{0.05}{Maroon}\DP{0.43}{Gray}\DP{0.05}{Maroon}\DP{0.36}{RedOrange}\DP{0.11}{Maroon}\DP{0.67}{RoyalBlue}\\25&\DP{0.09}{SkyBlue}\DP{0.06}{Maroon}\DP{0.09}{SkyBlue}\DP{0.10}{Maroon}\DP{0.12}{CarnationPink}\DP{0.06}{Maroon}\DP{0.76}{YellowGreen}\DP{0.08}{Maroon}\DP{0.35}{WildStrawberry}\DP{0.24}{Maroon}\DP{1.72}{ForestGreen}\DP{0.38}{Maroon}\DP{1.31}{Dandelion}\DP{0.03}{Maroon}\DP{0.34}{Tan}\DP{0.06}{Maroon}\DP{0.42}{Tan}\DP{0.04}{Maroon}\DP{0.14}{Tan}\DP{0.06}{Maroon}\DP{0.14}{CarnationPink}\DP{0.05}{Maroon}\DP{0.30}{Purple}\DP{0.03}{Maroon}\DP{0.16}{Orchid}\DP{0.04}{Maroon}\DP{0.00}{RedOrange}\DP{0.51}{RedOrange}\DP{0.01}{Maroon}\DP{0.31}{RedOrange}\DP{0.10}{Maroon}\DP{1.16}{RoyalBlue}\DP{0.10}{Maroon}\DP{0.20}{RedOrange}\DP{0.11}{Maroon}\DP{0.17}{RoyalBlue}\DP{0.08}{Maroon}\DP{0.08}{RoyalBlue}&26&\DP{0.09}{SkyBlue}\DP{0.07}{Maroon}\DP{0.10}{SkyBlue}\DP{0.07}{Maroon}\DP{0.16}{CarnationPink}\DP{0.07}{Maroon}\DP{0.66}{YellowGreen}\DP{0.06}{Maroon}\DP{0.63}{WildStrawberry}\DP{0.29}{Maroon}\DP{2.97}{ForestGreen}\DP{0.24}{Maroon}\DP{1.39}{Dandelion}\DP{0.08}{Maroon}\DP{0.58}{Tan}\DP{0.06}{Maroon}\DP{0.36}{Tan}\DP{0.03}{Maroon}\DP{0.12}{Tan}\DP{0.07}{Maroon}\DP{0.16}{CarnationPink}\DP{0.05}{Maroon}\DP{0.34}{Purple}\DP{0.05}{Maroon}\DP{0.30}{Orchid}\DP{0.06}{Maroon}\DP{0.31}{RedOrange}\DP{0.05}{Maroon}\DP{0.49}{Gray}\DP{0.00}{Maroon}\DP{0.10}{RoyalBlue}\\27&\DP{0.19}{SkyBlue}\DP{0.06}{Maroon}\DP{0.10}{SkyBlue}\DP{0.04}{Maroon}\DP{0.16}{CarnationPink}\DP{0.14}{Maroon}\DP{0.99}{YellowGreen}\DP{0.03}{Maroon}\DP{0.64}{WildStrawberry}\DP{0.04}{Maroon}\DP{0.15}{WildStrawberry}\DP{0.07}{Maroon}\DP{2.74}{ForestGreen}\DP{0.13}{Maroon}\DP{1.80}{Dandelion}\DP{0.04}{Maroon}\DP{0.26}{Tan}\DP{0.07}{Maroon}\DP{0.27}{Tan}\DP{0.08}{Maroon}\DP{0.16}{Tan}\DP{0.11}{Maroon}\DP{0.14}{CarnationPink}\DP{0.03}{Maroon}\DP{0.16}{Purple}\DP{0.04}{Maroon}\DP{0.24}{Orchid}\DP{0.06}{Maroon}\DP{0.41}{RedOrange}\DP{0.03}{Maroon}\DP{0.17}{Gray}\DP{0.04}{Maroon}\DP{0.12}{RedOrange}\DP{0.05}{Maroon}\DP{0.23}{RoyalBlue}&28&\DP{0.10}{SkyBlue}\DP{0.11}{Maroon}\DP{0.10}{SkyBlue}\DP{0.11}{Maroon}\DP{0.15}{CarnationPink}\DP{0.10}{Maroon}\DP{0.92}{YellowGreen}\DP{0.10}{Maroon}\DP{0.66}{WildStrawberry}\DP{0.13}{Maroon}\DP{2.95}{ForestGreen}\DP{0.36}{Maroon}\DP{0.90}{Dandelion}\DP{0.14}{Maroon}\DP{0.29}{Tan}\DP{0.01}{Maroon}\DP{0.12}{Tan}\DP{0.08}{Maroon}\DP{0.13}{CarnationPink}\DP{0.06}{Maroon}\DP{0.36}{Purple}\DP{0.06}{Maroon}\DP{0.20}{Orchid}\DP{0.05}{Maroon}\DP{0.33}{RedOrange}\DP{0.12}{Maroon}\DP{0.36}{Gray}\DP{0.07}{Maroon}\DP{0.22}{RedOrange}\DP{0.08}{Maroon}\DP{0.36}{RoyalBlue}\DP{0.06}{Maroon}\DP{0.20}{RoyalBlue}\\29&\DP{0.10}{SkyBlue}\DP{0.10}{Maroon}\DP{0.10}{SkyBlue}\DP{0.10}{Maroon}\DP{0.16}{CarnationPink}\DP{0.09}{Maroon}\DP{0.87}{YellowGreen}\DP{0.08}{Maroon}\DP{0.42}{WildStrawberry}\DP{0.14}{Maroon}\DP{2.68}{ForestGreen}\DP{0.28}{Maroon}\DP{1.32}{Dandelion}\DP{0.13}{Maroon}\DP{0.21}{Tan}\DP{0.01}{Maroon}\DP{0.18}{Tan}\DP{0.08}{Maroon}\DP{0.15}{CarnationPink}\DP{0.07}{Maroon}\DP{0.38}{Purple}\DP{0.06}{Maroon}\DP{0.20}{Orchid}\DP{0.05}{Maroon}\DP{0.26}{RedOrange}\DP{0.08}{Maroon}\DP{0.61}{Gray}\DP{0.08}{Maroon}\DP{0.28}{RedOrange}\DP{0.07}{Maroon}\DP{0.40}{RoyalBlue}\DP{0.07}{Maroon}\DP{0.17}{RoyalBlue}\DP{0.00}{Maroon}\DP{0.04}{RoyalBlue}&30&\DP{0.10}{SkyBlue}\DP{0.10}{Maroon}\DP{0.09}{SkyBlue}\DP{0.10}{Maroon}\DP{0.17}{CarnationPink}\DP{0.28}{Maroon}\DP{1.09}{YellowGreen}\DP{0.10}{Maroon}\DP{0.50}{WildStrawberry}\DP{0.13}{Maroon}\DP{2.50}{ForestGreen}\DP{0.28}{Maroon}\DP{0.98}{Dandelion}\DP{0.09}{Maroon}\DP{0.13}{Tan}\DP{0.02}{Maroon}\DP{0.12}{Tan}\DP{0.10}{Maroon}\DP{0.23}{CarnationPink}\DP{0.08}{Maroon}\DP{0.35}{Purple}\DP{0.07}{Maroon}\DP{0.32}{Orchid}\DP{0.06}{Maroon}\DP{0.34}{RedOrange}\DP{0.11}{Maroon}\DP{0.40}{Gray}\DP{0.07}{Maroon}\DP{0.29}{RedOrange}\DP{0.08}{Maroon}\DP{0.42}{RoyalBlue}\DP{0.09}{Maroon}\DP{0.21}{RoyalBlue}\\31&\DP{0.11}{SkyBlue}\DP{0.10}{Maroon}\DP{0.10}{SkyBlue}\DP{0.11}{Maroon}\DP{0.17}{CarnationPink}\DP{0.08}{Maroon}\DP{1.07}{YellowGreen}\DP{0.13}{Maroon}\DP{0.59}{WildStrawberry}\DP{0.17}{Maroon}\DP{2.58}{ForestGreen}\DP{0.26}{Maroon}\DP{0.68}{Dandelion}\DP{0.09}{Maroon}\DP{0.43}{Tan}\DP{0.08}{Maroon}\DP{0.23}{CarnationPink}\DP{0.07}{Maroon}\DP{0.62}{Purple}\DP{0.09}{Maroon}\DP{0.11}{Orchid}\DP{0.08}{Maroon}\DP{0.55}{RedOrange}\DP{0.10}{Maroon}\DP{0.33}{Gray}\DP{0.01}{Maroon}\DP{0.56}{RoyalBlue}\DP{0.31}{Maroon}\DP{0.22}{RoyalBlue}&32&\DP{0.07}{SkyBlue}\DP{0.14}{Maroon}\DP{0.07}{SkyBlue}\DP{0.09}{Maroon}\DP{0.09}{CarnationPink}\DP{0.10}{Maroon}\DP{0.80}{YellowGreen}\DP{0.07}{Maroon}\DP{0.24}{WildStrawberry}\DP{0.11}{Maroon}\DP{1.79}{ForestGreen}\DP{0.27}{Maroon}\DP{0.53}{Dandelion}\DP{0.09}{Maroon}\DP{0.12}{Tan}\DP{0.02}{Maroon}\DP{0.22}{Tan}\DP{0.06}{Maroon}\DP{0.17}{CarnationPink}\DP{0.06}{Maroon}\DP{0.21}{Purple}\DP{0.05}{Maroon}\DP{0.21}{Orchid}\DP{0.06}{Maroon}\DP{0.30}{RedOrange}\DP{0.19}{Maroon}\DP{0.43}{RedOrange}\DP{0.07}{Maroon}\DP{0.40}{Gray}\DP{0.01}{Maroon}\DP{0.17}{Gray}\DP{0.01}{Maroon}\DP{0.10}{RoyalBlue}\DP{0.10}{Maroon}\DP{0.33}{RedOrange}\DP{0.05}{Maroon}\DP{0.32}{RoyalBlue}\DP{0.05}{Maroon}\DP{0.24}{RoyalBlue}\DP{0.08}{Maroon}\DP{0.36}{RoyalBlue}\DP{0.33}{Maroon}\DP{0.22}{RedOrange}\DP{0.14}{Maroon}\DP{0.30}{RoyalBlue}\DP{0.07}{Maroon}\DP{0.10}{RoyalBlue}\\33&\DP{0.05}{SkyBlue}\DP{0.08}{Maroon}\DP{0.09}{SkyBlue}\DP{0.21}{Maroon}\DP{0.17}{CarnationPink}\DP{0.08}{Maroon}\DP{0.84}{YellowGreen}\DP{0.07}{Maroon}\DP{0.46}{WildStrawberry}\DP{0.11}{Maroon}\DP{2.72}{ForestGreen}\DP{0.22}{Maroon}\DP{1.05}{Dandelion}\DP{0.11}{Maroon}\DP{0.12}{Tan}\DP{0.02}{Maroon}\DP{0.31}{Tan}\DP{0.08}{Maroon}\DP{0.25}{CarnationPink}\DP{0.07}{Maroon}\DP{0.30}{Purple}\DP{0.06}{Maroon}\DP{0.35}{Orchid}\DP{0.05}{Maroon}\DP{0.43}{RedOrange}\DP{0.05}{Maroon}\DP{0.51}{Gray}\DP{0.07}{Maroon}\DP{0.31}{RedOrange}\DP{0.06}{Maroon}\DP{0.47}{RoyalBlue}\DP{0.07}{Maroon}\DP{0.14}{RoyalBlue}&34&\DP{0.11}{SkyBlue}\DP{0.06}{Maroon}\DP{0.12}{SkyBlue}\DP{0.06}{Maroon}\DP{0.13}{CarnationPink}\DP{0.06}{Maroon}\DP{1.17}{YellowGreen}\DP{0.08}{Maroon}\DP{0.81}{WildStrawberry}\DP{0.19}{Maroon}\DP{0.15}{WildStrawberry}\DP{0.13}{Maroon}\DP{3.00}{ForestGreen}\DP{0.17}{Maroon}\DP{0.84}{Dandelion}\DP{0.10}{Maroon}\DP{0.30}{Tan}\DP{0.02}{Maroon}\DP{0.13}{Tan}\DP{0.06}{Maroon}\DP{0.11}{CarnationPink}\DP{0.05}{Maroon}\DP{0.20}{Purple}\DP{0.04}{Maroon}\DP{0.21}{Orchid}\DP{0.04}{Maroon}\DP{0.41}{RedOrange}\DP{0.06}{Maroon}\DP{0.24}{Gray}\DP{0.01}{Maroon}\DP{0.26}{RoyalBlue}\DP{0.17}{Maroon}\DP{0.12}{RoyalBlue}\DP{0.06}{Maroon}\DP{0.16}{RoyalBlue}\DP{0.04}{Maroon}\DP{0.12}{RoyalBlue}\\35&\DP{0.06}{SkyBlue}\DP{0.06}{Maroon}\DP{0.06}{SkyBlue}\DP{0.08}{Maroon}\DP{0.08}{CarnationPink}\DP{0.05}{Maroon}\DP{0.89}{YellowGreen}\DP{0.08}{Maroon}\DP{0.24}{WildStrawberry}\DP{0.10}{Maroon}\DP{2.19}{ForestGreen}\DP{0.21}{Maroon}\DP{1.31}{Dandelion}\DP{0.07}{Maroon}\DP{0.45}{Tan}\DP{0.11}{Maroon}\DP{0.28}{Tan}\DP{0.07}{Maroon}\DP{0.14}{CarnationPink}\DP{0.06}{Maroon}\DP{0.28}{Purple}\DP{0.06}{Maroon}\DP{0.20}{Orchid}\DP{0.04}{Maroon}\DP{0.39}{RedOrange}\DP{0.05}{Maroon}\DP{0.34}{Gray}\DP{0.06}{Maroon}\DP{1.07}{RedOrange}\DP{0.05}{Maroon}\DP{0.14}{RoyalBlue}\DP{0.01}{Maroon}\DP{0.11}{RoyalBlue}\DP{0.01}{Maroon}\DP{0.34}{RoyalBlue}\DP{0.08}{Maroon}\DP{0.16}{RoyalBlue}&36&\DP{0.08}{SkyBlue}\DP{0.06}{Maroon}\DP{0.07}{SkyBlue}\DP{0.08}{Maroon}\DP{0.10}{CarnationPink}\DP{0.07}{Maroon}\DP{0.69}{YellowGreen}\DP{0.05}{Maroon}\DP{0.34}{WildStrawberry}\DP{0.52}{Maroon}\DP{2.01}{ForestGreen}\DP{0.24}{Maroon}\DP{1.14}{Dandelion}\DP{0.07}{Maroon}\DP{0.45}{Tan}\DP{0.09}{Maroon}\DP{0.34}{Tan}\DP{0.17}{Maroon}\DP{0.24}{Tan}\DP{0.07}{Maroon}\DP{0.13}{CarnationPink}\DP{0.06}{Maroon}\DP{0.36}{Purple}\DP{0.04}{Maroon}\DP{0.21}{Orchid}\DP{0.05}{Maroon}\DP{0.48}{RedOrange}\DP{0.07}{Maroon}\DP{0.51}{Gray}\DP{0.05}{Maroon}\DP{0.55}{RedOrange}\DP{0.06}{Maroon}\DP{0.25}{RoyalBlue}\DP{0.07}{Maroon}\DP{0.21}{RoyalBlue}\\37&\DP{0.08}{SkyBlue}\DP{0.07}{Maroon}\DP{0.13}{SkyBlue}\DP{0.09}{Maroon}\DP{0.15}{CarnationPink}\DP{0.06}{Maroon}\DP{1.01}{YellowGreen}\DP{0.16}{Maroon}\DP{0.55}{WildStrawberry}\DP{0.10}{Maroon}\DP{2.42}{ForestGreen}\DP{0.23}{Maroon}\DP{1.27}{Dandelion}\DP{0.08}{Maroon}\DP{0.13}{Tan}\DP{0.10}{Maroon}\DP{0.14}{CarnationPink}\DP{0.07}{Maroon}\DP{0.31}{Purple}\DP{0.05}{Maroon}\DP{0.28}{Orchid}\DP{0.04}{Maroon}\DP{0.38}{RedOrange}\DP{0.06}{Maroon}\DP{0.48}{Gray}\DP{0.05}{Maroon}\DP{0.19}{RedOrange}\DP{0.06}{Maroon}\DP{0.32}{Gray}\DP{0.12}{Maroon}\DP{0.14}{RedOrange}\DP{0.03}{Maroon}\DP{0.37}{RoyalBlue}\DP{0.08}{Maroon}\DP{0.22}{RoyalBlue}&38&\DP{0.17}{SkyBlue}\DP{0.08}{Maroon}\DP{0.09}{SkyBlue}\DP{0.08}{Maroon}\DP{0.15}{CarnationPink}\DP{0.07}{Maroon}\DP{0.84}{YellowGreen}\DP{0.08}{Maroon}\DP{0.63}{WildStrawberry}\DP{0.11}{Maroon}\DP{2.26}{ForestGreen}\DP{0.35}{Maroon}\DP{0.86}{Dandelion}\DP{0.13}{Maroon}\DP{0.40}{Tan}\DP{0.09}{Maroon}\DP{0.17}{CarnationPink}\DP{0.07}{Maroon}\DP{0.36}{Purple}\DP{0.08}{Maroon}\DP{0.57}{Orchid}\DP{0.06}{Maroon}\DP{0.30}{RedOrange}\DP{0.10}{Maroon}\DP{0.40}{Gray}\DP{0.06}{Maroon}\DP{0.18}{RedOrange}\DP{0.08}{Maroon}\DP{0.20}{RoyalBlue}\DP{0.14}{Maroon}\DP{0.10}{RedOrange}\DP{0.04}{Maroon}\DP{0.33}{RoyalBlue}\DP{0.18}{Maroon}\DP{0.20}{RoyalBlue}\\39&\DP{0.07}{SkyBlue}\DP{0.11}{Maroon}\DP{0.18}{SkyBlue}\DP{0.08}{Maroon}\DP{0.13}{CarnationPink}\DP{0.50}{Maroon}\DP{0.75}{YellowGreen}\DP{0.23}{Maroon}\DP{0.54}{WildStrawberry}\DP{0.55}{Maroon}\DP{2.48}{ForestGreen}\DP{0.26}{Maroon}\DP{1.48}{Dandelion}\DP{0.09}{Maroon}\DP{0.07}{Tan}\DP{0.05}{Maroon}\DP{0.11}{CarnationPink}\DP{0.03}{Maroon}\DP{0.16}{Purple}\DP{0.03}{Maroon}\DP{0.18}{Orchid}\DP{0.04}{Maroon}\DP{0.96}{RedOrange}\DP{0.06}{Maroon}\DP{0.10}{RoyalBlue}\DP{0.02}{Maroon}\DP{0.11}{RoyalBlue}\DP{0.29}{Maroon}\DP{0.08}{RoyalBlue}\DP{0.05}{Maroon}\DP{0.09}{RoyalBlue}\DP{0.05}{Maroon}\DP{0.08}{RoyalBlue}&40&\DP{0.08}{SkyBlue}\DP{0.08}{Maroon}\DP{0.11}{SkyBlue}\DP{0.12}{Maroon}\DP{0.23}{CarnationPink}\DP{0.05}{Maroon}\DP{0.78}{YellowGreen}\DP{0.07}{Maroon}\DP{0.38}{WildStrawberry}\DP{0.16}{Maroon}\DP{3.33}{ForestGreen}\DP{0.24}{Maroon}\DP{1.77}{Dandelion}\DP{0.10}{Maroon}\DP{0.05}{Tan}\DP{0.06}{Maroon}\DP{0.09}{CarnationPink}\DP{0.06}{Maroon}\DP{0.28}{Purple}\DP{0.07}{Maroon}\DP{0.09}{Orchid}\DP{0.04}{Maroon}\DP{0.34}{RedOrange}\DP{0.06}{Maroon}\DP{0.39}{Gray}\DP{0.17}{Maroon}\DP{0.30}{RedOrange}\DP{0.06}{Maroon}\DP{0.29}{RoyalBlue}\DP{0.07}{Maroon}\DP{0.09}{RoyalBlue}\\41&\DP{0.12}{SkyBlue}\DP{0.10}{Maroon}\DP{0.11}{SkyBlue}\DP{0.11}{Maroon}\DP{0.12}{CarnationPink}\DP{0.05}{Maroon}\DP{0.69}{YellowGreen}\DP{0.08}{Maroon}\DP{0.51}{WildStrawberry}\DP{0.09}{Maroon}\DP{3.76}{ForestGreen}\DP{0.24}{Maroon}\DP{1.22}{Dandelion}\DP{0.18}{Maroon}\DP{0.14}{Tan}\DP{0.05}{Maroon}\DP{0.15}{CarnationPink}\DP{0.06}{Maroon}\DP{0.22}{Purple}\DP{0.06}{Maroon}\DP{0.09}{Orchid}\DP{0.04}{Maroon}\DP{0.76}{RedOrange}\DP{0.05}{Maroon}\DP{0.33}{Gray}\DP{0.04}{Maroon}\DP{0.20}{RedOrange}\DP{0.04}{Maroon}\DP{0.07}{RoyalBlue}\DP{0.01}{Maroon}\DP{0.18}{RoyalBlue}\DP{0.05}{Maroon}\DP{0.08}{RoyalBlue}&42&\DP{0.10}{SkyBlue}\DP{0.10}{Maroon}\DP{0.12}{SkyBlue}\DP{0.10}{Maroon}\DP{0.13}{CarnationPink}\DP{0.06}{Maroon}\DP{0.93}{YellowGreen}\DP{0.06}{Maroon}\DP{0.43}{WildStrawberry}\DP{0.19}{Maroon}\DP{3.09}{ForestGreen}\DP{0.25}{Maroon}\DP{1.19}{Dandelion}\DP{0.10}{Maroon}\DP{0.29}{Tan}\DP{0.06}{Maroon}\DP{0.15}{CarnationPink}\DP{0.09}{Maroon}\DP{0.27}{Purple}\DP{0.11}{Maroon}\DP{0.23}{Orchid}\DP{0.06}{Maroon}\DP{0.51}{RedOrange}\DP{0.01}{Maroon}\DP{0.12}{RedOrange}\DP{0.07}{Maroon}\DP{0.41}{Gray}\DP{0.05}{Maroon}\DP{0.25}{RedOrange}\DP{0.05}{Maroon}\DP{0.08}{RoyalBlue}\DP{0.01}{Maroon}\DP{0.08}{RoyalBlue}\DP{0.01}{Maroon}\DP{0.05}{RoyalBlue}\DP{0.07}{Maroon}\DP{0.13}{RoyalBlue}\\43&\DP{0.09}{SkyBlue}\DP{0.11}{Maroon}\DP{0.12}{SkyBlue}\DP{0.18}{Maroon}\DP{0.21}{CarnationPink}\DP{0.13}{Maroon}\DP{1.30}{YellowGreen}\DP{0.05}{Maroon}\DP{0.37}{WildStrawberry}\DP{0.28}{Maroon}\DP{2.71}{ForestGreen}\DP{0.26}{Maroon}\DP{1.18}{Dandelion}\DP{0.12}{Maroon}\DP{0.12}{Tan}\DP{0.08}{Maroon}\DP{0.15}{CarnationPink}\DP{0.07}{Maroon}\DP{0.32}{Purple}\DP{0.13}{Maroon}\DP{0.24}{Orchid}\DP{0.05}{Maroon}\DP{0.51}{RedOrange}\DP{0.09}{Maroon}\DP{0.36}{Gray}\DP{0.06}{Maroon}\DP{0.20}{RedOrange}\DP{0.05}{Maroon}\DP{0.30}{RoyalBlue}\DP{0.06}{Maroon}\DP{0.11}{RoyalBlue}&44&\DP{0.05}{SkyBlue}\DP{0.16}{Maroon}\DP{0.14}{SkyBlue}\DP{0.09}{Maroon}\DP{0.22}{CarnationPink}\DP{0.05}{Maroon}\DP{0.88}{YellowGreen}\DP{0.19}{Maroon}\DP{0.35}{WildStrawberry}\DP{0.09}{Maroon}\DP{3.15}{ForestGreen}\DP{0.20}{Maroon}\DP{0.96}{Dandelion}\DP{0.09}{Maroon}\DP{0.12}{Tan}\DP{0.26}{Maroon}\DP{0.17}{CarnationPink}\DP{0.04}{Maroon}\DP{0.31}{Purple}\DP{0.06}{Maroon}\DP{0.13}{Orchid}\DP{0.05}{Maroon}\DP{0.51}{RedOrange}\DP{0.06}{Maroon}\DP{0.53}{Gray}\DP{0.04}{Maroon}\DP{0.40}{RedOrange}\DP{0.06}{Maroon}\DP{0.39}{RoyalBlue}\DP{0.07}{Maroon}\DP{0.20}{RoyalBlue}\\45&\DP{0.03}{SkyBlue}\DP{0.09}{Maroon}\DP{0.16}{SkyBlue}\DP{0.09}{Maroon}\DP{0.23}{CarnationPink}\DP{0.06}{Maroon}\DP{0.96}{YellowGreen}\DP{0.09}{Maroon}\DP{0.38}{WildStrawberry}\DP{0.18}{Maroon}\DP{2.78}{ForestGreen}\DP{0.21}{Maroon}\DP{1.46}{Dandelion}\DP{0.12}{Maroon}\DP{0.13}{Tan}\DP{0.06}{Maroon}\DP{0.20}{CarnationPink}\DP{0.05}{Maroon}\DP{0.29}{Purple}\DP{0.04}{Maroon}\DP{0.08}{Orchid}\DP{0.05}{Maroon}\DP{0.52}{RedOrange}\DP{0.09}{Maroon}\DP{0.63}{Gray}\DP{0.04}{Maroon}\DP{0.35}{RedOrange}\DP{0.08}{Maroon}\DP{0.21}{RoyalBlue}\DP{0.01}{Maroon}\DP{0.08}{RoyalBlue}\DP{0.09}{Maroon}\DP{0.17}{RoyalBlue}&46&\DP{0.02}{SkyBlue}\DP{0.07}{Maroon}\DP{0.08}{SkyBlue}\DP{0.11}{Maroon}\DP{0.13}{CarnationPink}\DP{0.11}{Maroon}\DP{0.63}{YellowGreen}\DP{0.06}{Maroon}\DP{0.38}{WildStrawberry}\DP{0.17}{Maroon}\DP{4.48}{ForestGreen}\DP{0.21}{Maroon}\DP{1.04}{Dandelion}\DP{0.06}{Maroon}\DP{0.08}{Tan}\DP{0.04}{Maroon}\DP{0.12}{CarnationPink}\DP{0.10}{Maroon}\DP{0.24}{Purple}\DP{0.03}{Maroon}\DP{0.11}{Orchid}\DP{0.04}{Maroon}\DP{0.38}{RedOrange}\DP{0.04}{Maroon}\DP{0.37}{Gray}\DP{0.02}{Maroon}\DP{0.23}{RedOrange}\DP{0.04}{Maroon}\DP{0.08}{RoyalBlue}\DP{0.00}{Maroon}\DP{0.07}{RoyalBlue}\DP{0.01}{Maroon}\DP{0.24}{RoyalBlue}\DP{0.01}{Maroon}\DP{0.05}{RoyalBlue}\DP{0.05}{Maroon}\DP{0.11}{RoyalBlue}\\
47&\DP{0.13}{SkyBlue}\DP{0.09}{Maroon}\DP{0.15}{SkyBlue}\DP{0.05}{Maroon}\DP{0.13}{CarnationPink}\DP{0.26}{Maroon}\DP{1.13}{YellowGreen}\DP{0.07}{Maroon}\DP{0.38}{WildStrawberry}\DP{0.20}{Maroon}\DP{1.74}{ForestGreen}\DP{0.34}{Maroon}\DP{1.95}{Dandelion}\DP{0.07}{Maroon}\DP{0.19}{Tan}\DP{0.07}{Maroon}\DP{0.11}{CarnationPink}\DP{0.08}{Maroon}\DP{0.39}{Purple}\DP{0.06}{Maroon}\DP{0.19}{Orchid}\DP{0.06}{Maroon}\DP{0.91}{RedOrange}\DP{0.08}{Maroon}\DP{0.96}{RoyalBlue}\DP{0.06}{Maroon}\DP{0.15}{RoyalBlue}&48&\DP{0.19}{SkyBlue}\DP{0.12}{Maroon}\DP{0.22}{SkyBlue}\DP{0.14}{Maroon}\DP{0.32}{CarnationPink}\DP{0.26}{Maroon}\DP{0.90}{YellowGreen}\DP{0.08}{Maroon}\DP{0.52}{WildStrawberry}\DP{0.19}{Maroon}\DP{2.18}{ForestGreen}\DP{0.39}{Maroon}\DP{1.10}{Dandelion}\DP{0.11}{Maroon}\DP{0.35}{Tan}\DP{0.11}{Maroon}\DP{0.14}{CarnationPink}\DP{0.16}{Maroon}\DP{0.73}{Purple}\DP{0.10}{Maroon}\DP{0.17}{Orchid}\DP{0.13}{Maroon}\DP{0.82}{RedOrange}\DP{0.08}{Maroon}\DP{0.48}{RoyalBlue}\\49&\DP{0.06}{SkyBlue}\DP{0.06}{Maroon}\DP{0.14}{SkyBlue}\DP{0.06}{Maroon}\DP{0.12}{CarnationPink}\DP{0.04}{Maroon}\DP{0.74}{YellowGreen}\DP{0.06}{Maroon}\DP{0.29}{WildStrawberry}\DP{0.07}{Maroon}\DP{3.71}{ForestGreen}\DP{0.13}{Maroon}\DP{0.99}{Dandelion}\DP{0.06}{Maroon}\DP{0.36}{Tan}\DP{0.08}{Maroon}\DP{0.16}{Tan}\DP{0.07}{Maroon}\DP{0.20}{CarnationPink}\DP{0.05}{Maroon}\DP{0.33}{Purple}\DP{0.05}{Maroon}\DP{0.22}{Orchid}\DP{0.03}{Maroon}\DP{0.36}{RedOrange}\DP{0.07}{Maroon}\DP{0.47}{Gray}\DP{0.05}{Maroon}\DP{0.22}{RedOrange}\DP{0.04}{Maroon}\DP{0.15}{RedOrange}\DP{0.05}{Maroon}\DP{0.40}{RoyalBlue}\DP{0.01}{Maroon}\DP{0.09}{RoyalBlue}&50&\DP{0.06}{SkyBlue}\DP{0.04}{Maroon}\DP{0.13}{SkyBlue}\DP{0.06}{Maroon}\DP{0.10}{CarnationPink}\DP{0.06}{Maroon}\DP{0.79}{YellowGreen}\DP{0.05}{Maroon}\DP{0.52}{WildStrawberry}\DP{0.30}{Maroon}\DP{3.62}{ForestGreen}\DP{0.21}{Maroon}\DP{0.79}{Dandelion}\DP{0.06}{Maroon}\DP{0.16}{Tan}\DP{0.17}{Maroon}\DP{0.06}{CarnationPink}\DP{0.07}{Maroon}\DP{0.72}{Purple}\DP{0.06}{Maroon}\DP{1.17}{RedOrange}\DP{0.13}{Maroon}\DP{0.53}{RoyalBlue}\DP{0.10}{Maroon}\DP{0.06}{RoyalBlue}\\51&\DP{0.11}{SkyBlue}\DP{0.06}{Maroon}\DP{0.09}{SkyBlue}\DP{0.07}{Maroon}\DP{0.09}{CarnationPink}\DP{0.05}{Maroon}\DP{0.91}{YellowGreen}\DP{0.08}{Maroon}\DP{0.31}{WildStrawberry}\DP{0.08}{Maroon}\DP{3.08}{ForestGreen}\DP{0.18}{Maroon}\DP{1.25}{Dandelion}\DP{0.07}{Maroon}\DP{0.47}{Tan}\DP{0.06}{Maroon}\DP{0.13}{CarnationPink}\DP{0.05}{Maroon}\DP{0.31}{Purple}\DP{0.06}{Maroon}\DP{0.92}{Orchid}\DP{0.05}{Maroon}\DP{0.34}{RedOrange}\DP{0.07}{Maroon}\DP{0.42}{Gray}\DP{0.08}{Maroon}\DP{0.17}{RedOrange}\DP{0.05}{Maroon}\DP{0.37}{RoyalBlue}&52&\DP{0.13}{SkyBlue}\DP{0.07}{Maroon}\DP{0.10}{SkyBlue}\DP{0.07}{Maroon}\DP{0.12}{CarnationPink}\DP{0.07}{Maroon}\DP{0.83}{YellowGreen}\DP{0.10}{Maroon}\DP{0.22}{WildStrawberry}\DP{0.01}{Maroon}\DP{0.51}{WildStrawberry}\DP{0.14}{Maroon}\DP{2.65}{ForestGreen}\DP{0.22}{Maroon}\DP{0.92}{Dandelion}\DP{0.08}{Maroon}\DP{0.31}{Tan}\DP{0.07}{Maroon}\DP{0.15}{CarnationPink}\DP{0.05}{Maroon}\DP{0.57}{Purple}\DP{0.08}{Maroon}\DP{0.11}{Orchid}\DP{0.04}{Maroon}\DP{0.42}{RedOrange}\DP{0.08}{Maroon}\DP{0.48}{Gray}\DP{0.05}{Maroon}\DP{0.34}{Gray}\DP{0.18}{Maroon}\DP{0.18}{RedOrange}\DP{0.06}{Maroon}\DP{0.35}{RoyalBlue}\DP{0.08}{Maroon}\DP{0.17}{RoyalBlue}\\53&\DP{0.10}{SkyBlue}\DP{0.08}{Maroon}\DP{0.12}{SkyBlue}\DP{0.09}{Maroon}\DP{0.21}{CarnationPink}\DP{0.06}{Maroon}\DP{0.56}{YellowGreen}\DP{0.06}{Maroon}\DP{0.56}{WildStrawberry}\DP{0.10}{Maroon}\DP{3.62}{ForestGreen}\DP{0.16}{Maroon}\DP{1.00}{Dandelion}\DP{0.06}{Maroon}\DP{0.33}{Tan}\DP{0.05}{Maroon}\DP{0.14}{CarnationPink}\DP{0.07}{Maroon}\DP{0.53}{Purple}\DP{0.05}{Maroon}\DP{0.68}{Orchid}\DP{0.05}{Maroon}\DP{0.45}{RedOrange}\DP{0.06}{Maroon}\DP{0.17}{Gray}\DP{0.04}{Maroon}\DP{0.28}{RedOrange}\DP{0.06}{Maroon}\DP{0.27}{RoyalBlue}&54&\DP{0.09}{SkyBlue}\DP{0.05}{Maroon}\DP{0.13}{SkyBlue}\DP{0.06}{Maroon}\DP{0.14}{CarnationPink}\DP{0.16}{Maroon}\DP{0.70}{YellowGreen}\DP{0.08}{Maroon}\DP{0.71}{WildStrawberry}\DP{0.04}{Maroon}\DP{0.48}{WildStrawberry}\DP{0.10}{Maroon}\DP{1.84}{ForestGreen}\DP{0.20}{Maroon}\DP{0.86}{Dandelion}\DP{0.06}{Maroon}\DP{0.19}{Tan}\DP{0.01}{Maroon}\DP{0.18}{Tan}\DP{0.04}{Maroon}\DP{0.20}{CarnationPink}\DP{0.05}{Maroon}\DP{0.21}{Purple}\DP{0.03}{Maroon}\DP{0.24}{Orchid}\DP{0.03}{Maroon}\DP{1.55}{RedOrange}\DP{0.07}{Maroon}\DP{0.43}{Gray}\DP{0.07}{Maroon}\DP{0.12}{RedOrange}\DP{0.05}{Maroon}\DP{0.58}{RoyalBlue}\DP{0.08}{Maroon}\DP{0.09}{RoyalBlue}\DP{0.04}{Maroon}\DP{0.05}{RoyalBlue}\\55&\DP{0.12}{SkyBlue}\DP{0.10}{Maroon}\DP{0.09}{SkyBlue}\DP{0.08}{Maroon}\DP{0.17}{CarnationPink}\DP{0.16}{Maroon}\DP{0.52}{YellowGreen}\DP{0.08}{Maroon}\DP{0.54}{WildStrawberry}\DP{0.21}{Maroon}\DP{3.16}{ForestGreen}\DP{0.20}{Maroon}\DP{1.85}{Dandelion}\DP{0.18}{Maroon}\DP{0.14}{Tan}\DP{0.08}{Maroon}\DP{0.12}{CarnationPink}\DP{0.12}{Maroon}\DP{0.36}{Purple}\DP{0.06}{Maroon}\DP{0.23}{Orchid}\DP{0.14}{Maroon}\DP{0.31}{RedOrange}\DP{0.06}{Maroon}\DP{0.25}{Gray}\DP{0.06}{Maroon}\DP{0.19}{RedOrange}\DP{0.05}{Maroon}\DP{0.38}{RoyalBlue}&56&\DP{0.11}{SkyBlue}\DP{0.08}{Maroon}\DP{0.07}{SkyBlue}\DP{0.06}{Maroon}\DP{0.16}{CarnationPink}\DP{0.05}{Maroon}\DP{0.56}{YellowGreen}\DP{0.07}{Maroon}\DP{0.27}{WildStrawberry}\DP{0.13}{Maroon}\DP{2.20}{ForestGreen}\DP{0.22}{Maroon}\DP{0.73}{Dandelion}\DP{0.08}{Maroon}\DP{0.36}{Tan}\DP{0.14}{Maroon}\DP{0.12}{CarnationPink}\DP{0.69}{Maroon}\DP{0.42}{Purple}\DP{0.06}{Maroon}\DP{0.41}{Orchid}\DP{0.01}{Maroon}\DP{0.46}{Orchid}\DP{0.26}{Maroon}\DP{0.28}{RedOrange}\DP{0.10}{Maroon}\DP{0.42}{Gray}\DP{0.06}{Maroon}\DP{0.53}{RedOrange}\DP{0.10}{Maroon}\DP{0.56}{RoyalBlue}\DP{0.08}{Maroon}\DP{0.18}{RoyalBlue}\\
\midrule
\end{tabular}
}
\resizebox{1\textwidth}{!}{%
\begin{tabular}{m{1.5cm}m{23cm}}
Colormap &  Incision \DP{0.9}{SkyBlue}, 
Viscoelastic \DP{0.9}{CarnationPink},
Capsulorhexis \DP{0.9}{YellowGreen},
Hydrodissection \DP{0.9}{WildStrawberry},
Phacoemulsification \DP{0.9}{ForestGreen},
Irrigation-Aspiration \DP{0.9}{Dandelion},
Capsule polishing \DP{0.9}{Tan}
\\
&
Lens Implantation \DP{0.9}{Purple},
Lens Positioning \DP{0.9}{Orchid},
Viscoelastic-Suction \DP{0.9}{RedOrange},
Anterior-Chamber Flushing \DP{0.9}{Gray},
Tonifying/Antibiotics \DP{0.9}{RoyalBlue}, Idle \DP{0.9}{Maroon}
\\
\specialrule{.12em}{.05em}{.05em}
\end{tabular}
}
\end{table}

\begin{figure*}[t!]
    \centering
    \includegraphics[width=0.68\textwidth]{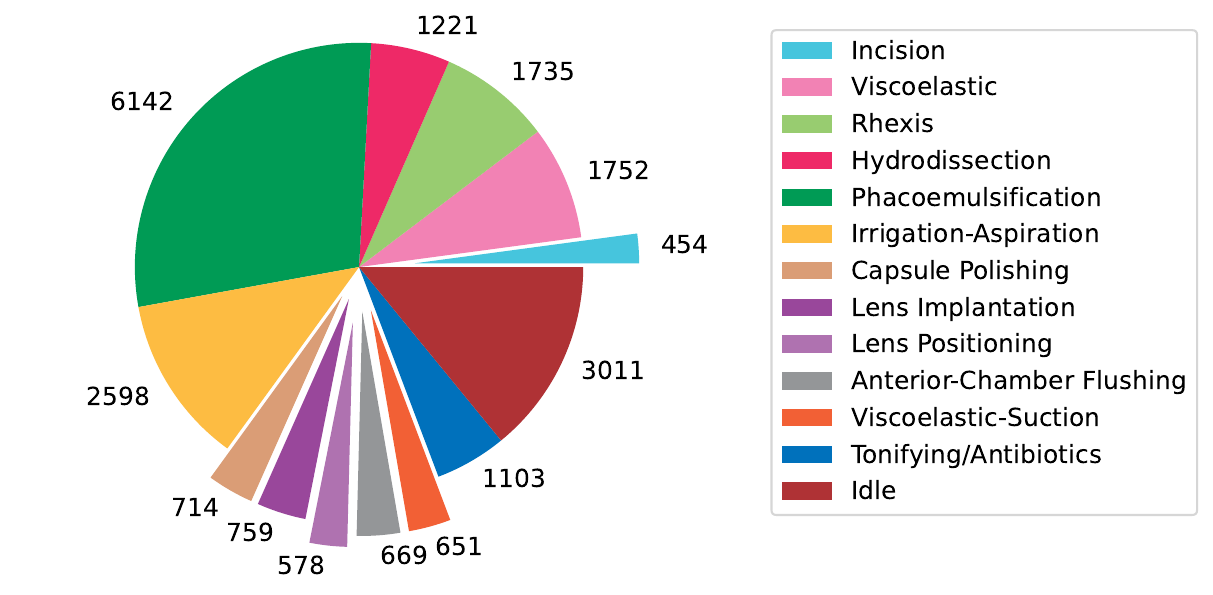}
    \caption{Total duration of the annotated phases in the 56 annotated cataract surgery videos (in seconds).}
    \label{fig:phase_statistics_pie}
\end{figure*}

\begin{itemize}
    \item As illustrated in Figure \ref{fig:phases-visualization}, instruments, which play a fundamental role in distinguishing between relevant phases, share a substantial resemblance in certain phases, leading to a narrow variation between different classes in a trained classification model.
    \item As shown in Figure \ref{fig:phase_statistics_pie}, phase recognition datasets for cataract surgery are extremely imbalanced, as the longest phase (phacoemulsification) and the shortest phase (incision) cover $28.72\%$ and $2.1\%$ of the annotations, respectively.
    \item Videos may exhibit defocus blur stemming from manual camera focus adjustments \cite{DCS}.
    \item Unintentional eye movements and rapid instrument motions close to the camera result in motion blur, impairing distinctive spatial details.
    
    \item Lack of metadata in stored videos precludes additional contextual information.
    \item Variances in patients' eye visuals generate substantial inter-video distribution disparities, demanding ample training data to build networks with generalizable performance.
\end{itemize}

\begin{figure*}[t!]
    \centering
    
    \begin{subfigure}[t]{1\textwidth}
      \includegraphics[width=0.33\textwidth]{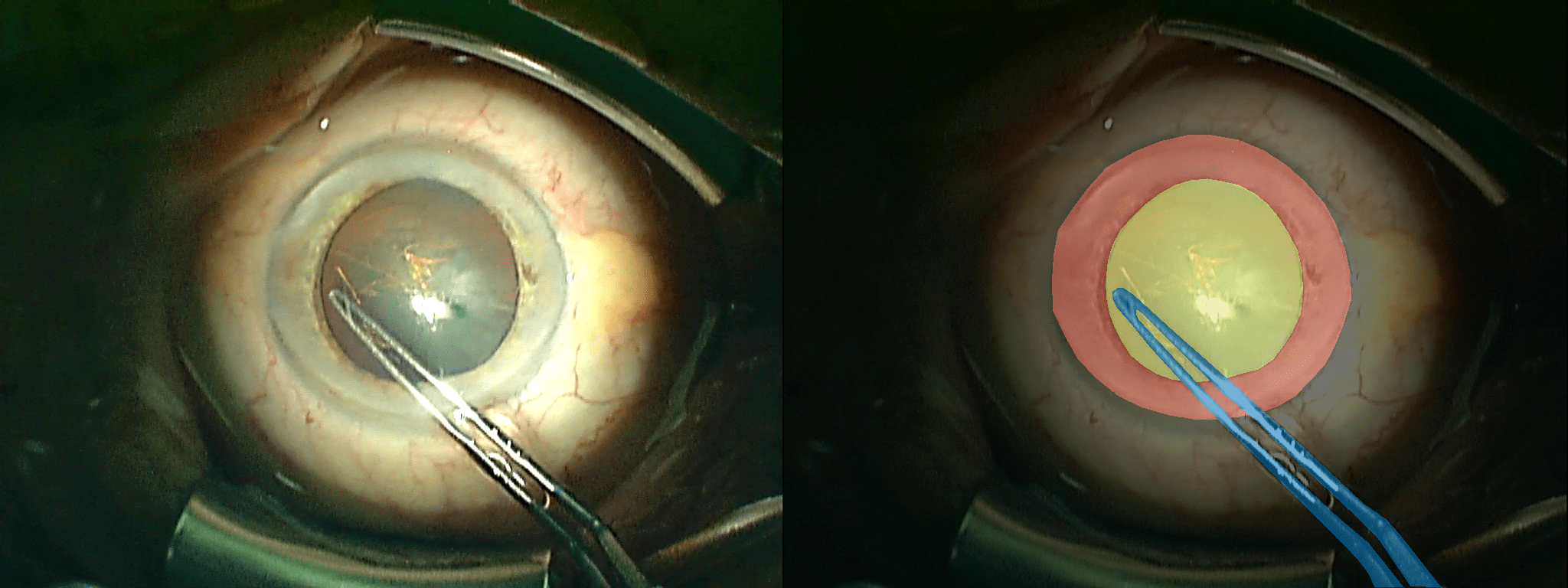}
      \includegraphics[width=0.33\textwidth]{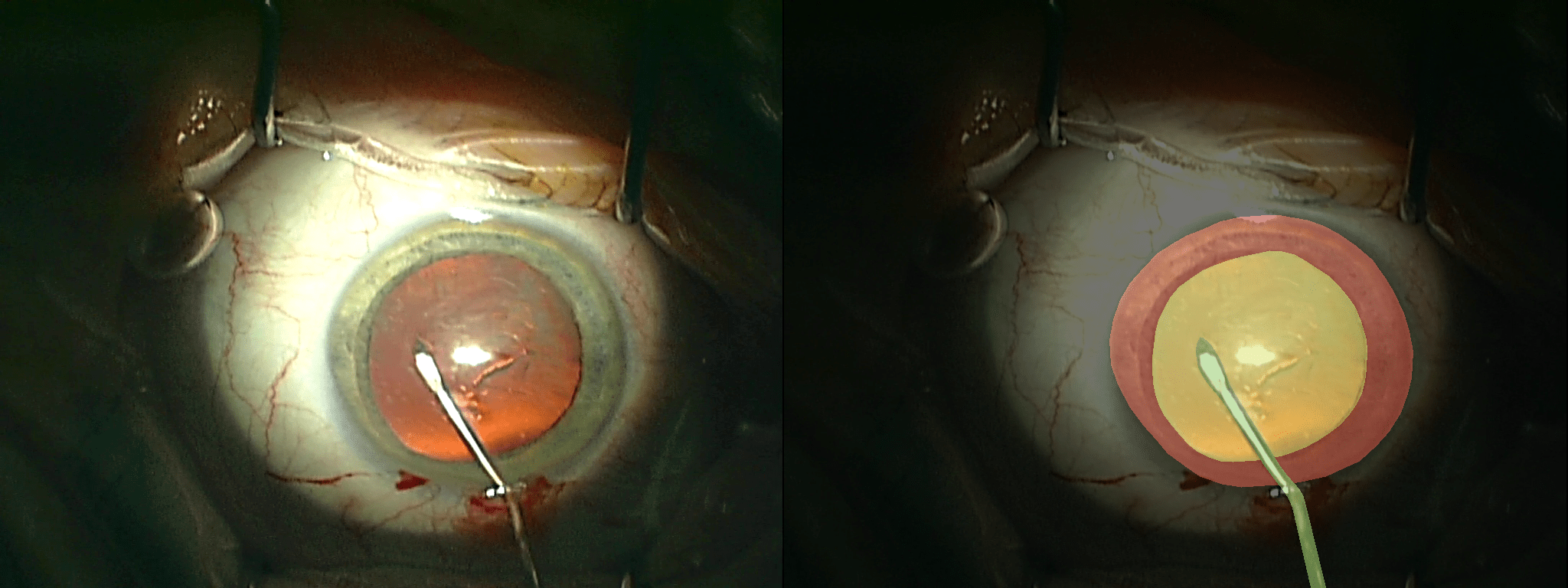}
        \includegraphics[width=0.33\textwidth]{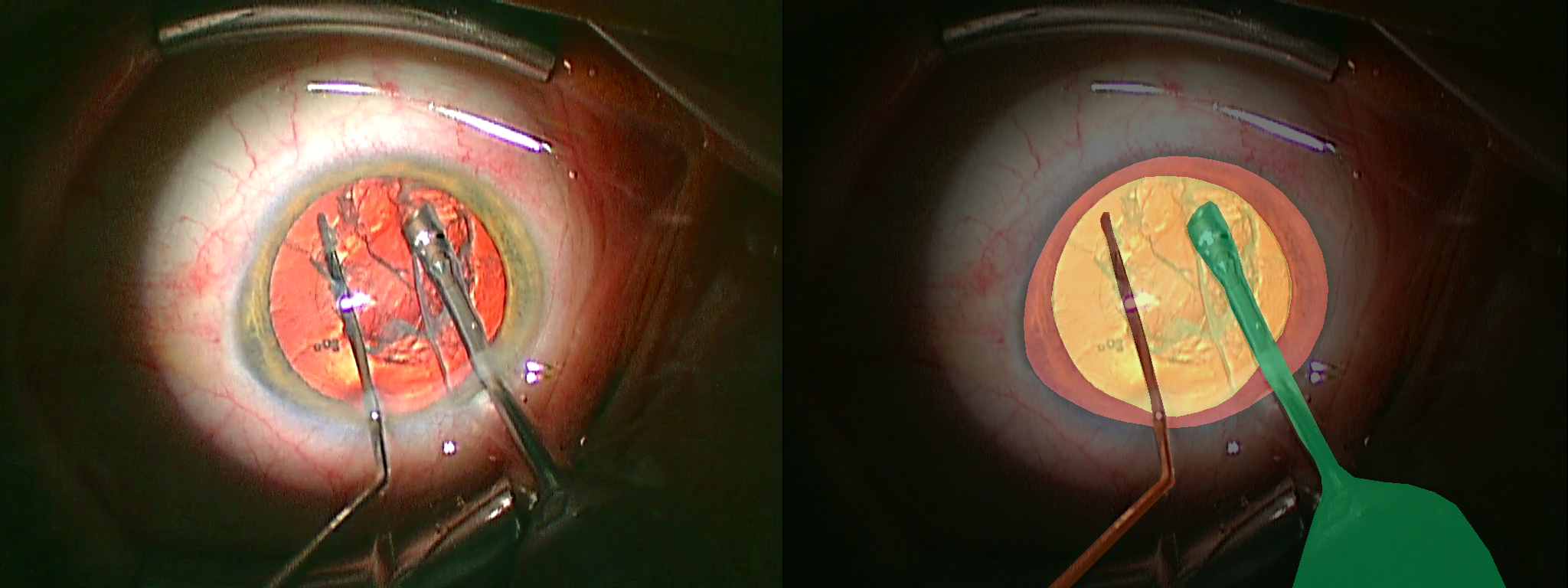}
        \caption{Color and texture variation in Pupil.}
    \end{subfigure}\\
    \begin{subfigure}[t]{0.33\textwidth}
        \includegraphics[width=\textwidth]{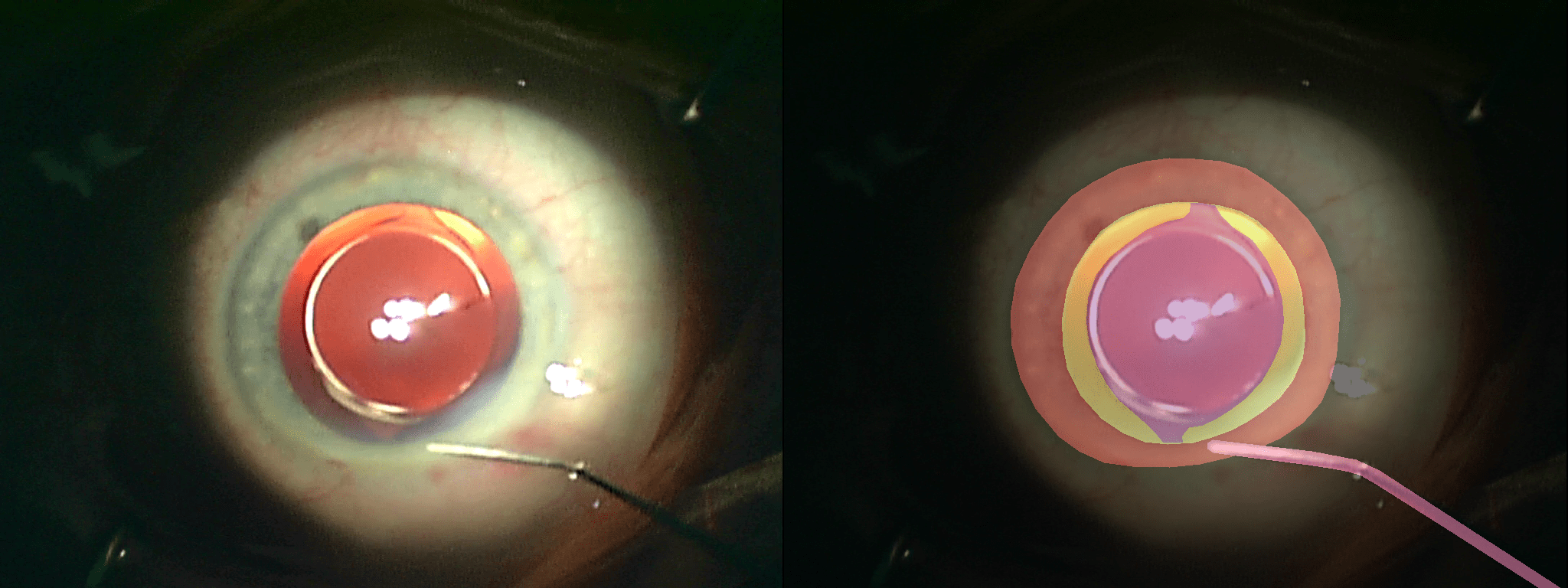}
        \caption{Smooth edges in Iris.}    
    \end{subfigure}
    \begin{subfigure}[t]{0.33\textwidth}
        \includegraphics[width=\textwidth]{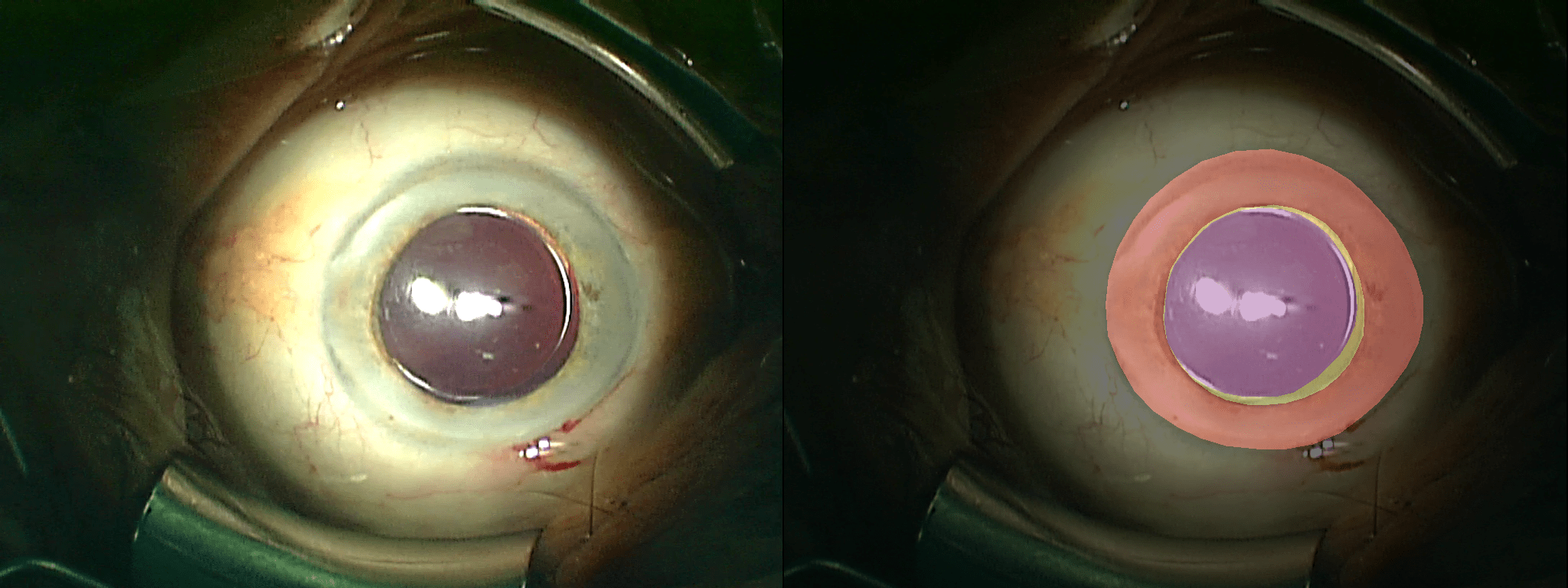}
        \caption{Transparency in intraocular lens.}
    \end{subfigure}
    \begin{subfigure}[t]{0.33\textwidth}
        \includegraphics[width=\textwidth]{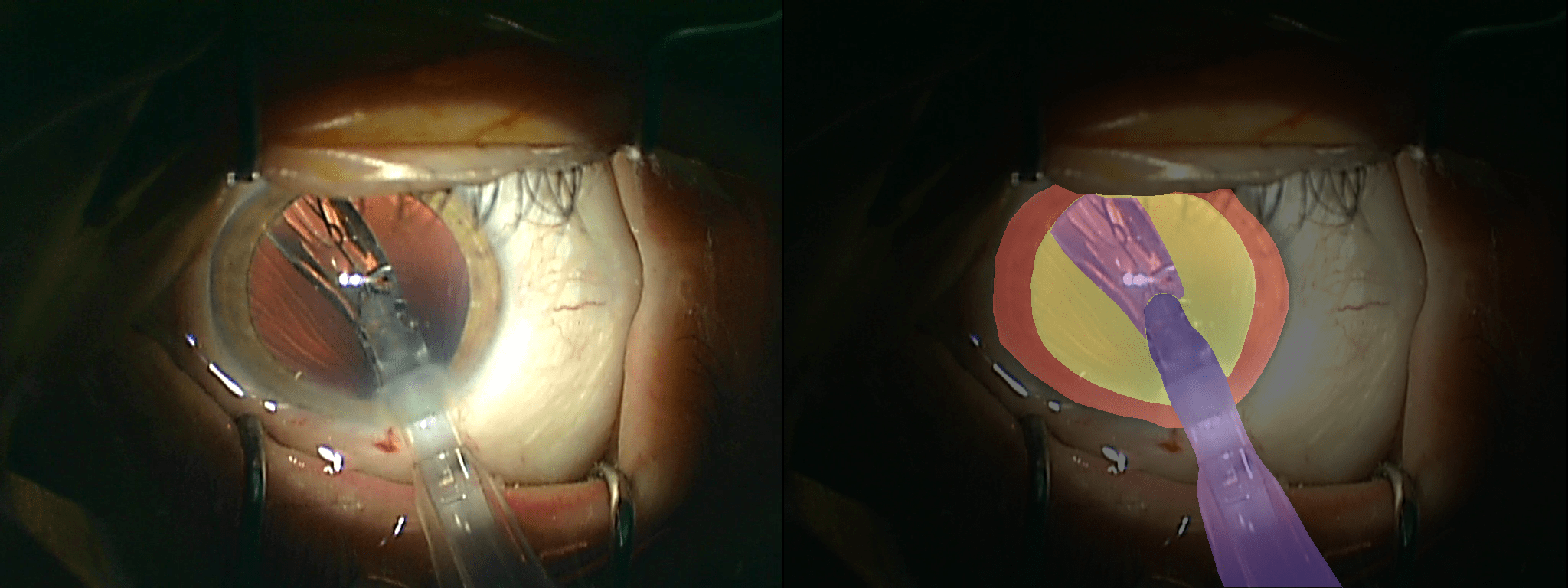}
        \caption{Deformations in the intraocular lens.}    
    \end{subfigure}\\
    \begin{subfigure}[t]{0.33\textwidth}
      \includegraphics[width=\textwidth]{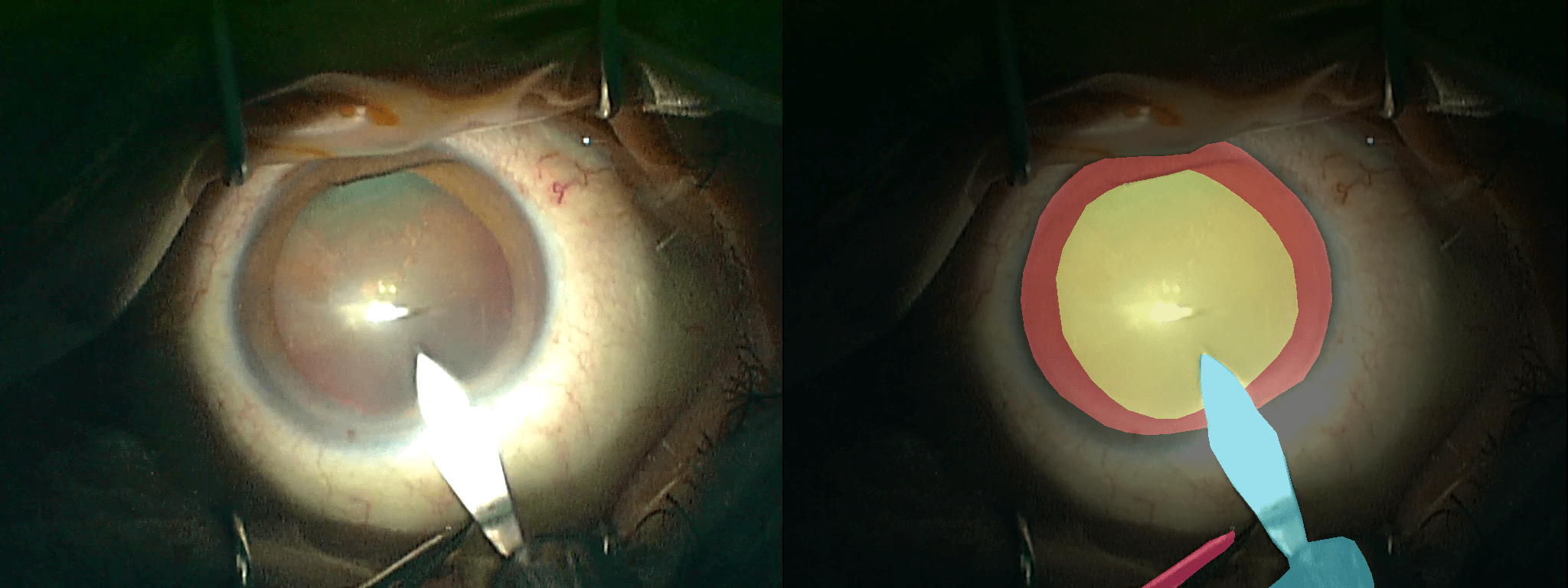}
      \caption{Reflection in the instruments.}
    \end{subfigure}
    \begin{subfigure}[t]{0.33\textwidth}
      \includegraphics[width=\textwidth]{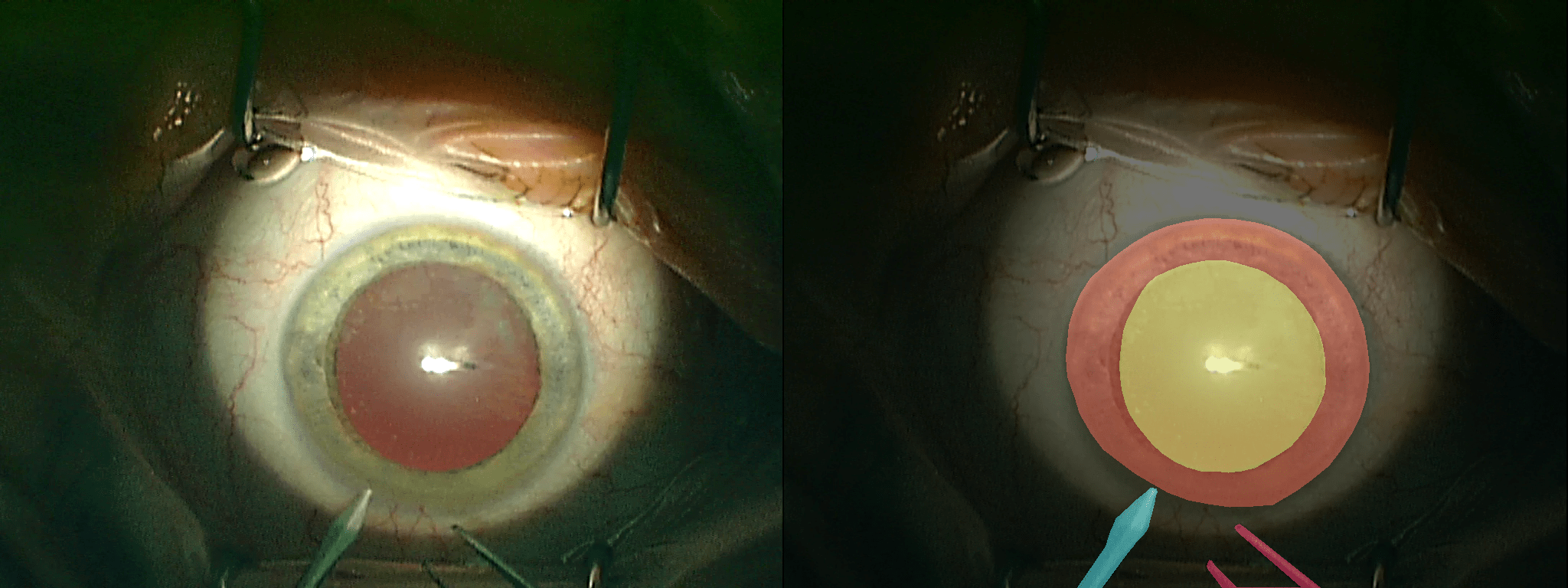}
      \caption{Katena Forceps with out-of-scene connecting point.}
    \end{subfigure}
    \begin{subfigure}[t]{0.33\textwidth}
        \includegraphics[width=\textwidth]{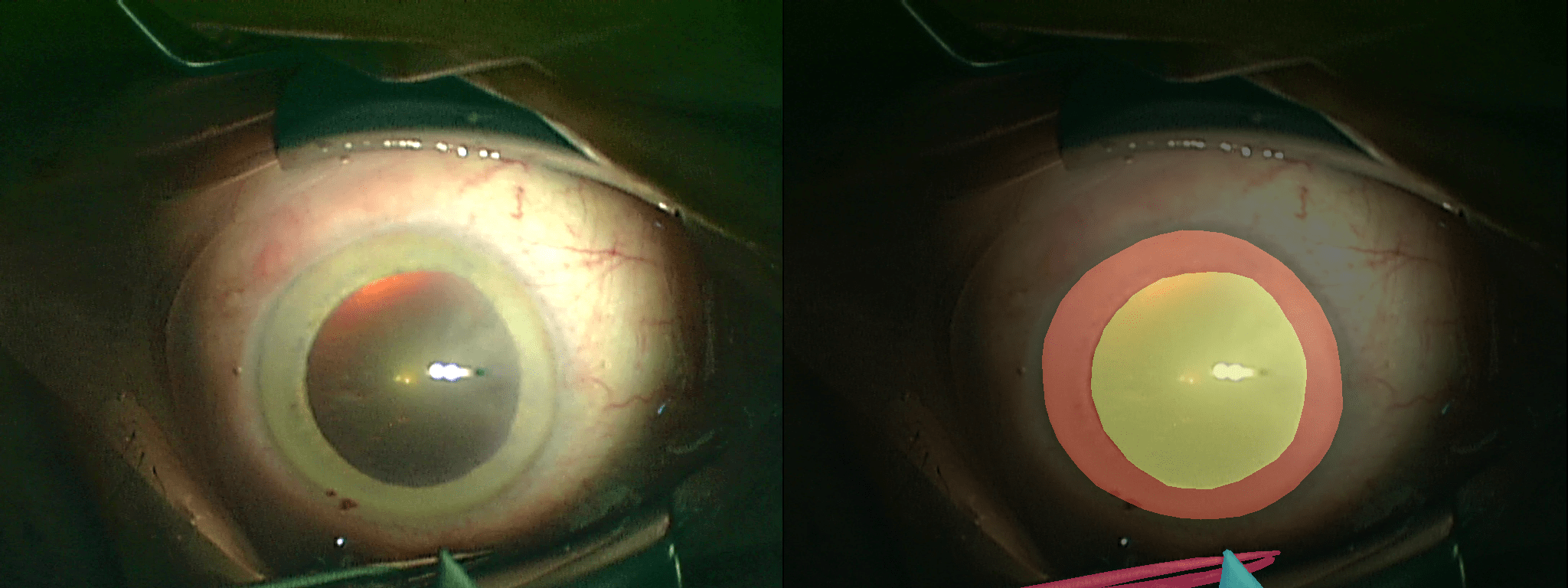}
        \caption{Katena Forceps and Incision Knife with partly visibility.}
    \end{subfigure}\\
    \begin{subfigure}[t]{1\textwidth}
      \includegraphics[width=0.33\textwidth]{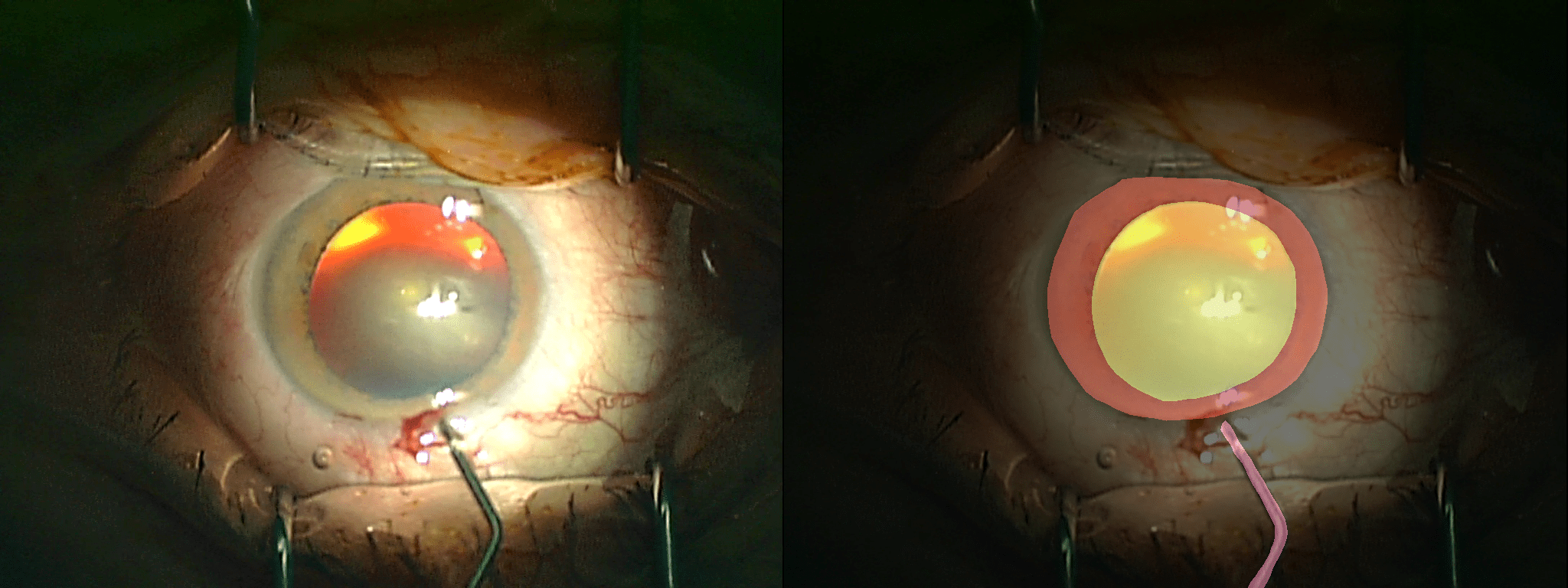}
      \includegraphics[width=0.33\textwidth]{Figures/case5016_13overlayed.png}
        \includegraphics[width=0.33\textwidth]{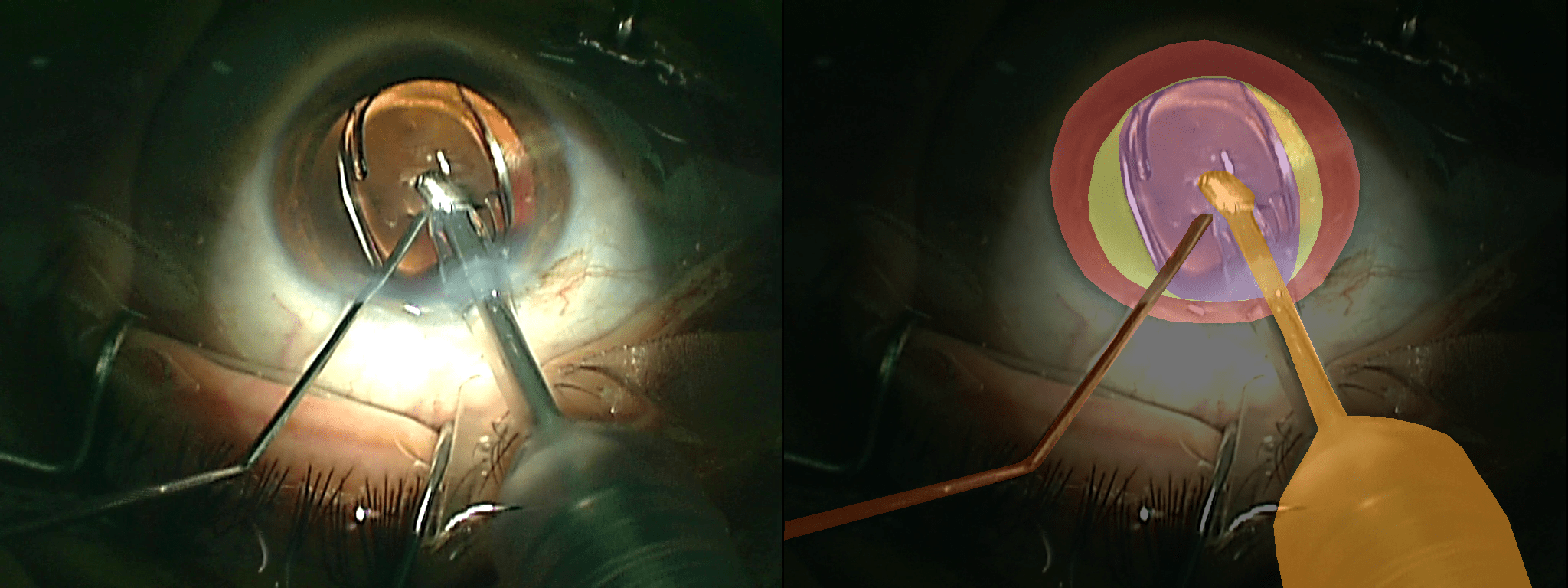}
        \caption{Visual similarities between Gauge, Capsulorhexis Cystotome, and Spatula.}
    \end{subfigure}\\
    \begin{subfigure}[t]{0.33\textwidth}
        \includegraphics[width=\textwidth]{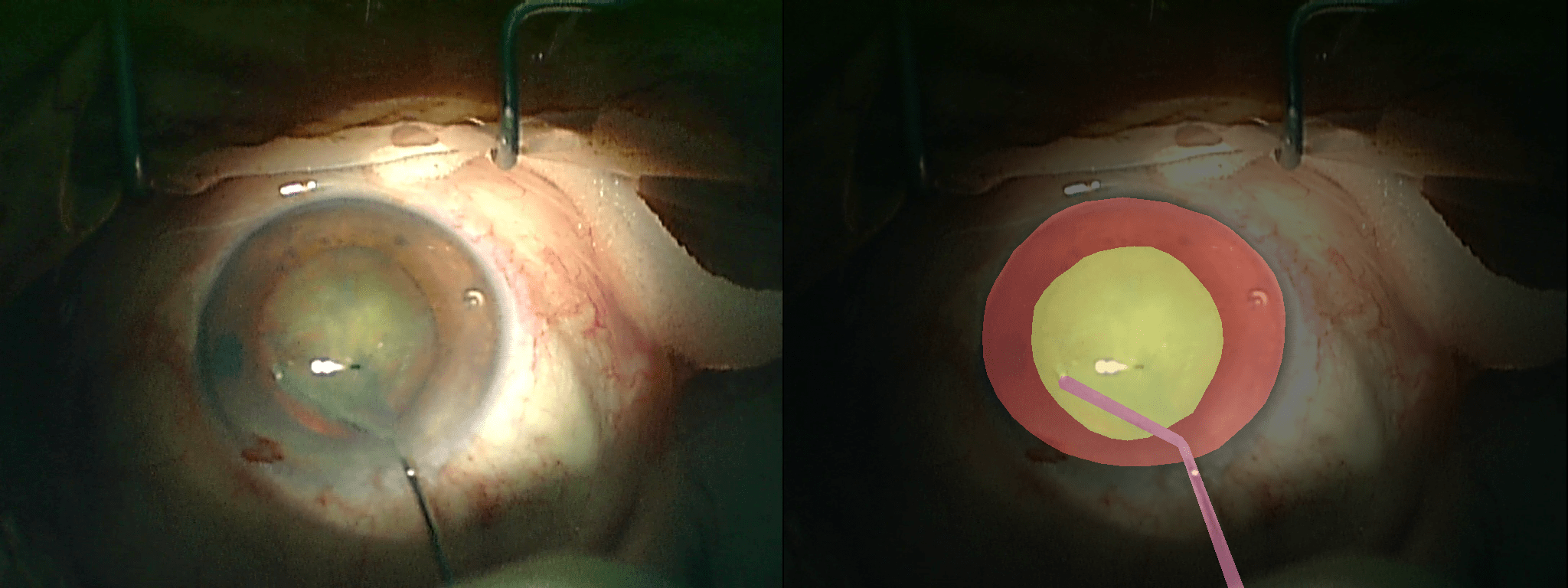}
        \caption{Instrument occlusion by corrupted natural lens.}
    \end{subfigure}
    \begin{subfigure}[t]{0.33\textwidth}
        \includegraphics[width=\textwidth]{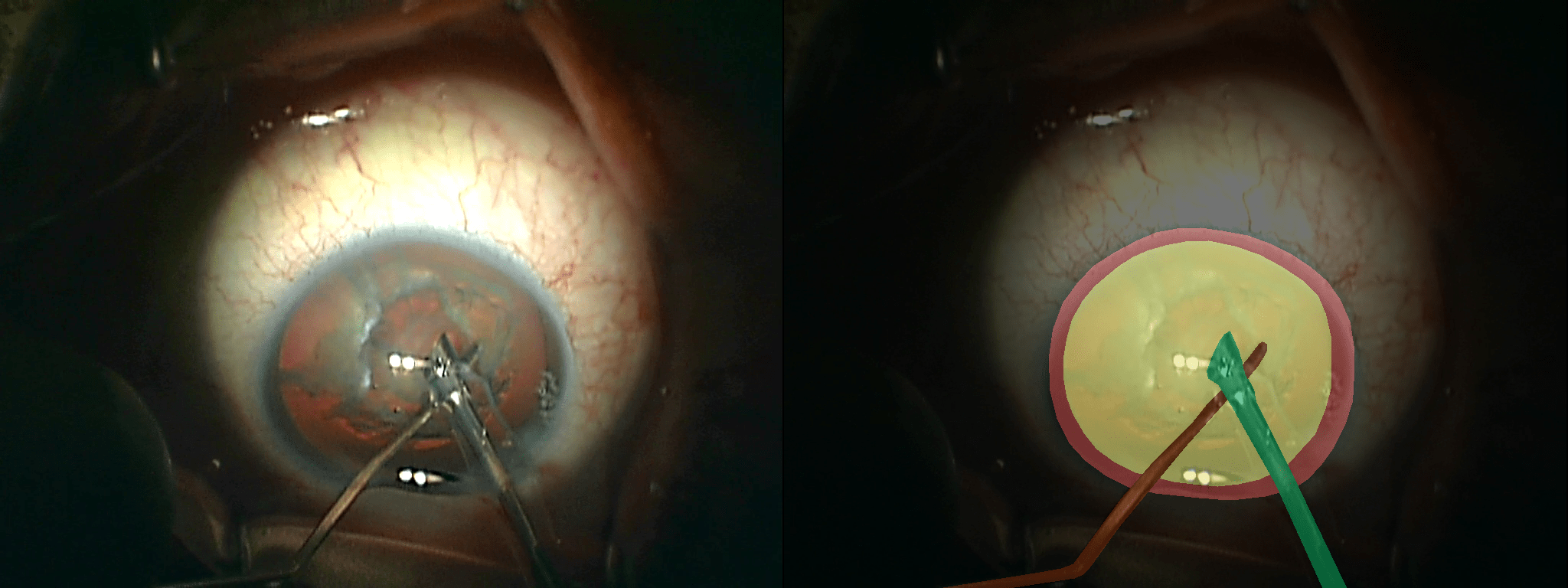}
        \caption{Instrument occlusion by other instruments.}
    \end{subfigure}
    \begin{subfigure}[t]{0.33\textwidth}
        \includegraphics[width=\textwidth]{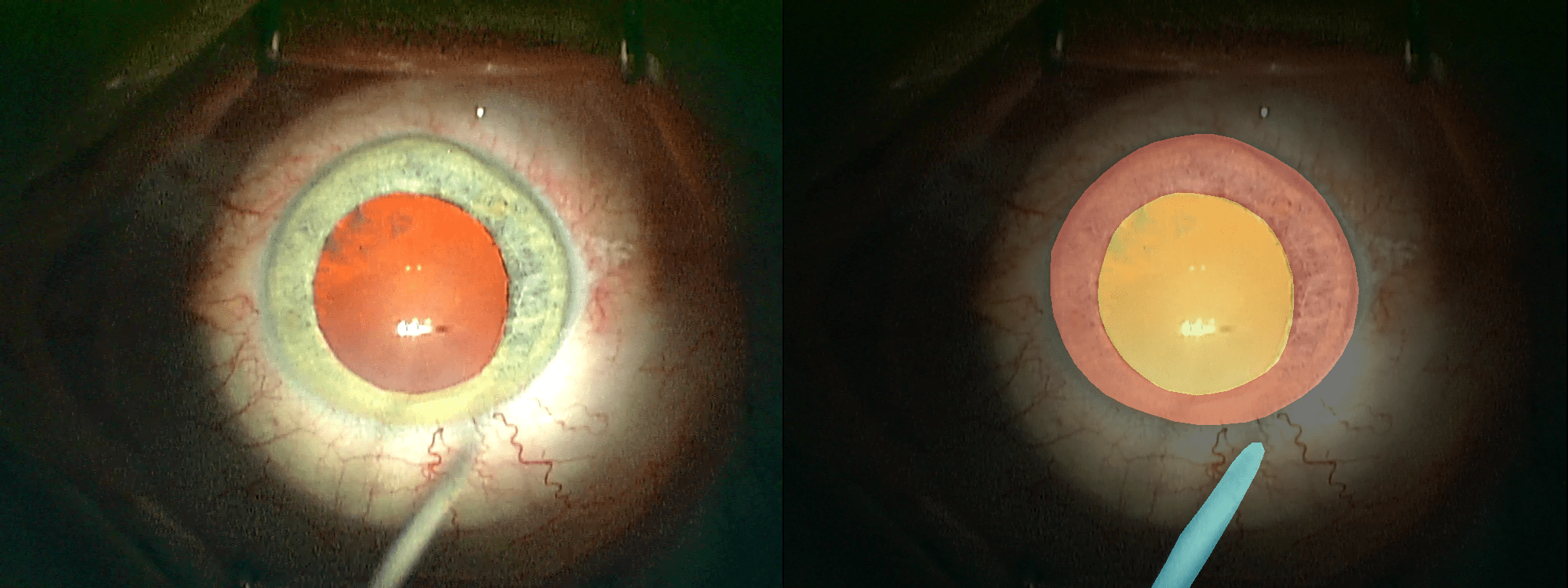}
        \caption{Motion blur in the instruments.}    
    \end{subfigure}
    \\
    \caption{Visualization of pixel-based annotations corresponding to relevant anatomical structures and instruments in cataract surgery and the challenges associated with different object 
(Iris: \DP{0.9}{Maroon}, Pupil: \DP{0.9}{GreenYellow}, Lens: \DP{0.9}{Orchid}, Slit/Incision Knife: \DP{0.9}{SkyBlue}, Gauge: \DP{0.9}{CarnationPink}, Spatula: \DP{0.9}{Sepia}, Capsulorhexis Cystotome: \DP{0.9}{YellowGreen}, Phacoemulsifier Tip: \DP{0.9}{ForestGreen}, Irrigation-Aspiration: \DP{0.9}{BurntOrange}, Lens Injector : \DP{0.9}{RoyalPurple}, Capsulorhexis Forceps: \DP{0.9}{RoyalBlue}, Katena Forceps: \DP{0.9}{WildStrawberry}).}
    \label{fig:segmentation-visualization-challenges}
\end{figure*}

As shown in Figure \ref{fig:phases-visualization}, regular cataract surgery can include twelve action phases, including incision, viscoelastic, capsulorhexis, hydrodissection, phacoemulsification, irrigation-aspiration, capsule polishing, lens implantation, lens positioning, viscoelastic-suction, anterior-chamber flushing, and tonifying/antibiotics. Besides, the idle phases refer to the time spans in the middle of a phase or between two phases when the surgeons mainly change the instruments and no instrument is visible inside the frames. 
We provide a large annotated dataset to enable comprehensive studies on deep-learning-based phase recognition in cataract surgery videos. 

Table \ref{tab:phase-annotations} visualizes the phase annotations corresponding to 56 regular cataract surgery videos, with a spatial resolution of $1024 \times 768$, a temporal resolution of 30 fps, an average duration of 6.45 minutes, and a standard deviation of 2.04. 
This dataset comprises patients with an average age of 75 years, ranging from 51 to 93 years, and a standard deviation of 8.69 years. 
The videos present in the phase recognition dataset correspond to surgeries executed by surgeons with an average experience of 8929 surgeries and a standard deviation of 6350 surgeries.

\paragraph{Semantic segmentation dataset. }Figure \ref{fig:segmentation-visualization-challenges} visualizes pixel-level annotations for relevant anatomical objects and instruments. As illustrated in Figure \ref{fig:segmentation-visualization-challenges}, semantic segmentation in cataract surgery videos poses the following challenges \cite{ReCal-Net,ghamsarian2022deeppyramid,ghamsarian2023deeppyramid+}:

\begin{itemize}
    \item Variations in color, shape, size, and texture in pupil.
    \item Transparency and deformations in the artificial lens,
    \item Smooth edges and color variations in iris,
    \item Occlusion, motion blur, reflection, and partly visibility in instruments,
    \item Visual similarities between different instruments in case of multi-class instrument segmentation,
\end{itemize}

\begin{figure*}[t!]
    \centering   
    \begin{subfigure}[t]{1\textwidth}
    \begin{overpic}[width=0.19\textwidth]{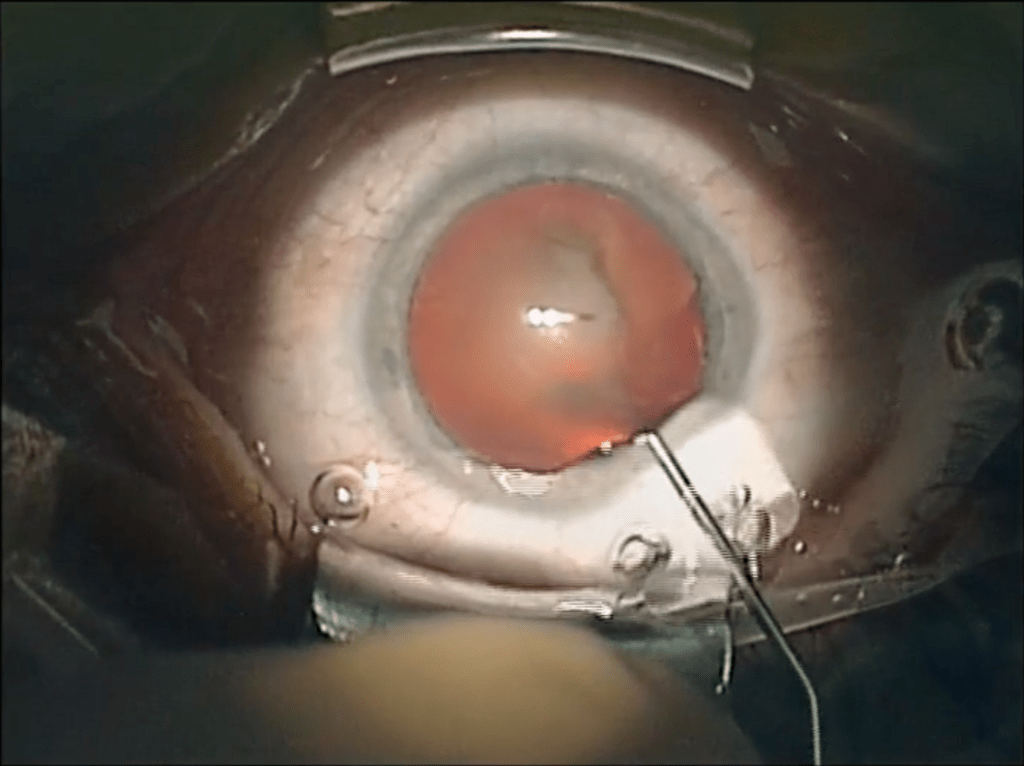}
      \put (3,65) {\footnotesize\textcolor{white}{00:00:27.41}}
     \end{overpic}
      \begin{overpic}[width=0.19\textwidth]{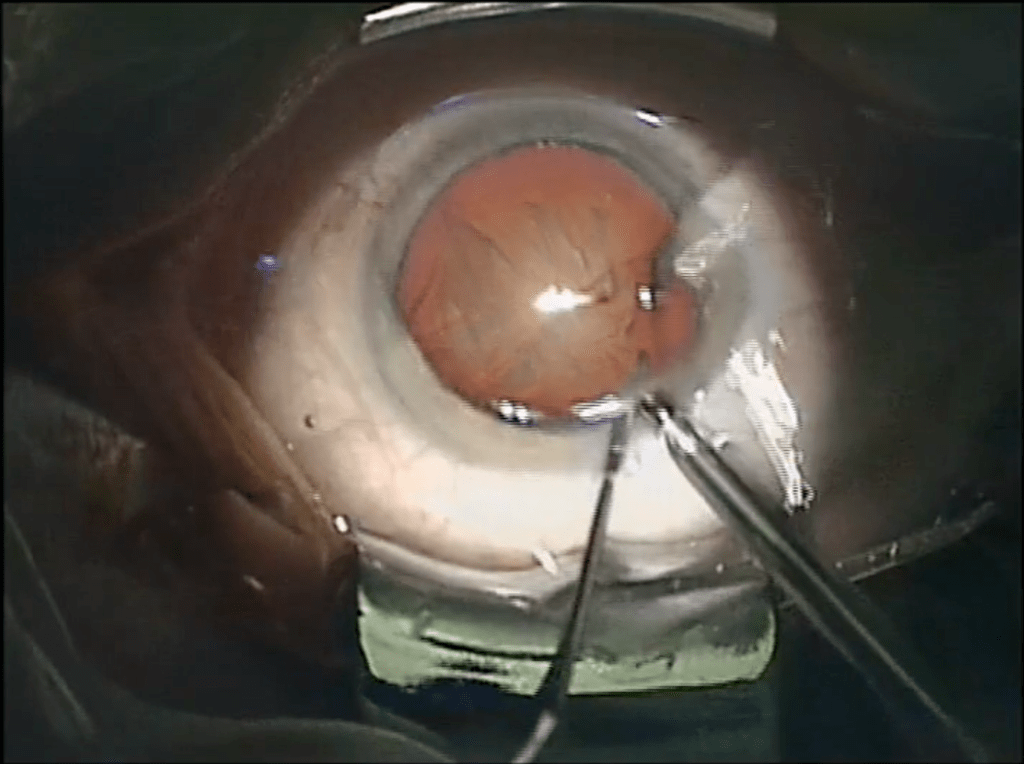}
      \put (3,65) {\footnotesize\textcolor{white}{00:01:2.89}}
     \end{overpic} 
     \begin{overpic}[width=0.19\textwidth]{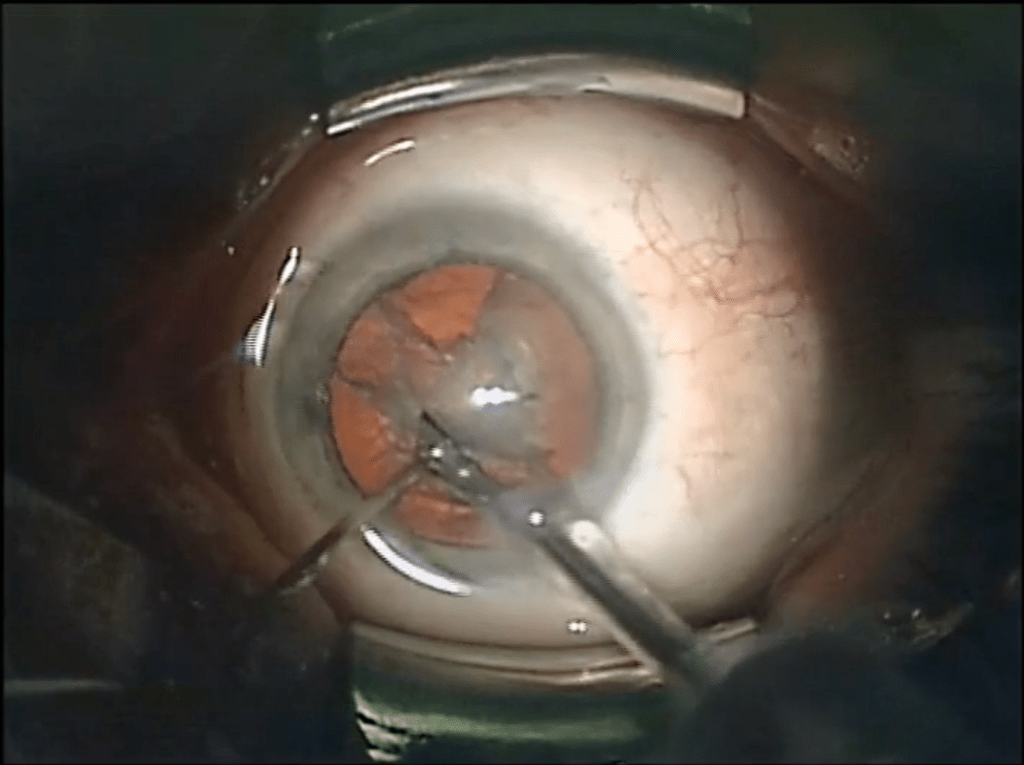}
      \put (3,65) {\footnotesize\textcolor{white}{00:01:40.30}}
     \end{overpic} 
     \begin{overpic}[width=0.19\textwidth]{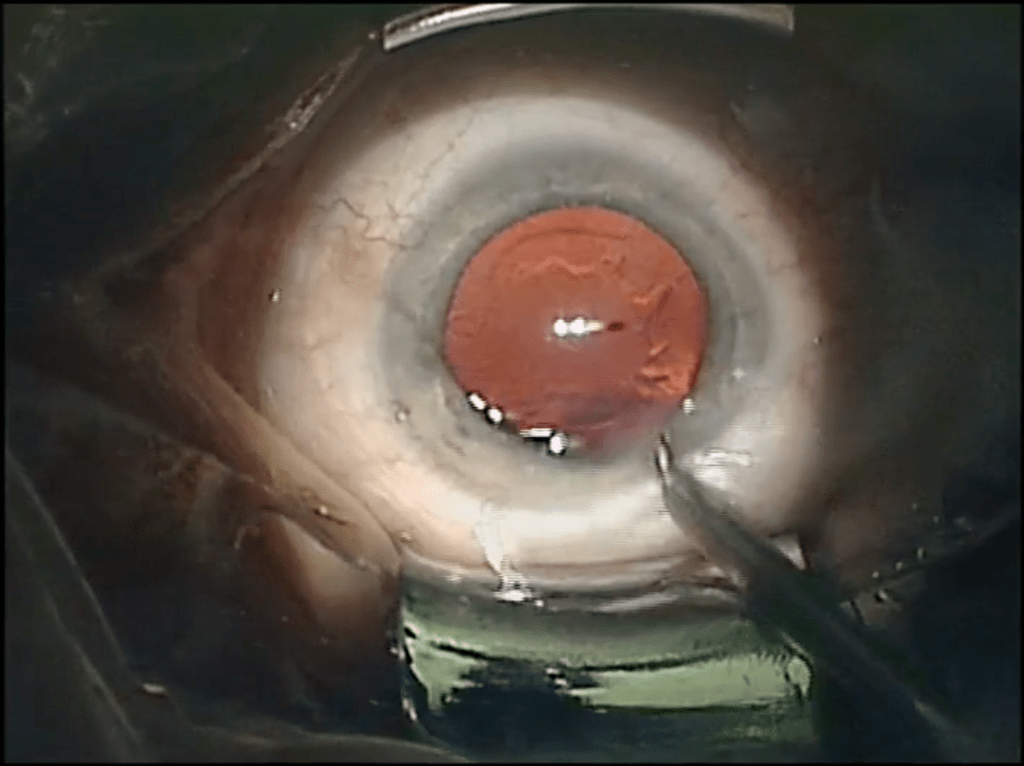}
      \put (3,65) {\footnotesize\textcolor{white}{00:01:58.31}}
     \end{overpic}  
     \begin{overpic}[width=0.19\textwidth]{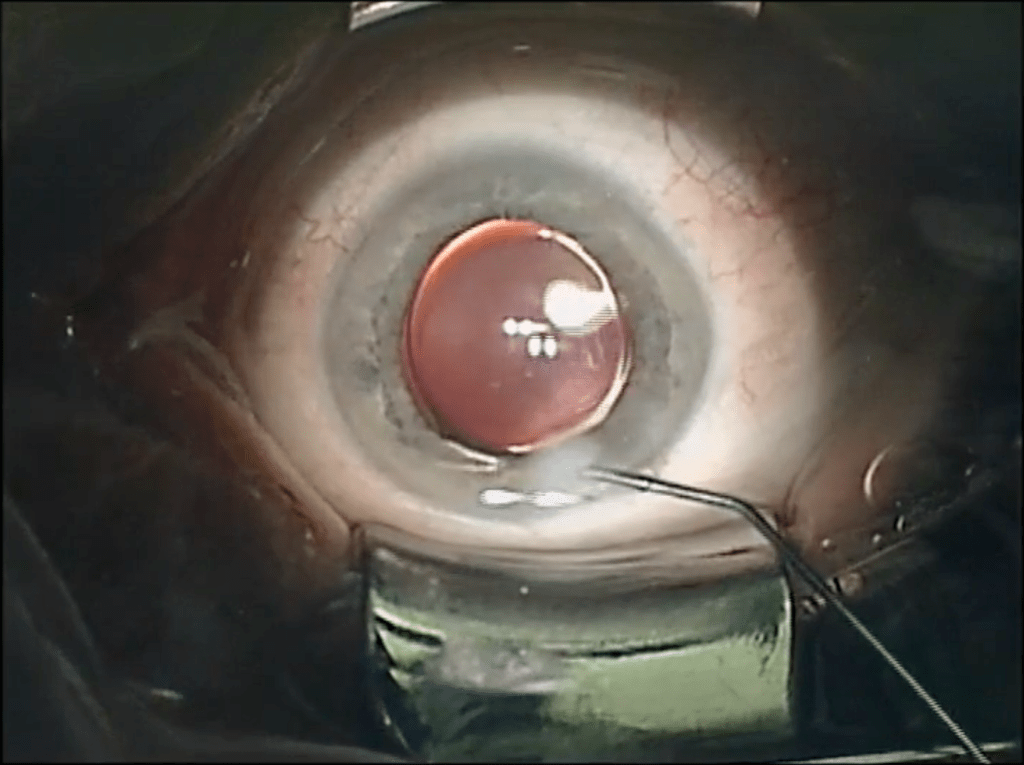}
      \put (3,65) {\footnotesize\textcolor{white}{00:03:5.45}}
     \end{overpic} 
        \caption{Pupil contraction during cataract surgery.}
    \end{subfigure}\\
        \begin{subfigure}[t]{1\textwidth}
      \begin{overpic}[width=0.19\textwidth]{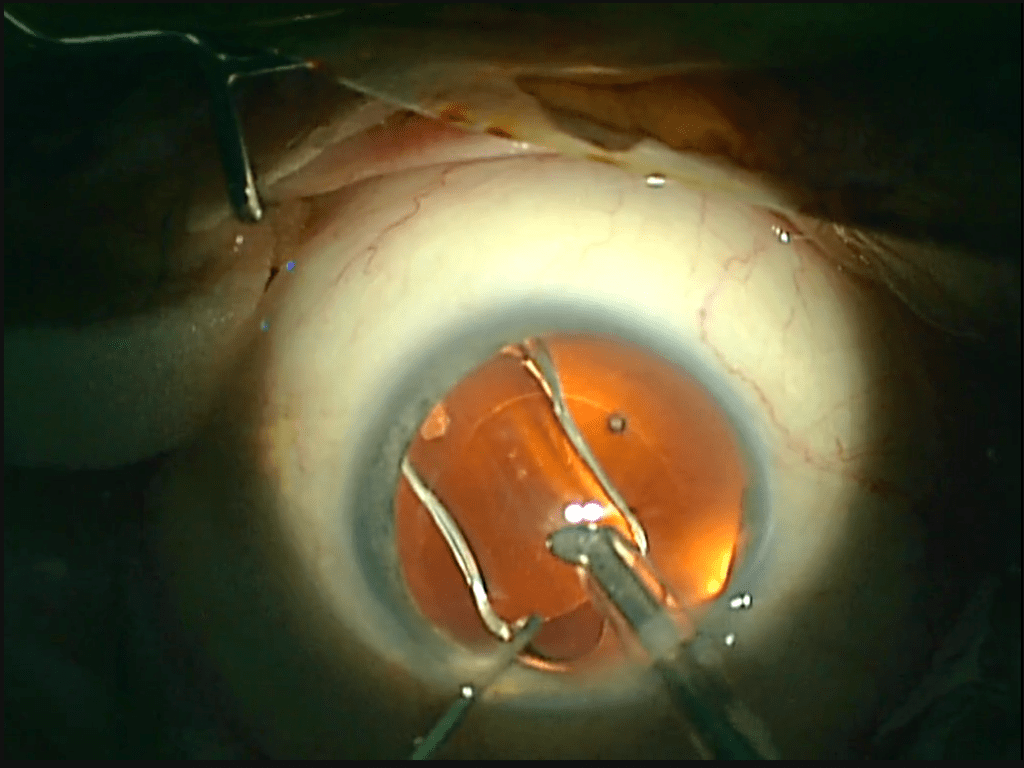}
      \put (3,65) {\footnotesize\textcolor{white}{00:02:50.07}}
     \end{overpic} 
     \begin{overpic}[width=0.19\textwidth]{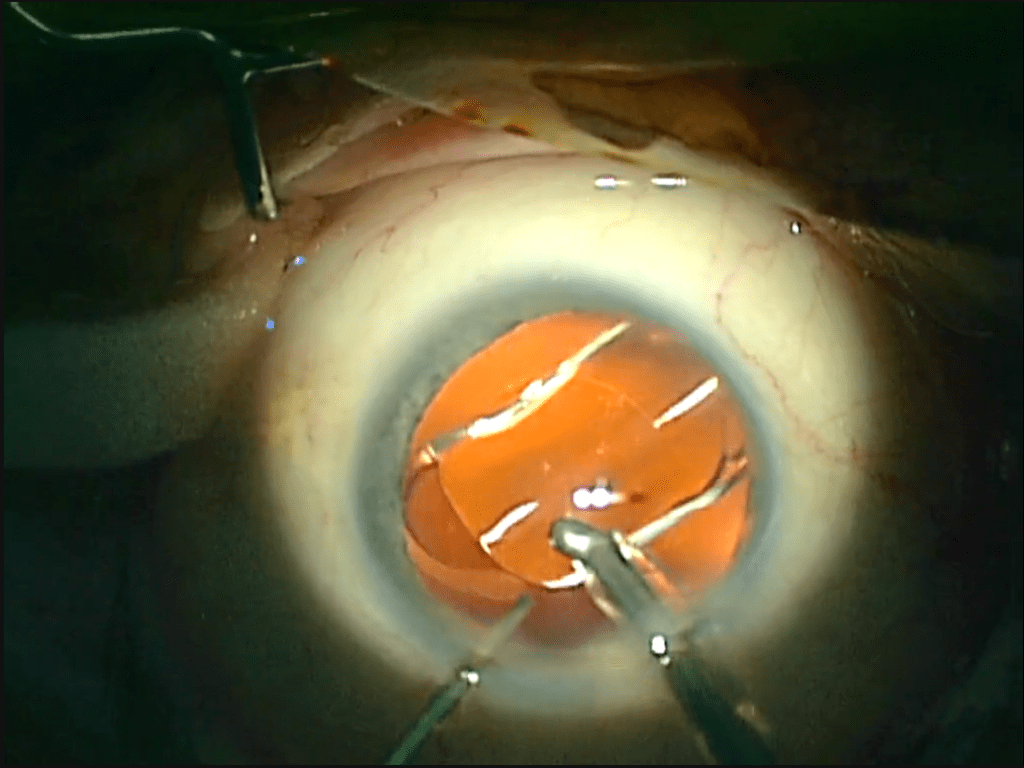}
      \put (3,65) {\footnotesize\textcolor{white}{00:02:53.20}}
     \end{overpic} 
     \begin{overpic}[width=0.19\textwidth]{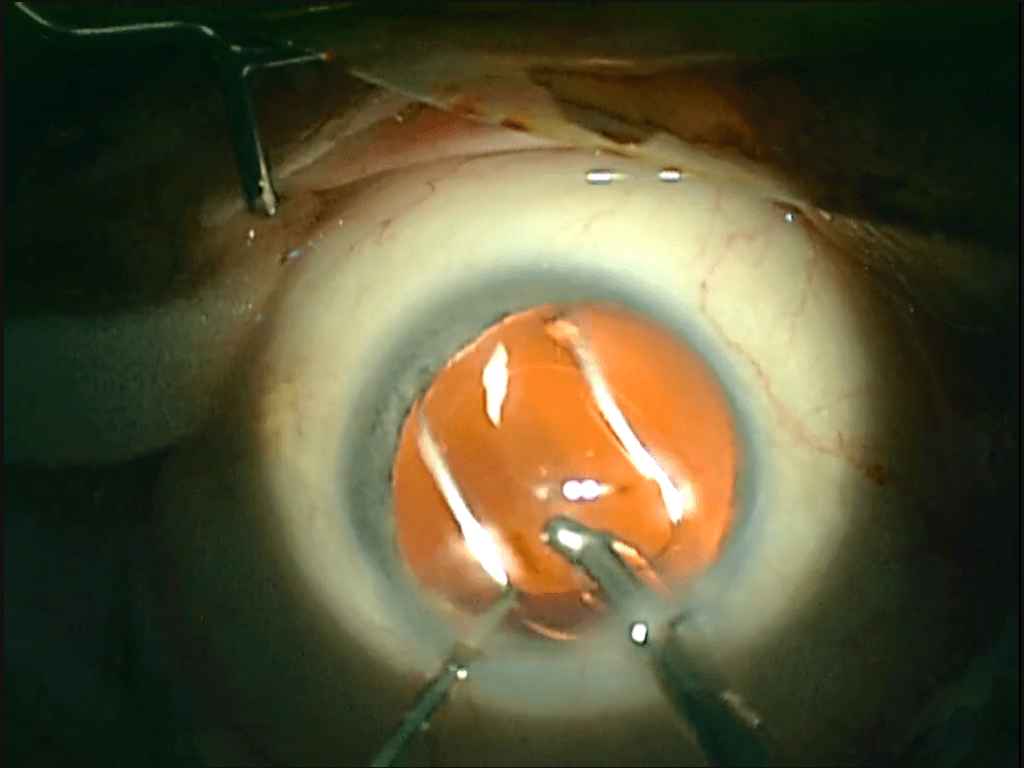}
      \put (3,65) {\footnotesize\textcolor{white}{00:02:53.65}}
     \end{overpic} 
     \begin{overpic}[width=0.19\textwidth]{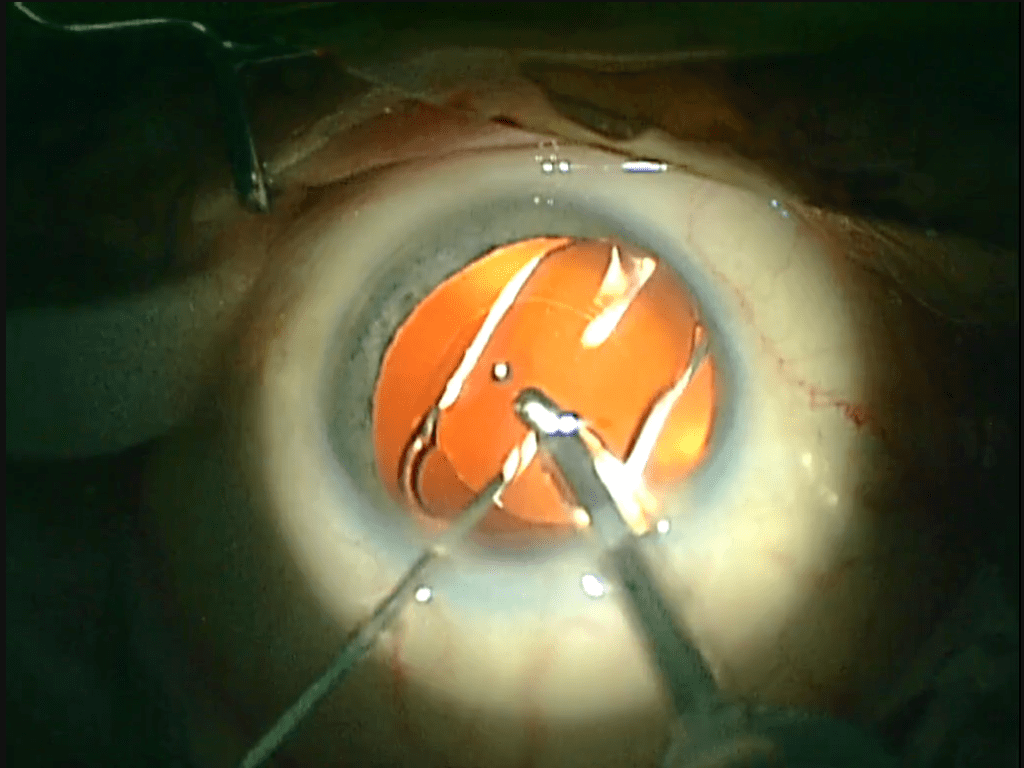}
      \put (3,65) {\footnotesize\textcolor{white}{00:02:54.37}}
     \end{overpic} 
     \begin{overpic}[width=0.19\textwidth]{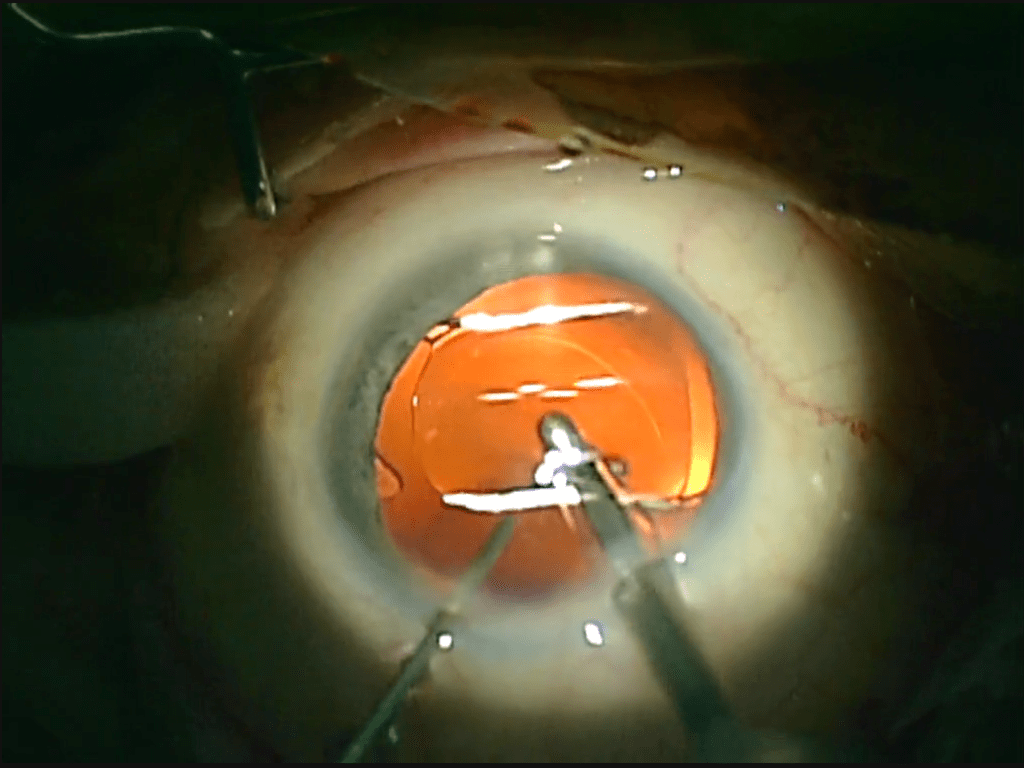}
      \put (3,65) {\footnotesize\textcolor{white}{00:02:56.69}}
     \end{overpic} 
     
        \caption{Sever clockwise IOL rotations.}
    \end{subfigure}\\
    \caption{Intra-operative irregularities in cataract surgery.}
    \label{fig:irregularities}
\end{figure*}

The semantic segmentation dataset includes frames from 30 regular cataract surgery videos with a spatial resolution of $1024 \times 768$. Frame extraction is performed at the rate of one frame per five seconds. Subsequently, the frames featuring very harsh motion blur or out-of-scene iris are excluded from the dataset. We provide pixel-level annotations for three relevant anatomical structures, including the iris, pupil, and intraocular lens, as well as nine instruments used in regular cataract surgeries, including slit/incision knife, gauge, spatula, capsulorhexis cystome, phacoemulsifier tip, irrigation-aspiration, lens injector, capsulorhexis forceps, and katana forceps. All annotations are performed using polygons in the Supervisely platform\footnote{https://supervisely.com/}, and exported as JSON files.
Within this dataset, the included individuals possess an average age of 74.5 years, spanning from 51 to 90 years, with a standard deviation of 8.43 years. Additionally, the videos contained in the phase recognition dataset depict surgeries conducted by surgeons whose collective experience averages 8033 surgeries, with a standard deviation of 3894 surgeries.
The provided dataset enables a reliable study of segmentation performance for relevant anatomical structures, binary instruments, and multi-class instruments.

\paragraph{Irregularity detection dataset. }
This dataset contains two small subsets of major intra-operative irregularities in cataract surgery, including pupil reaction \cite{sokolova2021automatic} and lens rotation \cite{LensID}.
\begin{itemize}
    \item \textit{Pupil Contraction:} During the phacoemulsification phase, where the occluded natural lens is fragmented and suctioned, there exists a heightened risk of causing damage to the delicate iris. Even very subtle trauma to the tissue can lead to undesirable pupil constriction \cite{Mirza2003}. These sudden reactions in pupil size can lead to serious intra-operative implications. Especially during
the phacoemulsification phase, where the instrument is deeply inserted inside the eye, sudden changes in pupil size may lead to
injuries to the eye’s tender tissues. Besides, achieving precise IOL alignment or centration becomes challenging in cases where intraoperative pupil contraction (miosis) occurs. Particularly in multifocal IOLs, minor displacements or tilts, which might be negligible for conventional mono-focal IOLs, can significantly compromise visual performance. In the case of toric IOLs, precise alignment of the torus is crucial, as any deviation diminishes the IOL's effectiveness. Detection of unusual pupil reactions and severe pupil contractions during the surgery can highly contribute to the overall outcomes of cataract surgery and provide important insight for further post-operative investigations. Figure \ref{fig:irregularities}-top demonstrates an example of severe pupil contraction during cataract surgery.
\item \textit{IOL Rotation:} Although aligned and centered upon surgery's conclusion, the IOL may rotate or dislocate following the surgery. Even slight deviations, such as minor misalignments of the torus in toric IOLs or the slight displacement and tilting of multifocal IOLs, can result in significant distortions in vision and leave patients dissatisfied. The sole way to address this postoperative complication is follow-up surgery, which entails added costs, heightened surgical risks, and patient discomfort.
Identification of intra-operative indicators for predicting and preventing post-surgical IOL dislocation is an unmet clinical need.
It is argued that intra-operative rotation of IOLs during cataract surgery is the leading cause of post-operative misalignments~\cite{Oshika2020}. Hence, automatic detection and measurement of intra-operative lens rotations can effectively contribute to preventing post-operative IOL dislocation. Figure \ref{fig:irregularities}-bottom represents fast clockwise rotations of IOL during unfolding, which occur in less than seven seconds.  
\end{itemize}

\subsection*{Experimental Settings for Phase Recognition}
\paragraph{Network Architectures.} We adopt a combined CNN-RNN framework for phase recognition. The CNN component, serving as the backbone model, is responsible for the extraction of distinctive features from individual frames within the video sequence. To achieve this, two different pre-trained CNN architectures, VGG16 and ResNet50, are employed. The output feature map of the CNN is fed into a recurrent neural network (RNN). The RNN component focuses on capturing temporal features from the input video clip. We compare the performance of four different RNN architectures, including LSTM, GRU, BiLSTM, and BiGRU. 

\paragraph{Training Settings.} We adopt a one-versus-rest strategy to evaluate phase recognition performance \cite{LocalPhase,10178763}. Accordingly, we segment all videos corresponding to each phase into three-second clips with an overlap of one second. Afterward, the entire dataset is split into two categories: the designated target phase and the remaining phases (the "rest" class). We apply offline augmentations to the videos across all categories. Typically, the number of clips in the target category is significantly lower than in the rest category. To rectify this imbalance problem, we employ a random selection process from the "rest" category, aligning it with the clip count in the target category. This strategy ensures an equivalent number of clips in both classes. 
The employed augmentations include gamma and contrast adjustments with a factor of 0.5, Gaussian blur with a sigma of 10, random rotation up to 20 degrees, brightness within a range of $[-0.3, 0.3]$, and saturation within a range of $[0.5, 1.5]$.
To maximize diversity within our training set, we employ a random sampling strategy during training. Specifically, we configure the network's input sequence to comprise 10 frames randomly selected from 90 frames within each three-second clip. In all settings, the backbone network employed for feature extraction is pre-trained on the ImageNet dataset. The RNN component is constructed with a single recurrent layer comprising 64 units. This is followed by a dense layer with 64 units, and finally, a two-unit layer with a Softmax activation function. To mitigate the risk of overfitting, the last four layers of the CNN component are kept frozen during training, and dropout regularization with a rate of 0.5 is applied to the output feature map of the recurrent layer.  All models are trained on 32 videos and tested on non-overlapping clips from the remaining videos. We use a binary cross-entropy loss function and Adam optimizer, a learning rate equal to 0.001, and a batch size of 16. The network's input dimensions are set to $224 \times 224$. We compare the performance of the trained models using accuracy and F1 score. 

\begin{table}[b!]
\renewcommand{\arraystretch}{1}
\caption{Specifications of the proposed and alternative approaches. In ``Upsampling" column, ``Trans Conv" stands for \textit{Transposed Convolution}.}

\label{tab:alternatives}
\centering
\resizebox{0.85\textwidth}{!}{%
\begin{tabular}{lccccc}
\specialrule{.12em}{.05em}{.05em}
Model & Backbone & Params. & Upsampling & Target & Reference\\\specialrule{.12em}{.05em}{.05em}
DeepPyramid &VGG16 & 33.57 M & Bilinear & Medical Images & \cite{ghamsarian2022deeppyramid}\\
Adapt-Net & VGG16 & 24.69 M & Bilinear & Medical Images & \cite{LensID}\\
UNet$++$~&VGG16&24.24 M& Bilinear & Medical Images & \cite{UNet++}\\
ReCal-Net & VGG16 & 22.93 M & Bilinear & Medical Images & \cite{ReCal-Net}\\
CPFNet & VGG16 $\vert$ ResNet34 & 39.17 M $\vert$ 34.66 M& Bilinear & Medical Images & \cite{CPFNet}\\
CE-Net &VGG16 $\vert$ ResNet34& 33.50 M $\vert$ 29.90 M& Trans Conv & Medical Images &  \cite{CE-Net}\\
FED-Net&ResNet50&59.52 M& Trans Conv \& PixelShuffle & Liver Lesion & \cite{FED-Net}\\
scSENet & VGG16 $\vert$ ResNet34& 22.90 M $\vert$ 25.25 M& Bilinear & Medical Images & \cite{SCSE}\\
DeepLabV3+& ResNet50& 26.68 M&Bilinear&Scene& \cite{DeepLabV3}\\
UPerNet & ResNet50& 51.26 M&Bilinear&Scene& \cite{UPerNet}\\
U-Net+\footnote{Note that UNet+ is an improved version of UNet, where we use VGG16 as the backbone network and double convolutional blocks (two consecutive convolutions followed by batch normalization and ReLU layers) as decoder modules.}& VGG16 &22.55 M& Bilinear & Medical Images & \cite{U-Net}\\
\specialrule{.12em}{.05em}{0.05em}
\end{tabular}
}
\end{table} 

\subsection*{Experimental Settings for Semantic Segmentation}
\paragraph{Network Architectures.} We perform experiments to validate the robustness of our pixel-level annotations using several state-of-the-art baselines targetting general images, medical images, and surgical videos. The specifications of the baselines are listed in Table~\ref{tab:alternatives}.

\paragraph{Training Settings.} For all neural networks, the backbones are initialized with ImageNet's pre-trained parameters \cite{ImageNet}. We train all networks with a batch size of eight and set the initial learning rate to 0.001, which decreases during training using polynomial decay $lr = lr_{init}\times (1-\frac{iter}{total-iter})^{0.9}$. The input size of the networks is set to $512\times 512$. We apply cropping and random rotation (up to 30 degrees), color jittering (brightness = 0.7, contrast = 0.7, saturation = 0.7), Gaussian blurring, and random sharpening as augmentations during training, and use the \textit{cross entropy log dice} loss during training as in eq. \eqref{eq: loss},

\begin{equation}
    \mathcal{L} = (\lambda)\times BCE(\mathcal{X}_{true}(i,j),\mathcal{X}_{pred}(i,j))
    -(1-\lambda)\times (\log \frac{2\sum \mathcal{X}_{true}\odot \mathcal{X}_{pred}+\sigma}{\sum \mathcal{X}_{true} + \sum \mathcal{X}_{pred}+ \sigma}),
\label{eq: loss}
\end{equation}

\noindent where $\mathcal{X}_{true}$ denotes the ground truth binary mask, and $\mathcal{X}_{pred}$ denotes the predicted mask ($0\leq \mathcal{X}_{pred}(i,j) \leq 1$). The parameter $\lambda \in [0,1]$ is set to $0.8$ in our experiments, and $\odot$ refers to the Hadamard product (element-wise multiplication). Besides, the parameter $\sigma$ is the Laplacian smoothing factor, which is added to (i) prevent division by zero and (ii) avoid overfitting (in experiments, $\sigma = 1$). 
We compare the performance of baselines using average dice and average intersection over union (IoU).

\begin{table}[bt!]
\centering
\caption{Number of instances and presence in the frames (\% of total number of frames in each fold).}
\label{tab:segmentation-statistics-presence}
\resizebox{1\textwidth}{!}{%

\begin{tabular}{lb{4cm}*{6}{>{\arraybackslash}b{2cm}}}
\specialrule{.12em}{.05em}{.05em}
Category & Class Name & All Videos & Fold1 & Fold2 & Fold3 & Fold4 & Fold5 \\\midrule
\multirow{3}{*}{Anatomy}&Iris&\DCB{2256}{100.0}&\DCB{561}{100.0}&\DCB{459}{100.0}&\DCB{420}{100.0}&\DCB{385}{100.0}&\DCB{431}{100.0}\\&Pupil&\DCB{2256}{100.0}&\DCB{561}{100.0}&\DCB{459}{100.0}&\DCB{420}{100.0}&\DCB{385}{100.0}&\DCB{431}{100.0}\\&Intraocular Lens&\DCB{537}{23.8}&\DCB{107}{19.07}&\DCB{119}{25.93}&\DCB{102}{24.29}&\DCB{106}{27.53}&\DCB{103}{23.9}\\\midrule\multirow{10}{*}{Instruments}&Slit/Incision Knife&\DCB{50}{2.22}&\DCB{12}{2.14}&\DCB{10}{2.18}&\DCB{12}{2.86}&\DCB{4}{1.04}&\DCB{12}{2.78}\\&Gauge&\DCB{426}{18.88}&\DCB{103}{18.36}&\DCB{90}{19.61}&\DCB{79}{18.81}&\DCB{76}{19.74}&\DCB{78}{18.1}\\&Spatula&\DCB{728}{32.27}&\DCB{214}{38.15}&\DCB{132}{28.76}&\DCB{148}{35.24}&\DCB{105}{27.27}&\DCB{129}{29.93}\\&Capsulorhexis Cystotome&\DCB{85}{3.77}&\DCB{20}{3.57}&\DCB{18}{3.92}&\DCB{12}{2.86}&\DCB{11}{2.86}&\DCB{24}{5.57}\\&Phacoemulsifier Tip&\DCB{547}{24.25}&\DCB{148}{26.38}&\DCB{91}{19.83}&\DCB{101}{24.05}&\DCB{95}{24.68}&\DCB{112}{25.99}\\&Irrigation-Aspiration&\DCB{456}{20.21}&\DCB{122}{21.75}&\DCB{91}{19.83}&\DCB{98}{23.33}&\DCB{71}{18.44}&\DCB{74}{17.17}\\&Lens Injector&\DCB{66}{2.93}&\DCB{14}{2.5}&\DCB{11}{2.4}&\DCB{14}{3.33}&\DCB{13}{3.38}&\DCB{14}{3.25}\\&Capsulorhexis Forceps&\DCB{108}{4.79}&\DCB{33}{5.88}&\DCB{21}{4.58}&\DCB{22}{5.24}&\DCB{21}{5.45}&\DCB{11}{2.55}\\&Katena Forceps&\DCB{29}{1.29}&\DCB{8}{1.43}&\DCB{3}{0.65}&\DCB{8}{1.9}&\DCB{3}{0.78}&\DCB{7}{1.62}\\&All&\DCB{1778}{78.81}&\DCB{462}{82.35}&\DCB{345}{75.16}&\DCB{344}{81.9}&\DCB{296}{76.88}&\DCB{331}{76.8}\\
\specialrule{.12em}{.05em}{.05em}
\end{tabular}
}
\end{table}

\begin{table}[bt!]
\centering
\caption{Average pixels corresponding to different labels per frame.}
\label{tab:segmentation-statistics-average-pixels}
\resizebox{1\textwidth}{!}{%

\begin{tabular}{lb{4cm}*{6}{>{\arraybackslash}b{2cm}}}
\specialrule{.12em}{.05em}{.05em}
Category & Class Name & All Videos & Fold1 & Fold2 & Fold3 & Fold4 & Fold5 \\\midrule
\multirow{3}{*}{Anatomy}&Iris&45939&41874&47792&44867&47963&48494\\&Pupil&36013&38594&33578&35900&35291&35999\\&Intraocular Lens&9135&7056&10017&9153&10405&9748\\\midrule\multirow{10}{*}{Instruments}&Slit/Incision Knife&1140&1088&1179&1163&1206&1086\\&Gauge&299&222&337&454&168&326\\&Spatula&2613&3163&2078&2893&2309&2466\\&Capsulorhexis Cystotome&5523&4760&4773&6551&6345&5580\\&Phacoemulsifier Tip&5230&4388&5526&7646&4451&4354\\&Irrigation-Aspiration&1083&790&1138&1311&1153&1123\\&Lens Injector&512&465&543&673&318&556\\&Capsulorhexis Forceps&172&288&104&225&23&176\\&Katena Forceps&823&906&678&1065&1133&357\\&All&17397&16069&16357&21981&17105&16025\\
\specialrule{.12em}{.05em}{.05em}
\end{tabular}
}
\end{table}

\section*{Data Records}

All datasets and annotations will be publicly released in Synapse upon the acceptance of the paper (accessible for anonymous review in Figshare).

Frame-level annotations for phase recognition are provided in CSV files, determining the first and the last frames for all action phases per video. The preprocessing codes to extract all action and idle phases from a video using the CSV files are provided in the GitHub repository of the paper. Table \ref{tab:phase-annotations} visualizes our phase annotations for 56 cataract surgery videos. Furthermore, Figure \ref{fig:phase_statistics_pie} demonstrates the total duration of the annotations corresponding to each phase from 56 videos.

Pixel-level annotations are provided in two formats: (1) Supervisely format, for which we provide Python codes for mask creation from JSON files, and (2) COCO format, which also provides bounding box annotations for all pixel-level annotated objects. The latter annotations can be used for object localization problems. The preprocessing codes to create training masks for "anatomy plus instrument segmentation", "binary instrument segmentation", and "multi-class instrument segmentation" are provided in the GitHub repository of the paper. We have formed five folds with patient-wise separation, meaning every fold consists of the frames corresponding to six distinct videos. Table \ref{tab:segmentation-statistics-presence} compares the number of instances and their appearance percentage in the frames. Besides, Table \ref{tab:segmentation-statistics-average-pixels} lists the average number of pixels per frame corresponding to each label.

\section*{Technical Validation}

Table \ref{tab:phase-recognition} showcases the phase recognition performance of several CNN-RNN architectures. In our evaluations, we have combined the phases of viscoelastic and anterior-chamber flushing due to their shared visual features. The collective findings reveal commendable and satisfactory phase recognition performance across diverse backbones and recurrent network setups. Notably, the incorporation of bidirectional recurrent layers has consistently amplified detection accuracy and F1-Score across all configurations.

\begin{table}[b!]
\centering
\caption{Phase recognition performance of several CNN-RNN architectures.}
\label{tab:phase-recognition}
\resizebox{1\textwidth}{!}{%

\begin{tabular}{l*{8}{>{\centering\arraybackslash}m{2cm}}}
\specialrule{.12em}{.05em}{.05em}
Network & \multicolumn{2}{c}{\footnotesize{ResNet50-LSTM}} & \multicolumn{2}{c}{\footnotesize{ResNet50-GRU}} & \multicolumn{2}{c}{\footnotesize{ResNet50-BiLSTM}} & \multicolumn{2}{c}{\footnotesize{ResNet50-BiGRU}} \\ \cmidrule(lr){2-3}\cmidrule(lr){4-5}\cmidrule(lr){6-7}\cmidrule(lr){8-9}
Phases &  Accuracy (\%) & F1-Score (\%) & Accuracy (\%) & F1-Score (\%) & Accuracy (\%) & F1-Score (\%) & Accuracy (\%) & F1-Score (\%) \\ \specialrule{.12em}{.05em}{.05em}
Incision& \NP{83.35}& \NP{81.48}& \NP{83.33}& \NP{83.31}& \NP{86.67}& \NP{86.61}& \NP{86.67}& \NP{86.61}\\Viscoelastic/AC Flushing& \NP{65.12}& \NP{64.81}& \NP{69.77}& \NP{69.17}& \NP{62.21}& \NP{59.43}& \NP{60.47}& \NP{57.42}\\Capsulorhexis& \NP{85.71}& \NP{85.54}& \NP{86.05}& \NP{85.85}& \NP{90.14}& \NP{90.09}& \NP{90.82}& \NP{90.81}\\Hydrodissection& \NP{88.27}& \NP{88.22}& \NP{86.42}& \NP{86.16}& \NP{88.89}& \NP{88.81}& \NP{87.04}& \NP{86.86}\\Phacoemulsification& \NP{95.17}& \NP{95.17}& \NP{94.16}& \NP{94.16}& \NP{95.17}& \NP{95.37}& \NP{94.67}& \NP{94.67}\\Irrigation-Aspiration& \NP{89.91}& \NP{89.88}& \NP{87.39}& \NP{87.32}& \NP{86.47}& \NP{86.44}& \NP{87.84}& \NP{87.82}\\Capsule Polishing& \NP{86.17}& \NP{85.90}& \NP{81.91}& \NP{81.66}& \NP{88.30}& \NP{88.30}& \NP{87.23}& \NP{87.21}\\Lens Implantation& \NP{85.14}& \NP{84.80}& \NP{81.08}& \NP{80.38}& \NP{86.49}& \NP{86.24}& \NP{90.54}& \NP{90.50}\\Lens Positioning& \NP{87.50}& \NP{87.49}& \NP{87.50}& \NP{87.45}& \NP{92.19}& \NP{92.14}& \NP{89.06}& \NP{89.00}\\Viscoelastic-Suction& \NP{91.73}& \NP{91.72}& \NP{89.10}& \NP{89.03}& \NP{90.23}& \NP{90.15}& \NP{90.98}& \NP{90.97}\\Tonifying/Antibiotics& \NP{85.83}& \NP{85.75}& \NP{81.67}& \NP{81.33}& \NP{86.67}& \NP{86.66}& \NP{88.33}& \NP{88.30}\\
\specialrule{.12em}{.05em}{.05em}
Network & \multicolumn{2}{c}{\footnotesize{VGG16-LSTM}} & \multicolumn{2}{c}{\footnotesize{VGG16-GRU}} & \multicolumn{2}{c}{\footnotesize{VGG16-BiLSTM}} & \multicolumn{2}{c}{\footnotesize{VGG16-BiGRU}} \\ \cmidrule(lr){2-3}\cmidrule(lr){4-5}\cmidrule(lr){6-7}\cmidrule(lr){8-9}
Phases &  Accuracy (\%) & F1-Score (\%) & Accuracy (\%) & F1-Score (\%) & Accuracy (\%) & F1-Score (\%) & Accuracy (\%) & F1-Score (\%) \\ \specialrule{.12em}{.05em}{.05em}
Incision& \NP{83.33}& \NP{82.86}& \NP{86.67}& \NP{86.43}& \NP{86.67}& \NP{86.43}& \NP{90.00}& \NP{89.90}\\Viscoelastic/AC Flushing& \NP{64.53}& \NP{63.18}& \NP{63.37}& \NP{62.30}& \NP{64.53}& \NP{63.89}& \NP{66.28}& \NP{64.91}\\Capsulorhexis& \NP{87.07}& \NP{87.04}& \NP{87.76}& \NP{87.73}& \NP{88.44}& \NP{88.42}& \NP{89.80}& \NP{89.79}\\Hydrodissection& \NP{86.42}& \NP{86.42}& \NP{85.23}& \NP{85.79}& \NP{87.89}& \NP{88.89}& \NP{87.04}& \NP{87.02}\\Phacoemulsification& \NP{93.86}& \NP{93.86}& \NP{93.36}& \NP{93.36}& \NP{93.26}& \NP{93.26}& \NP{92.86}& \NP{92.86}\\Irrigation-Aspiration& \NP{86.70}& \NP{86.55}& \NP{86.47}& \NP{86.31}& \NP{88.53}& \NP{88.51}& \NP{88.53}& \NP{88.48}\\Capsule Polishing& \NP{87.23}& \NP{87.23}& \NP{86.17}& \NP{85.90}& \NP{90.43}& \NP{90.37}& \NP{88.30}& \NP{88.14}\\Lens Implantation& \NP{82.43}& \NP{82.04}& \NP{86.49}& \NP{86.24}& \NP{83.78}& \NP{83.35}& \NP{85.14}& \NP{85.00}\\Lens Positioning& \NP{85.94}& \NP{85.93}& \NP{82.81}& \NP{82.47}& \NP{87.50}& \NP{87.49}& \NP{84.38}& \NP{84.13}\\Viscoelastic-Suction& \NP{81.95}& \NP{81.72}& \NP{78.95}& \NP{78.24}& \NP{82.71}& \NP{82.31}& \NP{82.33}& \NP{81.90}\\Tonifying/Antibiotics& \NP{82.50}& \NP{82.50}& \NP{83.33}& \NP{83.33}& \NP{81.67}& \NP{81.33}& \NP{85.00}& \NP{84.96}\\
\specialrule{.12em}{.05em}{.05em}
\end{tabular}
}
\end{table}

\begin{table}[b!]
\centering
\caption{Quantitative evaluations of "anatomy plus instrument" segmentation performance for neural network architectures listed in Table \ref{tab:alternatives}.}
\label{tab:quantitative-anatomy-inst}
\resizebox{1\textwidth}{!}{%

\begin{tabular}{lm{1.6cm}*{9}{>{\centering\arraybackslash}m{1.8cm}}}
\specialrule{.12em}{.05em}{.05em}
Object && \multicolumn{2}{c}{\footnotesize{Iris}} & \multicolumn{2}{c}{\footnotesize{Pupil}} & \multicolumn{2}{c}{\footnotesize{Lens}} & \multicolumn{2}{c}{\footnotesize{Instruments}} \\ \cmidrule(lr){3-4}\cmidrule(lr){5-6}\cmidrule(lr){7-8}\cmidrule(lr){9-10}
Backbone &  Network & IoU (\%) & Dice (\%) & IoU (\%) & Dice (\%) & IoU (\%) & Dice (\%) & IoU (\%) & Dice (\%) \\ \specialrule{.12em}{.05em}{.05em}
\multirow{9}{*}{VGG16} & UNet+ & \NP{85.43} & \NP{92.13} & \NP{89.50} & \NP{94.46} & \NP{79.39} & \NP{88.47} & \NP{63.37} & \NP{77.56}\\
 & scSENet & \NP{85.52} & \NP{92.18} & \NP{89.34} & \NP{94.37} & \NP{78.89} & \NP{88.18} & \NP{63.53} & \NP{77.68}\\
 & FEDNet & \NP{84.09} & \NP{91.34} & \NP{88.25} & \NP{93.76} & \NP{77.93} & \NP{87.56} & \NP{60.12} & \NP{75.07}\\
 & CE-Net & \NP{82.98} & \NP{90.68} & \NP{86.54} & \NP{92.78} & \NP{73.40} & \NP{84.53} & \NP{56.50} & \NP{72.17}\\
 & CPFNet & \NP{85.10} & \NP{91.93} & \NP{89.50} & \NP{94.45} & \NP{80.66} & \NP{89.26} & \NP{62.90} & \NP{77.21}\\
 & UNetPP & \NP{85.20} & \NP{91.99} & \NP{89.46} & \NP{94.43} & \NP{79.07} & \NP{88.26} & \NP{63.63} & \NP{77.76}\\
 & AdaptNet & \NP{85.50} & \NP{92.18} & \NP{90.29} & \NP{94.89} & \NP{83.02} & \NP{90.70} & \NP{62.18} & \NP{76.68}\\
 & ReCal-Net & \NP{85.33} & \NP{92.07} & \NP{90.19} & \NP{94.84} & \NP{82.53} & \NP{90.43} & \NP{62.55} & \NP{76.95}\\ 
 & DeepPyramid & \NP{86.10} & \NP{92.52} & \NP{90.61} & \NP{95.07} & \NP{83.72} & \NP{91.11} & \NP{63.91} & \NP{77.98}\\ \midrule
\multirow{4}{*}{ResNet34} 
         & scSENet & \NP{84.77} & \NP{91.74} & \NP{87.72} & \NP{93.42} & \NP{65.31} & \NP{76.02} & \NP{27.11} & \NP{34.64}\\
 & FEDNet & \NP{81.83} & \NP{89.99} & \NP{84.13} & \NP{91.35} & \NP{51.75} & \NP{65.61} & \NP{53.51} & \NP{69.69}\\
 & CE-Net & \NP{76.52} & \NP{86.43} & \NP{84.84} & \NP{91.80} & \NP{66.21} & \NP{79.48} & \NP{30.88} & \NP{43.86}\\
 & CPFNet &  \NP{83.59} & \NP{91.05} & \NP{88.74} & \NP{94.03} & \NP{81.24} & \NP{89.63} & \NP{61.71} & \NP{76.31}\\ \midrule
\multirow{2}{*}{ResNet50}
 & UPerNet & \NP{85.15} & \NP{91.97} & \NP{90.03} & \NP{94.75} & \NP{83.72} & \NP{91.11} & \NP{63.48} & \NP{77.65}\\

 & DeepLabV3+ & \NP{79.97} & \NP{88.83} & \NP{86.24} & \NP{92.61} & \NP{77.23} & \NP{87.12} & \NP{53.53}& \NP{69.67}\\

\specialrule{.12em}{.05em}{.05em}
\end{tabular}
}
\end{table}

Furthermore, networks leveraging the ResNet50 backbone display marginally superior performance compared to those utilizing VGG16. This outcome can be attributed to the deeper architecture of ResNet50, facilitating the extraction of intricate features essential for accurate recognition.  
The results also reveal the distinguishability of different phases in cataract surgery. Precisely, the phacoemulsification phase consistently attains the highest accuracy and F1 score, attributed to the distinctive phacoemulsification instrument and the unique texture of the pupil during this phase. Conversely, the least robust detection performance aligns with the viscoelastic/AC flushing phases, accentuating the visual resemblances shared between these phases and other phases within cataract surgery videos.

\begin{figure*}[t!]
\begin{tabular}{c}
\begin{subfigure}{0.8\textwidth}\vspace{-0.2\baselineskip}
  \centering
  \begin{adjustbox}{width=1\textwidth}
  \begin{tikzpicture}
\begin{axis}[
ybar=0pt,
axis on top,
bar width=0.6cm,
width=\textwidth,
height=.4\textwidth,
xlabel={\ VGG16},
enlarge x limits=0.08,
every axis x label/.append style={xshift=0.2cm},
symbolic x coords={UNet+, scSENet, FEDNet, CE-Net, CPFNet, UNet++, AdaptNet, ReCal-Net, DeepPyramid},
xtick=data,
nodes near coords align={vertical},
cycle list/Set2,
ymin=60,ymax=110,
ymajorgrids = true,
yminorgrids = true,
bar width=0.5cm,
every node near coord/.append style={yshift=+4pt,xshift=-5pt,anchor=east,font=\scriptsize},
nodes near coords,
minor ytick={70,90},
ticklabel style = {font=\scriptsize},
major x tick style = transparent,
x tick label style={rotate=25,anchor=east},
legend style={at={(0.5,+1.18)},
	            anchor=north,legend columns=2},
every node near coord/.append style={rotate=90, anchor=west, font=\scriptsize, color = black, opacity=1}	            
]
\addplot+[style={draw=black,solid,fill,opacity=0.6,
}, 
             error bars/.cd, 
             y dir=both,y explicit]
             coordinates {
                  (UNet+,79.42) +- (0,1.48)
                  (scSENet,79.32) +- (0,1.33)
                  (FEDNet,77.60) +- (0,1.25)
                  (CE-Net,74.86) +- (0,2.23)
                  (CPFNet,79.52) +- (0,1.56)
                  (UNet++,79.34) +- (0,1.61)
                  (AdaptNet,80.25) +- (0,1.42)
                  (ReCal-Net,80.15) +- (0,1.08)
                  (DeepPyramid,81.07) +- (0,1.34)
                  }; 
                  
\addplot+[style={draw=black,solid,fill,opacity=0.6,
}, 
             error bars/.cd, 
             y dir=both,y explicit]
             coordinates {
                  (UNet+,88.16) +- (0,0.91)
                  (scSENet,88.10) +- (0,0.83)
                  (FEDNet,86.93) +- (0,0.78)
                  (CE-Net,85.04) +- (0,1.48)
                  (CPFNet,88.21) +- (0,0.95)
                  (UNet++,88.11) +- (0,0.99)
                  (AdaptNet,88.61) +- (0,0.87)
                  (ReCal-Net,88.57) +- (0,0.67)
                  (DeepPyramid,89.17) +- (0,0.80)
                  }; 
\legend{IoU(\%),Dice(\%)}
\end{axis}
\end{tikzpicture}

\end{adjustbox}
\end{subfigure}\\

\begin{subfigure}{0.44\textwidth}\vspace{-0.2\baselineskip}
  \centering
  \begin{adjustbox}{width=1\textwidth}
  \begin{tikzpicture}
\begin{axis}[
ybar=0pt,
axis on top,
width=\textwidth,
height=.6\textwidth,
xlabel={\ ResNet34},
enlarge x limits=0.25,
every axis x label/.append style={xshift=0.2cm},
symbolic x coords={scSENet, FEDNet, CE-Net, CPFNet},
xtick=data,
nodes near coords align={vertical},
every node near coord/.append style={yshift=+4pt,xshift=-5pt,anchor=east,font=\scriptsize},
cycle list/Set2,
ymin=50,ymax=110,
ymajorgrids = true,
yminorgrids = true,
bar width=0.5cm,
nodes near coords,
minor ytick={70,90},
ticklabel style = {font=\scriptsize},
major x tick style = transparent,
x tick label style={rotate=25,anchor=east},
legend style={at={(0.5,+1.15)},
	            anchor=north,legend columns=2},
every node near coord/.append style={rotate=90, anchor=west, font=\scriptsize, color = black, opacity=1}	            
]
\addplot+[style={draw=black,solid,fill,opacity=0.6,
}, 
             error bars/.cd, 
             y dir=both,y explicit]
             coordinates {
                  
                  (scSENet,66.23) +- (0,12.60)
                  (FEDNet,67.80) +- (0,5.84)
                  (CE-Net,64.61) +- (0,7.29)
                  (CPFNet,78.82) +- (0,1.53)
                  
                  }; 
                  
\addplot+[style={draw=black,solid,fill,opacity=0.6,
}, 
             error bars/.cd, 
             y dir=both,y explicit]
             coordinates {
                  
                  (scSENet,73.96) +- (0,12.94)
                  (FEDNet,79.16) +- (0,5.61)
                  (CE-Net,75.39) +- (0,7.97)
                  (CPFNet,87.76) +- (0,0.72)
                  
                  }; 
\end{axis}
\end{tikzpicture}

\end{adjustbox}
\end{subfigure}

\begin{subfigure}{0.35\textwidth}\vspace{-0.2\baselineskip}
  \centering
  \begin{adjustbox}{width=1\textwidth}
  \begin{tikzpicture}
\begin{axis}[
ybar=0pt,
axis on top,
width=\textwidth,
height=.69\textwidth,
xlabel={\ ResNet50},
enlarge x limits=0.8,
every axis x label/.append style={xshift=0.2cm},
symbolic x coords={UPerNet, DeepLabV3+},
xtick=data,
nodes near coords align={vertical},
cycle list/Set2,
ymin=50,ymax=110,
ymajorgrids = true,
yminorgrids = true,
bar width=0.5cm,
every node near coord/.append style={yshift=+4pt,xshift=-5pt,anchor=east,font=\scriptsize},
nodes near coords,
minor ytick={70,90},
ticklabel style = {font=\scriptsize},
major x tick style = transparent,
x tick label style={rotate=25,anchor=east},
legend style={at={(0.5,+1.15)},
	            anchor=north,legend columns=2},
every node near coord/.append style={rotate=90, anchor=west, font=\scriptsize, color = black, opacity=1}	            
]
\addplot+[style={draw=black,solid,fill,opacity=0.6,
}, 
             error bars/.cd, 
             y dir=both,y explicit]
             coordinates {
                  
                  (UPerNet,80.59) +- (0,1.24)
                  (DeepLabV3+,74.24) +- (0,2.19)

                  }; 
                  
\addplot+[style={draw=black,solid,fill,opacity=0.6,
}, 
             error bars/.cd, 
             y dir=both,y explicit]
             coordinates {
                  
                  (UPerNet,88.87) +- (0,0.76)
                  (DeepLabV3+,84.56) +- (0,1.51)

                  }; 
\end{axis}
\end{tikzpicture}

\end{adjustbox}
\end{subfigure}

\end{tabular}
\caption{Average and standard deviation of "anatomy plus instrument" segmentation results 
for neural network architectures listed in Table \ref{tab:alternatives}.}
\label{fig:Dice-IoU}
\end{figure*}
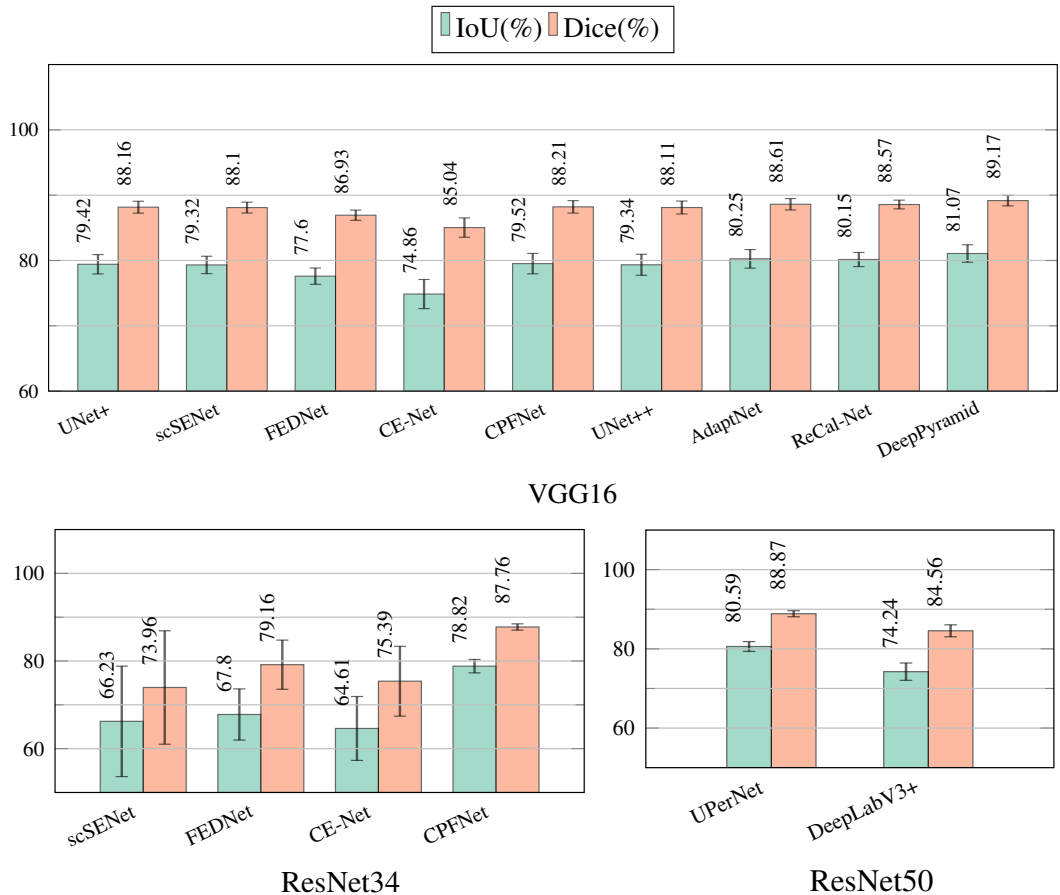

Table \ref{tab:quantitative-anatomy-inst} provides a quantitative analysis of "anatomy plus instrument" segmentation performance for various neural network architectures. The results notably highlight that segmenting the relevant anatomical structures emerges as a comparatively less challenging task than instrument segmentation for all networks. Specifically, the best performance corresponds to pupil segmentation, attributable to its distinct features and sharp boundaries. In contrast, lens segmentation demonstrates relatively lower performance due to its transparent nature and an inherent imbalance issue (outlined in Table \ref{tab:segmentation-statistics-presence}). The segments involving instruments, however, confront significant challenges. This class is marked by major distortions, encompassing motion blur, reflections, and occlusions, collectively contributing to the relatively low performance of the networks. The best performance corresponds to the DeepPyramid network with a VGG16 backbone, consistently yielding optimal results across all classes. 

Figure \ref{fig:Dice-IoU} visually compares the Dice and IoU metrics' averages and standard deviations across five folds for the evaluated neural networks.  According to the results, DeepPyramid, AdaptNet, and ReCal-Net are the three best-performing networks for anatomy and instrument segmentation in cataract surgery videos.

\begin{table}[t!]
\centering
\caption{Single domain and cross-domain binary instrument segmentation performance for neural network architectures listed in Table \ref{tab:alternatives}.}
\label{tab:quantitative-single-cross-domain}
\resizebox{0.9   \textwidth}{!}{%

\begin{tabular}{lm{1.6cm}*{4}{>{\centering\arraybackslash}m{2.8cm}}}
\specialrule{.12em}{.05em}{.05em}
Domain && \multicolumn{2}{c}{\footnotesize{Source (Cataract-1K)}} & \multicolumn{2}{c}{\footnotesize{Target (CaDIS)}}  \\ \cmidrule(lr){3-4}\cmidrule(lr){5-6}
Backbone &  Network & IoU (\%) & Dice (\%) & IoU (\%) & Dice (\%) \\ \specialrule{.12em}{.05em}{.05em}
\multirow{9}{*}{VGG16}&UNet+ &\NSP{71.58}{3.03}&\NSP{79.06}{2.53}&\NSP{29.89}{2.48}&\NSP{39.19}{2.84}\\&scSENet &\NSP{71.53}{3.57}&\NSP{79.05}{2.99}&\NSP{25.56}{1.97}&\NSP{34.11}{2.32}\\&FEDNet &\NSP{69.40}{3.68}&\NSP{77.62}{3.07}&\NSP{21.45}{4.35}&\NSP{29.08}{5.54}\\&CE-Net &\NSP{71.27}{4.57}&\NSP{79.65}{3.98}&\NSP{14.46}{5.16}&\NSP{20.17}{6.40}\\&CPFNet &\NSP{78.38}{2.53}&\NSP{86.00}{1.99}&\NSP{16.86}{5.42}&\NSP{22.84}{6.64}\\&UNetPP &\NSP{71.66}{3.48}&\NSP{79.15}{2.93}&\NSP{30.54}{1.50}&\NSP{40.01}{1.65}\\&AdaptNet &\NSP{74.42}{3.25}&\NSP{81.49}{2.64}&\NSP{49.55}{1.49}&\NSP{61.65}{1.63}\\&ReCal-Net &\NSP{66.88}{5.66}&\NSP{74.15}{5.67}&\NSP{37.99}{4.15}&\NSP{49.02}{4.64}\\
&DeepPyramid&\NSP{77.97}{3.78}&\NSP{84.95}{3.02}&\NSP{49.43}{2.06}&\NSP{60.79}{1.96}\\
\midrule\multirow{6}{*}{ResNet34}&scSENet &\NSP{77.30}{2.79}&\NSP{84.64}{2.21}&\NSP{44.36}{1.51}&\NSP{55.26}{1.62}\\&FEDNet &\NSP{76.86}{2.66}&\NSP{85.01}{2.05}&\NSP{40.46}{1.35}&\NSP{51.32}{1.42}\\&CE-Net &\NSP{34.78}{2.88}&\NSP{47.29}{2.90}&\NSP{36.55}{2.59}&\NSP{50.61}{2.82}\\&CPFNet &\NSP{44.92}{1.20}&\NSP{56.17}{1.50}&\NSP{43.71}{1.93}&\NSP{57.43}{2.01}\\&AdaptNet &\NSP{75.10}{2.99}&\NSP{82.34}{2.43}&\NSP{54.15}{0.80}&\NSP{66.23}{0.80}\\&ReCal-Net &\NSP{69.76}{5.96}&\NSP{77.27}{5.98}&\NSP{48.66}{2.35}&\NSP{60.36}{2.84}\\\midrule\multirow{2}{*}{ResNet50}&UPerNet &\NSP{78.36}{2.72}&\NSP{85.51}{2.13}&\NSP{40.28}{1.33}&\NSP{50.82}{1.52}\\&DeepLabV3+ &\NSP{68.77}{2.31}&\NSP{78.14}{2.13}&\NSP{30.72}{4.95}&\NSP{41.50}{5.94}\\

\specialrule{.12em}{.05em}{.05em}
\end{tabular}
}
\end{table}

Within Table \ref{tab:quantitative-single-cross-domain}, a thorough comparison is made between the performance of various neural network architectures concerning intra-domain and cross-domain scenarios. These architectures are trained using our binary instrument annotations. The results clearly indicate statistical differences between Cataract-1k and CaDIS datasets. Concretely, the average dice coefficient for binary instrument segmentation equals $77\%$ within the Cataract-1k dataset. However, this performance metric markedly contracts, remaining confined to around $67\%$ (with AdaptNet illustrating $66.23\%$) when extended to the CaDIS dataset. This considerable variance starkly underscores the substantial domain shift inherently present between these two datasets.
These results demonstrate the necessity of strategically exploring semi-supervised and domain adaptation techniques to elevate instrument segmentation performance in cataract surgery videos with cross-dataset domain shifts \cite{ghamsarian2023domain}.

\section*{Usage Notes}

The datasets are licensed under CC BY. For further legal details, we kindly request the readers to refer to the complete \href{https://creativecommons.org/licenses/by/4.0/}{license terms}. 

Besides, anyone can view the sample videos and images from the dataset and access the GitHub repository for dataset preparation codes.

\section*{Code availability}
We provide all code for mask creation using JSON annotations and phase extraction using CSV files, as well as their usage instruction in the GitHub repository of this paper (accessible for anonymous review in Figshare).

\section*{Acknowledgements} 
We would like to thank Daniela Stefanics for helping us in annotating the datasets to the highest quality. 

\noindent This work was funded by Haag-Streit Foundation Switzerland and the FWF Austrian Science Fund under grant P~32010-N38.

\noindent This work is performed under ethics committee approval (EK 28/17).

\section*{Author contributions statement}

N.G. wrote the original draft. R.S., M.Z., S.W., K.S., Y.E., and D.P. acquired the projects' funding. R.S. and K.S. were responsible for the projects' supervision. N.G. and K.S. organized the annotation process. Y.E. and D.P. provided expert information on the cataract surgery phases, relevant anatomical structures, and instruments used in regular cataract surgery videos. N.G. and D.P. reviewed and corrected the annotations. N.G. designed, implemented, and evaluated semantic segmentation experiments of the technical validation. N.G. and S.N. designed phase recognition experiments of technical validation. S.N. implemented and evaluated phase recognition experiments of technical validation. All authors reviewed the manuscript.
\section*{Competing interests} 

The authors declare no competing interests.
\bibliography{bibtex}
\end{document}